\documentclass[a4paper,fleqn]{cas-sc}

\usepackage[numbers]{natbib}

\usepackage{multirow}
\usepackage{adjustbox}
\usepackage{booktabs}
\usepackage{arydshln}
\usepackage{pgfplots}
\usepackage{caption}
\usepackage{subcaption}
\usepackage{color, xcolor}
\usepackage[framemethod=TikZ]{mdframed}
\usepackage{tikz}
\usepackage[edges]{forest}

\pgfplotsset{compat=1.18}

\newcommand{\quotes}[1]{``#1''}
\renewcommand{\vec}[1]{\mathbf{#1}}

\definecolor{lightgray}{rgb}{0.97,0.97,0.97} 
\definecolor{darkgray}{rgb}{0.8,0.8,0.8} 
\definecolor{lightgreenprompt}{HTML}{90EE90}
\definecolor{hiddendraw}{RGB}{205, 44, 36}

\colorlet{CellColor}{blue!50}

\mdfsetup{frametitlealignment=\center}

\makeatletter
\def\adl@drawiv#1#2#3{%
        \hskip.5\tabcolsep
        \xleaders#3{#2.5\@tempdimb #1{1}#2.5\@tempdimb}%
                #2\z@ plus1fil minus1fil\relax
        \hskip.5\tabcolsep}
\newcommand{\cdashlinelr}[1]{%
  \noalign{\vskip\aboverulesep
           \global\let\@dashdrawstore\adl@draw
           \global\let\adl@draw\adl@drawiv}
  \cdashline{#1}
  \noalign{\global\let\adl@draw\@dashdrawstore
           \vskip\belowrulesep}}
\makeatother

\begin{document}
\let\WriteBookmarks\relax
\def\floatpagepagefraction{1}
\def\textpagefraction{.001}

\title [mode = title]{Cross-lingual Aspect-Based Sentiment Analysis: A Survey on Tasks, Approaches, and Challenges\tnotemark[1]}                      
\tnotetext[1]{\texorpdfstring{Submitted version prior to peer review. Updated version accepted in \textit{Information Fusion}:  \href{https://doi.org/10.1016/j.inffus.2025.103073}{doi.org/10.1016/j.inffus.2025.103073}.}{Submitted version prior to peer review. Updated version accepted in Information Fusion: doi.org/10.1016/j.inffus.2025.103073}}

\shorttitle{Cross-lingual Aspect-Based Sentiment Analysis: A Survey on Tasks, Approaches, and Challenges}

\shortauthors{\v{S}m\'{i}d et~al.}

\author[1, 2]{Jakub \v{S}m\'{i}d}[orcid=0000-0002-4492-5481]
\ead{jaksmid@kiv.zcu.cz}

\cormark[1]

\credit{Conceptualization, Validation, Formal analysis, Investigation, Resources, Writing – original draft, Visualization, Supervision, Project administration}

\author[1,2]{Pavel Kr\'{a}l}[
        orcid=0000-0002-3096-675X
]

\ead{pkral@kiv.zcu.cz}

\credit{Writing – review \& editing, Supervision}

\affiliation[1]{organization={University of West Bohemia in Pilsen, Faculty of Applied Sciences, Department of Computer Science and Engineering},
    addressline={Univerzitni 8}, 
    city={Pilsen},
    citysep={}, 
    postcode={301 00}, 
    country={Czech Republic},
    }
    
\affiliation[2]{organization={NTIS -- New Technologies for the Information Society},
    addressline={Univerzitni 8}, 
    city={Pilsen},
    citysep={}, 
    postcode={301 00}, 
    country={Czech Republic}}

\cortext[cor1]{Corresponding author}

\begin{abstract}
    Aspect-based sentiment analysis (ABSA) is a fine-grained sentiment analysis task that focuses on understanding opinions at the aspect level, including sentiment towards specific aspect terms, categories, and opinions. While ABSA research has seen significant progress, much of the focus has been on monolingual settings. Cross-lingual ABSA, which aims to transfer knowledge from resource-rich languages (such as English) to low-resource languages, remains an under-explored area, with no systematic review of the field. This paper aims to fill that gap by providing a comprehensive survey of cross-lingual ABSA. We summarize key ABSA tasks, including aspect term extraction, aspect sentiment classification, and compound tasks involving multiple sentiment elements. Additionally, we review the datasets, modelling paradigms, and cross-lingual transfer methods used to solve these tasks. We also examine how existing work in monolingual and multilingual ABSA, as well as ABSA with LLMs, contributes to the development of cross-lingual ABSA. Finally, we highlight the main challenges and suggest directions for future research to advance cross-lingual ABSA systems.

\end{abstract}


\maketitle

\section{Introduction}
Aspect-based sentiment analysis (ABSA), a natural language processing (NLP) task, has become essential for analysing opinions expressed in online content, especially in environments where customer feedback and reviews shape decisions. Businesses, for example, rely heavily on customer sentiment to refine their products, improve services, and develop marketing strategies. However, given the sheer volume of unstructured online text, manually processing these opinions is infeasible. This growing demand for scalable solutions has driven the need for automated systems capable of extracting and interpreting sentiment from user-generated content, leading to the rise of sentiment analysis and opinion mining~\cite{liu2010sentiment}. 

Traditional sentiment analysis often focuses on broader evaluation of entire documents or sentences. These approaches assume that the text conveys a single sentiment towards a single topic, which oversimplifies the complexity of real-world opinions. In reality, different aspects of a product or service may evoke distinct sentiments within a single review. For instance, a customer might express satisfaction with a product's design but disappointment with its price. This gap led to the development of aspect-based sentiment analysis, which focuses on extracting sentiment at a more granular level. Rather than analysing the sentiment of the entire text, ABSA identifies specific entities (such as products or services) and their individual attributes (such as features or qualities) that users express opinions about. By distinguishing between these aspects ABSA enables a deeper understanding of customer feedback and provides more actionable insights. It has proven particularly valuable for businesses seeking to build fine-grained opinion summaries, offering a richer perspective on customer sentiment for applications in various sectors.

\begin{figure}[ht!]
    \centering
    \includegraphics{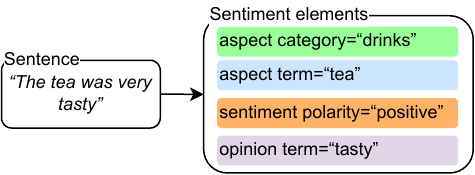}
    \caption{Example of a review with four ABSA sentiment elements.}
    \label{fig:example}
\end{figure}

ABSA research focuses on four sentiment elements~\cite{zhang2022survey}: aspect term ($a$), aspect category ($c$), sentiment polarity ($p$), and opinion term ($o$). Given the review: \textit{\quotes{The tea was very tasty}}, the corresponding sentiment elements are \textit{\quotes{tea}}, \textit{\quotes{drinks}}, \textit{\quotes{positive}}, and \textit{\quotes{tasty}}, respectively, as shown in Figure~\ref{fig:example}. Early ABSA research primarily focuses on \textit{single} ABSA tasks identifying each sentiment element separately, such as aspect term extraction or aspect category detection. More recent work shifts towards \textit{compound} ABSA tasks aiming at extracting two, three or four sentiment elements together, making these tasks more complex and challenging. For example, end-to-end ABSA identifies the aspect terms and their corresponding sentiment polarities simultaneously.

\begin{figure}[ht!]
    \centering
    \includegraphics[width=0.9\linewidth]{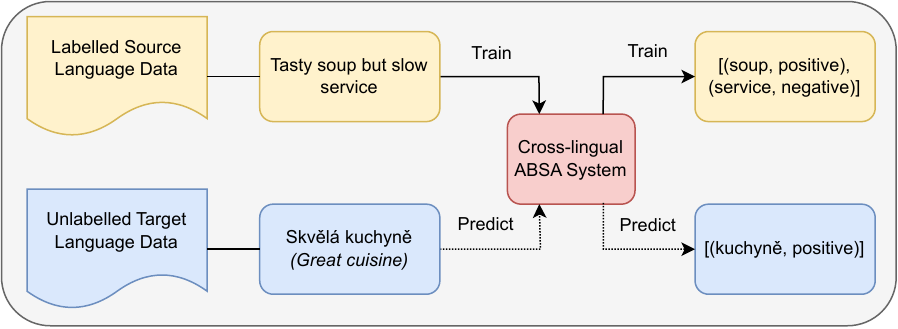}
    \caption{Visualization of cross-lingual ABSA system.}
    \label{fig:xabsa}
\end{figure}

Most existing research on ABSA has been conducted in a monolingual setting, with a strong focus on English. However, real-world users write reviews in many different languages~\cite{pontiki-etal-2016-semeval}. A significant challenge is the limited availability of labelled datasets in languages other than English, and obtaining such annotated data is both expensive and time-consuming, especially for low-resource languages. This has led to the rise of cross-lingual ABSA as an important area of research. The goal of cross-lingual aspect-based sentiment analysis is to leverage labelled data from a \textit{source language}, usually a resource-rich language like English, and transfer that knowledge to a \textit{target language}, which typically lacks annotated data, as depicted in Figure~\ref{fig:xabsa}. Due to the complexity of this task, cross-lingual ABSA often focuses on simpler tasks like aspect term extraction, aspect sentiment classification, and end-to-end ABSA, while monolingual studies have more recently tackled more intricate, compound tasks.

Several cross-lingual ABSA approaches have been explored, leveraging various forms of information fusion to address linguistic and cultural diversity across languages. Earlier methods combine machine translation with alignment algorithms to label translated data~\cite{lambert-2015-aspect, zhou2015clopinionminer} or employ cross-lingual word embeddings to enable models to switch languages by replacing the embedding layer~\cite{barnes-etal-2016-exploring, akhtar-etal-2018-solving}. More recent work leverages multilingual pre-trained language models (mPLMs) such as multilingual BERT (mBERT)~\cite{devlin-etal-2019-bert} and XLM-RoBERTa (XLM-R)~\cite{conneau-etal-2020-unsupervised}, which have become standard tools for cross-lingual NLP tasks~\cite{pmlr-v119-hu20b}. These models, trained on large multilingual corpora, fuse syntactic and semantic patterns across languages, enabling improved cross-lingual transfer. However, cross-lingual ABSA remains challenging, particularly in zero-shot settings where no labelled target language data is available, due to language-specific nuances such as slang, abbreviations, and unique aspect terms~\cite{li2020unsupervised}. Recent advancements in this field often involve the fusion of mPLMs and machine translation with complementary techniques, such as alignment-free label projection~\cite{zhang-etal-2021-cross}, distillation on unlabelled target language data~\cite{zhang-etal-2021-cross}, contrastive learning~\cite{lin2023clxabsa}, and dynamic weighted loss strategies~\cite{LinXABSA}, which collectively aim to bridge the linguistic gap and enhance model robustness in diverse languages. Recent NLP research has heavily focused on large language models (LLMs), but their application to cross-lingual ABSA remains limited to a few exploratory studies.

Existing ABSA surveys have a number of limitations. Earlier surveys~\cite{liu2010sentiment, 9260964} take a broad approach to sentiment analysis without providing a detailed examination of ABSA. While later surveys focus on ABSA specifically~\cite{7286808, 8976252, 10.1145/3503044, zhang2022survey, CHAUHAN2023100576}, including modern neural network approaches, and some delve into specialized areas such as multimodal ABSA~\cite{ZHAO2024102552}, they either completely overlook or provide limited coverage of cross-lingual approaches. This oversight leaves a gap in the systematic review of this increasingly important area. The primary aim of this paper is to bridge this gap by presenting a comprehensive and focused review of cross-lingual aspect-based sentiment analysis. We provide an in-depth discussion of ABSA tasks, methods for addressing them, available datasets, and general modelling paradigms. Furthermore, we review various techniques for cross-lingual transfer in ABSA, highlight existing challenges, and propose future research directions to advance the field.

The main contributions of this paper include:
\begin{itemize}
    \item This paper is the first detailed survey focusing on cross-lingual aspect-based sentiment analysis, addressing a gap in the existing literature.
    \item This survey provides a comprehensive overview of the available datasets, ABSA tasks, and general modelling paradigms relevant to cross-lingual ABSA.
    \item This survey systematically reviews techniques for cross-lingual transfer in ABSA, highlighting advancements in multilingual pre-trained language models, machine translation, and other fusion-based approaches.
    \item This survey offers an in-depth analysis of solutions for specific cross-lingual ABSA tasks.
    \item This survey reviews related work in monolingual and multilingual ABSA, cross-lingual and multilingual sentiment analysis, as well as ABSA with LLMs, and examines their relevance to cross-lingual ABSA.
    \item This survey discusses the remaining challenges in cross-lingual ABSA and outlines potential future research directions to inspire advancements in this field.
\end{itemize}

The structure of the paper is as follows. Section~\ref{sec:background} provides an overview of ABSA, covering its definitions, tasks, and available multilingual datasets. Section~\ref{sec:paradigms} discusses various modelling paradigms for ABSA. Section~\ref{sec:cl_transfer} focuses on techniques for cross-lingual transfer. Section~\ref{sec:cl_tasks} reviews existing solutions for different cross-lingual ABSA tasks in detail. Section~\ref{sec:related} provides an overview of related research on innovative approaches in monolingual ABSA, multilingual ABSA, multilingual and cross-lingual sentiment analysis, and the application of large language models for ABSA, while also highlighting their connection to cross-lingual ABSA. Section~\ref{sec:future} explores the limitations of current cross-lingual ABSA research and outlines potential directions for future studies. Finally, Section~\ref{sec:conclusion} summarizes the findings and contributions of this study to the field of cross-lingual ABSA.

\section{ABSA Background}
\label{sec:background}
This section provides the background for aspect-based sentiment analysis, including definitions, tasks and datasets.

\subsection{Definitions}
Aspect-based sentiment analysis, also known as aspect-level sentiment analysis, performs fine-grained analysis, looking at opinions consisting of a sentiment and a target instead of looking at documents or sentences as a whole~\cite{liu2022sentiment}. The target can be described by aspect term ($a$) or aspect category ($c$), while the sentiment can be described by sentiment polarity ($p$) or opinion term ($o$). These sentiment elements can be described as:
\begin{itemize}
    \item \textbf{Aspect term ($a$)} is the specific target of the opinion directly mentioned in the text. For example, in the sentence \textit{\quotes{The laptop's battery lasts all day}}, the term \textit{\quotes{battery}} is the aspect term. If the target is implied rather than explicitly stated (e.g. in \textit{\quotes{It lasts forever!}}), the aspect term may be labelled as \textit{\quotes{null}}.
    \item \textbf{Aspect category ($c$)} refers to a distinct attribute of an entity and is expected to belong to a predefined category set for a particular domain. For instance, in the electronics domain, categories like \textit{\quotes{battery life}} or \textit{\quotes{design}} could be used to evaluate different aspects of a smartphone.
    \item \textbf{Sentiment polarity ($p$)} refers to the emotional tone or attitude associated with an aspect category or aspect term, typically classified as positive, negative, or neutral. For example, the sentiment in \textit{\quotes{The battery lasts all day}} is positive, while in \textit{\quotes{The battery drains too fast}}, it is negative.
    \item \textbf{Opinion term ($o$)} represents the words or phrases the speaker uses to convey their sentiment about the target. In the sentence \textit{\quotes{The laptop's battery lasts all day}}, the term \textit{\quotes{lasts all day}} is the opinion term describing the battery. Like aspect terms, opinion terms can be implicit, for example, in \textit{\quotes{I am going back}}.
\end{itemize}

Given the sentiment elements definitions, we define ABSA as follows:
\begin{quote}
    \textbf{Aspect-based sentiment analysis} involves identifying and extracting sentiment elements of interest from a given text, focusing on either individual sentiment elements or multiple elements that may exhibit dependency relationships. This process aims to uncover the sentiments associated with specific aspects of entities, enabling a comprehensive understanding of consumer opinions and facilitating informed decision-making in various applications.
\end{quote}

We can also define cross-lingual ABSA as follows:
\begin{quote}
    \textbf{Cross-lingual aspect-based sentiment analysis} aims to leverage labelled data from a resource-rich source language to enable aspect-based sentiment analysis in a low-resource target language. This approach facilitates the effective transfer of knowledge across languages, addressing the challenges posed by linguistic differences and enhancing the capacity for sentiment extraction regarding specific aspects in diverse languages.
\end{quote}

In the literature, some terms are often used interchangeably but can have different meanings depending on the context. For instance, \textit{\quotes{aspect}}, \textit{\quotes{opinion target}}, \textit{\quotes{entity}}, and \textit{\quotes{target}} are commonly used to describe what the opinion is about. However, they can refer to either an aspect category or an aspect term, which can lead to confusion and incomplete literature reviews. In this survey, we use the most widely accepted terminology and clearly distinguish related concepts. As such, we differentiate between \textit{\quotes{aspect term}} and \textit{\quotes{aspect category}} while using \textit{\quotes{target}} or \textit{\quotes{aspect}} as general terms for an opinion target.

\subsection{Tasks}
\label{sec:tasks}
\citet{zhang2022survey} classify ABSA tasks into two main categories, single and compound, depending on whether the output involves a single sentiment element or multiple linked elements. While single tasks are generally simpler, they fail to capture more comprehensive opinions at the aspect level. Consequently, recent research has shifted towards tackling compound tasks, which provide a more complete understanding of opinions. This shift is significant as it allows for a more nuanced analysis of opinions, leading to more accurate sentiment analysis results.

\tikzstyle{mybox}=[
    rectangle,
    draw=hiddendraw,
    rounded corners,
    text opacity=1,
    minimum height=2em,
    minimum width=22em,
    inner sep=2pt,
    align=center,
    fill opacity=.5,
    ]
    
\tikzstyle{leaf}=[
    mybox,
    minimum height=1em,
    fill=green!40, 
    text width=23.7em,
    text=black,
    align=left,
    font=\scriptsize,
    inner xsep=2pt,
    inner ysep=1pt,
]

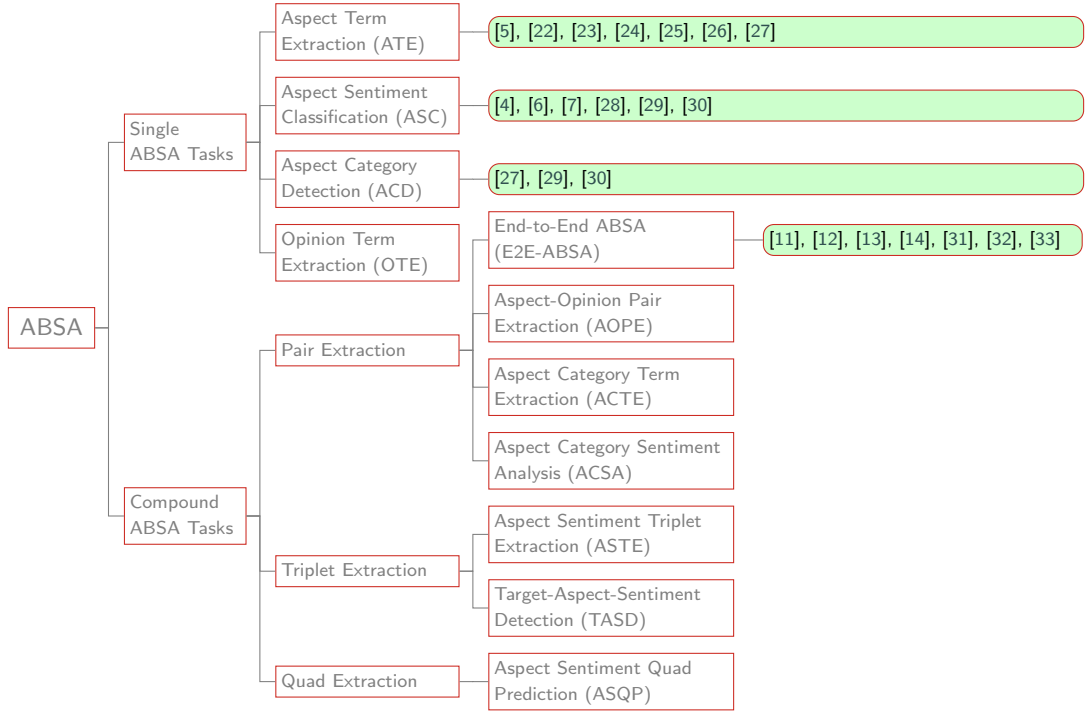
\begin{figure}[ht!]
  \centering
  \begin{forest}
    forked edges,
    for tree={
      grow=east,
      reversed=true,
      anchor=base west,
      parent anchor=east,
      child anchor=west,
      base=left,
      font=\small,
      rectangle,
      draw=hiddendraw,
      align=left,
      minimum width=3.5em,
      s sep=6pt,
      inner xsep=2pt,
      inner ysep=1pt,
      ver/.style={rotate=90, child anchor=north, parent anchor=south, anchor=center},
    },
    where level=1{text width=4.5em,font=\scriptsize}{},
    where level=2{text width=7em,font=\scriptsize}{},
    where level=3{text width=9.5em,font=\scriptsize}{},
    [ABSA
    [Single\\ABSA Tasks
        [Aspect Term\\Extraction (ATE)
           [\cite{zhou2015clopinionminer}{,} \cite{10.1145/2661829.2662019}{,} \cite{klinger-cimiano-2015-instance}{,} \cite{10.1145/3273931}{,} \cite{wang2018transition}{,} \cite{jebbara-cimiano-2019-zero}{,} \cite{9585242}
            ,leaf
            ]
        ]
        [Aspect Sentiment\\Classification (ASC)
            [\cite{lambert-2015-aspect}{,} \cite{barnes-etal-2016-exploring}{,} \cite{akhtar-etal-2018-solving}{,} \cite{wu2021reinforced}{,} \cite{doi:10.1080/24751839.2023.2173843}{,} \cite{nam2024kpc},leaf]
        ]
        [Aspect Category\\Detection (ACD)
            [\cite{9585242}{,} \cite{doi:10.1080/24751839.2023.2173843}{,} \cite{nam2024kpc}
                ,leaf
            ]
        ]
        [Opinion Term\\Extraction (OTE)]
    ]
    [Compound \\ ABSA Tasks
        [Pair Extraction
            [End-to-End ABSA\\ (E2E-ABSA)         
                [\cite{li2020unsupervised}{,} \cite{zhang-etal-2021-cross}{,} \cite{lin2023clxabsa}{,} \cite{LinXABSA}{,} \cite{9550786}{,} \cite{10031113}{,} \cite{wu2024evaluatingzeroshotmultilingualaspectbased}
                ,leaf,text width=12.5em, minimum width=12.5em]
            ]
            [Aspect-Opinion Pair\\Extraction (AOPE)]
            [Aspect Category Term\\Extraction (ACTE)]
            [Aspect Category Sentiment\\Analysis (ACSA)]
        ]
        [Triplet Extraction
            [Aspect Sentiment Triplet\\ Extraction (ASTE)]
            [Target-Aspect-Sentiment\\ Detection (TASD)]
        ]
        [Quad Extraction
            [Aspect Sentiment Quad\\ Prediction (ASQP)]
        ]
    ]
    ]
  \end{forest}
  \caption{Taxonomy of ABSA tasks, highlighting representative papers addressing each task in cross-lingual settings. Tasks without references indicate gaps in cross-lingual research.}
  \label{fig:absa-tasks}
\end{figure}

Figure~\ref{fig:absa-tasks} illustrates the taxonomy of ABSA tasks, highlighting representative works in cross-lingual settings. Additionally, the figure identifies tasks for which no references are available, indicating gaps in cross-lingual ABSA research and highlighting opportunities for future exploration. Table~\ref{tab:absa-tasks} presents the input and output examples for each task. Below is a brief overview of the key ABSA tasks:
\begin{itemize}
    \item \textbf{Aspect term extraction} (ATE) focuses on identifying specific aspect terms that are the target of opinions expressed in the text.
    \item \textbf{Aspect category detection} (ACD) aims to determine the broad categories under which the aspects discussed in the text fall.
    \item \textbf{Opinion term extraction} (OTE) identifies opinion expressions towards specific aspects. It includes two sub-tasks. Aspect opinion co-extraction (AOCE) extracts aspect and opinion terms without linking them.
    However, it does not link the aspect terms and opinions together, meaning it is still considered a single ABSA task.
    The second task is target-oriented opinion word extraction (TOWE), which aims to extract the corresponding opinion terms given a specific aspect term.
    \item \textbf{Aspect sentiment classification} (ASC) predicts the sentiment polarity for a specific aspect, which may be an aspect term, category, or a combination of both.
    \item \textbf{Aspect-opinion pair extraction} (AOPE) focuses on extracting both aspect terms and their corresponding opinion expressions from the text.
    \item \textbf{End-to-end ABSA} (E2E-ABSA) extracts both aspect terms and their associated sentiment polarities.
    \item \textbf{Aspect category term extraction} (ACTE) aims to extract aspect terms alongside their respective aspect categories.
    \item \textbf{Aspect category sentiment analysis} (ACSA) involves detecting aspect categories discussed in the text and predicting their corresponding sentiment polarities.
    \item \textbf{Aspect sentiment triplet extraction} (ASTE) extracts (aspect term, opinion term, sentiment polarity) triplets.
    \item \textbf{Target-aspect-sentiment detection} (TASD) extracts (aspect term, aspect category, sentiment polarity) triplets for a given sentence. It is also referred to as aspect-category-sentiment detection (ACSD).
    \item \textbf{Aspect sentiment quad prediction} (ASQP) focuses on extracting all (aspect term, opinion term, aspect category, sentiment polarity) quadruplets from a sentence.
\end{itemize}

\begin{table}[ht!]
    \centering
    \caption{An overview of the input and output examples for each ABSA task considering an input sentence \textit{s}: \textit{\quotes{The tea was tasty, but the waiter was rude.}.}}
    \begin{adjustbox}{width=\linewidth}
    \begin{tabular}{@{}llll@{}}
        \toprule
        \textbf{Task} & \textbf{\begin{tabular}[c]{@{}l@{}}
                                    Example\\ input
        \end{tabular}} & \textbf{Output} & \textbf{\begin{tabular}[c]{@{}l@{}}
                                     Example\\ output
        \end{tabular}}                   \\
        \midrule
        \multicolumn{4}{@{}l@{}}{\textit{Single tasks}}                   \\
        Aspect term extraction (ATE)                  & \textit{s}    & \{$a$\}      & \{\textit{\quotes{tea}}, \textit{\quotes{service}}\}                                              \\
        Aspect category detection (ACD) & \textit{s} & \{$c$\}    & \{\textit{\quotes{food}}, \textit{\quotes{service}}\}                                                                               \\
        Aspect opinion co-extraction (AOCE)                      & \textit{s} & \{$a$\}, \{$o$\} & \{\textit{\quotes{tea}}, \textit{\quotes{waiter}}\}, \{\textit{\quotes{tasty}}, \textit{\quotes{rude}}\}                                                                                \\
        \multirow{2}{*}{Target-oriented opinion word extraction (TOWE)}  & \textit{s}, \textit{\quotes{tea}}    & $o_1$ & \textit{\quotes{tasty}}                                                                       \\
        & \textit{s}, \textit{\quotes{waiter}} & $o_2$ & \textit{\quotes{rude}}                                                                       \\
        \multirow{2}{*}{Aspect sentiment classification (ASC)} & \textit{s}, \textit{\quotes{tea}} & $p_1$ &\textit{\quotes{positive}} \\
        & \textit{s}, \textit{\quotes{waiter}}          & $p_2$ & \textit{\quotes{negative}}                                                  \\
        \hdashline
        \multicolumn{4}{@{}l@{}}{\textit{Compound tasks}}              \\
        Aspect-opinion pair extraction (AOPE)                  & \textit{s}     & \{($a$, $o$)\}     & \{(\textit{\quotes{tea}}, \textit{\quotes{tasty}}), (\textit{\quotes{waiter}}, \textit{\quotes{rude}})\}                         \\
        End-to-end ABSA (E2E-ABSA)                  & \textit{s}     & \{($a$, $p$)\}     & \{(\textit{\quotes{tea}}, \textit{\quotes{positive}}), (\textit{\quotes{waiter}}, \textit{\quotes{negative}})\} \\
        Aspect category term extraction (ACTE)                  & \textit{s}     & \{($a$, $c$)\}     & \{(\textit{\quotes{tea}}, \textit{\quotes{food}}), (\textit{\quotes{waiter}}, \textit{\quotes{service}})\}                        \\
        Aspect category sentiment analysis (ACSA)           & \textit{s}     & \{($c$, $p$)\}     & \{(\textit{\quotes{food}}, \textit{\quotes{positive}}), (\textit{\quotes{service}}, \textit{\quotes{negative}})\}             \\
        Aspect sentiment triplet extraction (ASTE)                  & \textit{s}    & \{($a$, $o$, $p$)\}      & \begin{tabular}[c]{@{}l@{}}\{(\textit{\quotes{tea}}, \textit{\quotes{tasty}}, \textit{\quotes{positive}}),\\ (\textit{\quotes{waiter}}, \textit{\quotes{rude}}, \textit{\quotes{negative}})\}
        \end{tabular}                  \\
        Target-aspect-sentiment detection (TASD)                  & \textit{s}     &  \{($a$, $c$, $p$)\}   & \begin{tabular}[c]{@{}l@{}}\{(\textit{\quotes{tea}}, \textit{\quotes{food}}, \textit{\quotes{positive}}),\\ (\textit{\quotes{waiter}}, \textit{\quotes{service}}, \textit{\quotes{negative}})\}
        \end{tabular}                  \\
        Aspect sentiment quad prediction (ASQP)                  & \textit{s}      & \{($a$, $c$, $o$, $p$)\}    & \begin{tabular}[c]{@{}l@{}}\{(\textit{\quotes{tea}}, \textit{\quotes{food}}, \textit{\quotes{tasty}}, \textit{\quotes{positive}}),\\ (\textit{\quotes{waiter}}, \textit{\quotes{service}}, \textit{\quotes{rude}}, \textit{\quotes{negative}})\}
        \end{tabular}                  \\ \bottomrule
    \end{tabular}
    \end{adjustbox}
    \label{tab:absa-tasks}
\end{table}

Given the difficulty of cross-lingual ABSA compared to monolingual ABSA and also much less cross-lingual aspect-based sentiment analysis research, most of the cross-lingual ABSA work focuses on simple tasks such as aspect term extraction and aspect sentiment classification, and from compound tasks, only on end-to-end ABSA. This lack of research on compound ABSA tasks in cross-lingual settings presents one of the key challenges and limitations of current ABSA research.

\subsection{Datasets}
\label{sec:datasets}
Annotated datasets are vital for the development of ABSA methods. While most datasets are in English, this section focuses on commonly used multilingual datasets for cross-lingual ABSA and briefly mentions widely used monolingual datasets. Table~\ref{tab:datasets} summarizes the key multilingual datasets, including languages, domains, and annotated sentiment elements.

\begin{table}[ht!]
    \centering
    \caption{An overview of common multilingual ABSA benchmark datasets. Languages are denoted using ISO 639-1 codes. *The languages in the MultiAspectEmo dataset, except for Polish, are translations of the Polish reviews.}
    \begin{adjustbox}{width=0.8\linewidth}
    \begin{tabular}{llll}
    \toprule
    Dataset & Languages & Major Domains & Annotations \\
    \midrule
    SemEval-2016~\cite{pontiki-etal-2016-semeval} & ar, en, es, fr, nl, ru, tr, zh & restaurants, hotels, electronics & $a$, $c$, $p$ \\
    USAGE~\cite{klinger-cimiano-2014-usage} & en, de & products & $a$, $p$ \\
    OpeNER~\cite{agerri2013opener} & en, es & hotels & $a$, $p$ \\
    \cite{hyun-etal-2020-building} & en, kr & automotive & $a$, $p$ \\
    MultiAspectEmo*~\cite{10031113} & cs, en, es, fr, nl, pl  & school, medicine, hotels, products & $a$, $p$ \\
    PanoSent~\cite{10.1145/3664647.3680705} & en, es, zh & more than 100 domains & \textit{multimodal annotations} \\
    \bottomrule
    \end{tabular}
    \end{adjustbox}
    \label{tab:datasets}
\end{table}

The SemEval-2016 dataset~\cite{pontiki-etal-2016-semeval} contains real-world restaurant reviews and is one of the most widely used in cross-lingual ABSA. It builds on earlier shared task datasets from SemEval-2014~\cite{pontiki-etal-2014-semeval} and SemEval-2015~\cite{pontiki-etal-2015-semeval}, expanding from English to include Spanish, French, Dutch, Russian, and Turkish. This dataset features aspect terms, categories, and sentiment polarities, enabling various compound ABSA tasks. However, it lacks opinion term annotations, which were later introduced in other works~\cite{fan-etal-2019-target, xu-etal-2020-position, aste, cai-etal-2021-aspect, zhang-etal-2021-aspect-sentiment}, though only for English. In addition to the restaurant domain, the SemEval-2016 dataset includes hotel reviews in Arabic and electronics reviews in Chinese. Introduction of a new Czech dataset in the SemEval-2016 format, expanded and reannotated based on prior datasets~\cite{steinberger-etal-2014-aspect, hercig2016unsupervised}, enables future cross-lingual comparisons~\cite{smid-etal-2024-czech}.

Other notable multilingual datasets include the USAGE corpus~\cite{klinger-cimiano-2014-usage}, featuring English and German product reviews with annotated aspect terms and sentiment polarities, and the OpeNER dataset~\cite{agerri2013opener}, which contains hotel reviews in English and Spanish. A dataset in the automotive domain for English and Korean contains annotations of aspect terms and sentiment polarities~\cite{hyun-etal-2020-building}. The MultiAspectEmo dataset~\cite{10031113} contains original reviews in Polish in four domains, along with translations into English, Czech, Spanish, French, and Dutch. The PanoSent dataset~\cite{10.1145/3664647.3680705} covers more than 100 domains for multimodality in English, Chinese, and Spanish.

The Korean dataset~\cite{nam2024kpc}, focused on the restaurant domain, follows a format similar to the SemEval-2014 dataset~\cite{pontiki-etal-2014-semeval}. Notably, a portion of this dataset is a direct translation of the SemEval-2014 dataset into Korean.

For monolingual datasets, a Hindi dataset with electronics reviews is available~\cite{akhtar-etal-2016-aspect}, as well as a Japanese dataset for hotel reviews annotated with aspect categories and sentiment polarities~\cite{nakayama-etal-2022-large}. For Chinese, there is a dataset for ASC in QA-style reviews~\cite{wang-etal-2019-aspect}, and ASAP~\cite{bu-etal-2021-asap}, a large-scale dataset in the restaurant domain. Other English datasets include SentiHood~\cite{saeidi-etal-2016-sentihood}, MAMS~\cite{jiang-etal-2019-challenge}, and ARTS~\cite{xing-etal-2020-tasty}.

One key limitation of multilingual ABSA datasets is their limited language coverage and the absence of opinion term annotations. This gap has prevented the exploration of ABSA tasks involving opinion terms across multiple languages.

\subsection{Evaluation Metrics}
For evaluation, an exact match is the dominant method across ABSA datasets, where a prediction is correct only if all elements match annotations. Standard classification metrics like accuracy, precision, recall, and F1 score are then used to assess model performance.

\section{Modelling Paradigms}
\label{sec:paradigms}

Several paradigms are commonly used to solve ABSA tasks~\cite{zhang2022survey}, including sequence-level classification, token-level classification, and sequence-to-sequence modelling. Each paradigm represents a general computational framework designed to handle a specific input-output format, allowing them to be applied across various tasks. Beyond these unified paradigms, which address tasks in an end-to-end manner, more complex ABSA tasks may require the pipeline paradigm, where multiple models are connected sequentially to produce the final prediction. However, most cross-lingual ABSA research primarily relies on sequence-level and token-level classification paradigms; therefore, this section will discuss them in greater detail compared to other paradigms. Table~\ref{tab:modelling} presents representative cross-lingual ABSA works for different tasks and modelling paradigms.

\begin{table}[ht!]
\caption{Overview of various modelling paradigms used in cross-lingual ABSA, the tasks they address, and representative papers.}
\begin{adjustbox}{width=0.7\linewidth}
\begin{tabular}{@{}lll@{}}
\toprule
\textbf{Modelling Paradigm}                    & \textbf{Task}                                                                   & \textbf{Sources}                                                                                                                                                                                                             \\ \midrule
\multirow{2}{*}{Sequence-Level Classification} & Aspect Sentiment Classification & \cite{lambert-2015-aspect, barnes-etal-2016-exploring, akhtar-etal-2018-solving, wu2021reinforced, doi:10.1080/24751839.2023.2173843, nam2024kpc}                                                   \\ \cdashlinelr{2-3}
                                               & Aspect Category Detection        & \cite{9585242, doi:10.1080/24751839.2023.2173843, nam2024kpc}                                                                                                                                                                   \\ \cdashlinelr{1-3}
\multirow{2}{*}{Token-Level Classification}    & Aspect Term Extraction          & \cite{zhou2015clopinionminer, 10.1145/2661829.2662019, klinger-cimiano-2015-instance, 10.1145/3273931, wang2018transition, jebbara-cimiano-2019-zero, 9585242} \\ \cdashlinelr{2-3} 
                                               & End-to-End ABSA            & \cite{li2020unsupervised, zhang-etal-2021-cross, lin2023clxabsa, LinXABSA, 9550786, 10031113}                                                           \\ \cdashlinelr{1-3}
Sequence-to-Sequence Modelling & End-to-End ABSA                            & \cite{wu2024evaluatingzeroshotmultilingualaspectbased}\\
                                               \bottomrule
\end{tabular}
\end{adjustbox}
\label{tab:modelling}
\end{table}

In this section, we denote the dataset for a given ABSA task as $\mathcal{D}=\{\boldsymbol{x}_i, \boldsymbol{y}_i\}_{i=1}^{|\mathcal{D}|}$, where $\boldsymbol{x}_i=\{x_1, x_2, \ldots, x_n\}$ represents the $i$-th input sentence consisting of $n$ tokens, and $\boldsymbol{y}_i$ is the corresponding label.

\subsection{Sequence-Level Classification}
In sequence-level classification, a model typically consists of an encoder $\textbf{Enc}(\cdot)$ that processes input text $\boldsymbol{x}$ to extract features specific to the task. The extracted features are then passed through a classifier $\mathbf{CLS}(\cdot)$ to produce the label $\boldsymbol{y}$ as
\begin{equation}
    \boldsymbol{y} = \mathbf{CLS}(\mathbf{Enc}(\boldsymbol{x})).
\end{equation}

With advancements in deep learning, the encoder architecture can range from convolutional networks (CNNs)~\cite{fukushima2007neocognitron}, recurrent neural networks (RNNs) like LSTMs~\cite{hochreiter1997long}, and Transformers~\cite{vaswani2023attentionneed}, enabling the capture of rich contextual information.
For tasks like aspect sentiment classification, where inputs include both a sentence and an associated aspect term or category, the encoder must model not only the individual inputs but also their interactions. The classifier component is commonly designed as a multi-layer perceptron, often incorporating a pooling mechanism to derive the final prediction.

\begin{figure}[ht!]
    \centering
    \includegraphics[scale=1.3]{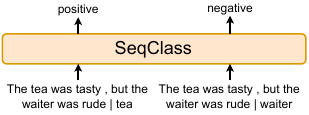}
    \caption{Sequence-level classification paradigm for the aspect sentiment classification task.}
    \label{fig:seqclass}
\end{figure}

Aspect category detection is often treated as a multi-label classification problem in this paradigm, where each aspect category is treated as a separate label. Sequence-level classification is frequently used for aspect sentiment classification, and this paradigm is commonly explored in cross-lingual settings. Figure~\ref{fig:seqclass} illustrates the sequence-level classification paradigm for aspect sentiment classification.

\subsection{Token-Level Classification}
In contrast to sequence-level classification, which assigns a single label to the entire input, token-level classification (also known as sequence labelling or sequence tagging) assigns a label to each individual token in the input sequence.
Similar to sequence-level classification, token-level classification begins with encoding the input text into contextualized features using an encoder $\mathbf{Enc} (\cdot)$.
A decoder $\mathbf{Dec}(\cdot)$ then predicts labels $y_1, y_2, \ldots, y_n$ for each token $x_1, x_2, \ldots, x_n$ in the input $\boldsymbol{x}$ as
\begin{equation}
    y_1, y_2, \ldots, y_n = \mathbf{Dec}(\mathbf{Enc}(x_1, x_2, \ldots, x_n)).
\end{equation}
The decoder is often implemented as a multi-layer perceptron with a softmax layer or with conditional random fields (CRF)~\cite{lafferty2001conditional}.
Various tagging schemes, such as the BIOES scheme (\textbf{B}eginning, \textbf{I}nside, \textbf{O}utside, \textbf{E}end, \textbf{S}ingleton)~\cite{tjong-kim-sang-veenstra-1999-representing} or the BIO scheme, can also be employed.

\begin{figure}[ht!]
    \centering
    \includegraphics[scale=1.3]{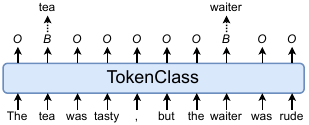}
    \caption{Token-level classification paradigm for the aspect term extraction task.}
    \label{fig:tokenclass}
\end{figure}

Aspect term extraction problem is commonly framed as a token-level classification task, as aspect terms are typically words or phrases within a sentence. This paradigm is also used for tasks like aspect opinion co-extraction and end-to-end ABSA, often with tagging schemes like BIOES. Since aspect term extraction and end-to-end ABSA are commonly examined in cross-lingual research, token-level classification is a popular approach in cross-lingual ABSA studies. Figure~\ref{fig:tokenclass} shows an example of a token-level classification paradigm for aspect term extraction.

\subsection{Sequence-to-Sequence Modelling}
\label{sec:seq2seq}
The sequence-to-sequence (Seq2Seq) paradigm processes an input sequence $\boldsymbol{x} = \{x_1, x_2, \ldots, x_n\}$ and generates an output sequence $\boldsymbol{y} = \{y_1, y_2, \ldots, y_m\}$.
Although commonly used in machine translation tasks, it is also applied in ABSA tasks, where it can generate label sequences or extract sentiment-related elements from input text. For instance, in the aspect term extraction task, an input $\boldsymbol{x}$ like \textit{\quotes{The steak with potatoes was delicious}} could yield an output $\boldsymbol{y}$ like \textit{\quotes{steak with potatoes}}.

The sequence-to-sequence paradigm typically employs an encoder-decoder model, such as the Transformer, where the encoder $\mathbf{Enc}(\cdot)$ extracts contextualized features from the input, and the decoder $\mathbf{Dec}(\cdot)$ generates tokens based on the encoded input and previously generated tokens as
\begin{equation}
    y_1, y_2, \ldots, y_m = \mathbf{Dec}(\mathbf{Enc}(x_1, x_2, \ldots, x_n)).
\end{equation}

With the advent of large language models, which are typically decoder-only models, this paradigm has gained prominence. In this case, the decoder generates token $y_i$ based only on previously generated tokens as
\begin{equation}
    y_i = \mathbf{Dec}(y_1, y_2, \ldots, y_{i-1}).
\end{equation}

\begin{figure}[ht!]
    \centering
    \includegraphics[scale=1.3]{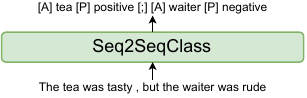}
    \caption{Sequence-to-sequence modelling paradigm for end-to-end ABSA.}
    \label{fig:seq2seq2clas}
\end{figure}

Although Seq2Seq modelling can be used to solve individual ABSA tasks (e.g. aspect term extraction), it is mainly employed for compound ABSA tasks. Figure~\ref{fig:seq2seq2clas} illustrates the use of sequence-to-sequence modelling for end-to-end ABSA. This paradigm has gained significant traction in recent years within monolingual ABSA research~\cite{zhang-etal-2021-aspect-sentiment, zhang-etal-2021-towards-generative, mao-etal-2022-seq2path, gou-etal-2023-mvp, xianlong-etal-2023-tagging, smid-priban-2023-prompt}, including studies with LLMs~\cite{gou-etal-2023-mvp, zhang-etal-2024-sentiment,smid-etal-2024-llama}. However, its potential in cross-lingual ABSA remains largely unexplored, with only one preliminary findings reported in a single study~\cite{wu2024evaluatingzeroshotmultilingualaspectbased} using fine-tuned LLMs.

\subsection{Pipeline Method}
\begin{figure}[ht!]
    \centering
    \includegraphics[scale=1.1]{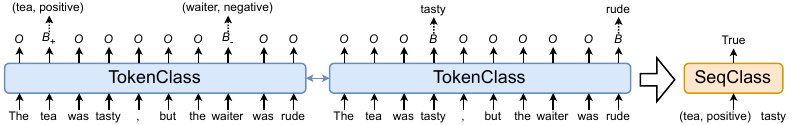}
    \caption{Pipeline method for aspect sentiment triplet extraction.}
    \label{fig:pipeline}
\end{figure}

The pipeline approach is frequently used for complex, compound ABSA tasks. It involves connecting multiple models sequentially, where the output of one model becomes the input for the next, as shown in~\ref{fig:pipeline}.
For instance, in the AOPE problem, an aspect term extraction model identifies aspect terms, which are then passed to a subsequent model to extract corresponding opinion terms. However, this method is prone to error propagation, where mistakes made by earlier models can negatively impact the overall performance. Despite its utility in handling complex tasks, the pipeline method has not been widely adopted in cross-lingual ABSA research.

\section{Cross-lingual Transfer}
\label{sec:cl_transfer}
In cross-lingual ABSA, cross-lingual transfer methods facilitate the fusion of linguistic knowledge between source and target languages, reducing the gap and enabling effective sentiment analysis across diverse linguistic contexts. The three primary approaches to achieving this transfer are the fusion of cross-lingual word embeddings, the integration of machine translation techniques, and the use of multilingual pre-trained language models.

\subsection{Cross-lingual Word Embeddings}
Pre-trained word embeddings, such as Word2Vec~\cite{word2vec}, FastText~\cite{bojanowski-etal-2017-enriching}, and GloVe~\cite{pennington-etal-2014-glove}, have been widely used in NLP to capture semantic relationships between words. These embeddings represent words as dense, continuous vectors in a high-dimensional space, capturing semantic relationships and syntactic patterns. Unlike sparse one-hot encodings, word embeddings leverage context from large corpora to create vectorized representations, where semantically similar words are close to each other in the embedding space.

Word2Vec, one of the pioneering models, learns embeddings using \textit{skip-gram} or \textit{continuous bag-of-words} (CBOW) techniques, predicting the context of a word or vice versa. FastText extends Word2Vec by incorporating subword information, which improves its ability to handle rare and morphologically complex words. GloVe, on the other hand, focuses on word co-occurrence statistics to capture the global context. These pre-trained embeddings have been widely adopted in various NLP tasks, including text classification, sentiment analysis, and machine translation, due to their ability to improve performance with minimal training data.

\begin{figure}[ht!]
    \centering
    \includegraphics[scale=1.3]{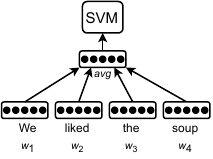}
    \caption{Example of averaging word embeddings for each word $w_n$ to obtain a single vector representing a whole sentence, which is then passed to a classifier such as SVM.}
    \label{fig:svm}
\end{figure}

\begin{figure}[ht!]
    \centering
    \includegraphics[scale=1.3]{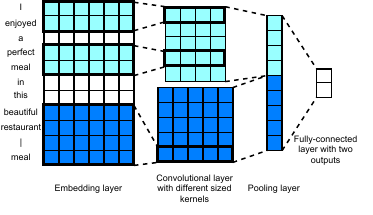}
    \caption{Example of a convolutional neural network with input embeddings.}
    \label{fig:cnn}
\end{figure}

These embeddings are widely used in tasks like text classification, sentiment analysis, and machine translation. For classifiers like support vector machines (SVMs), word embeddings are typically averaged across all words in a sentence, as shown in Figure~\ref{fig:svm}. While effective, this method may lose some of the embeddings' rich information. Neural network models, such as CNNs and LSTMs, can process sequences of word vectors more effectively, as illustrated in Figure~\ref{fig:cnn}, which shows a simple CNN architecture.

\subsubsection{Bilingual Word Embeddings}
In cross-lingual tasks, bilingual word embeddings (BWEs) extend monolingual embeddings to multilingual scenarios. For instance, the \textit{bilingual skip-gram algorithm}~\cite{luong-etal-2015-bilingual} modifies the traditional skip-gram model by incorporating bilingual objectives, ensuring that words in parallel sentences have similar embeddings across languages.

\citet{barnes-etal-2016-exploring} train bilingual embeddings on the English-Spanish Europarl corpus~\cite{koehn-2005-europarl}. Similarly, \citet{akhtar-etal-2018-solving} train bilingual word embeddings on a parallel English-Hindi and English-French corpora. They propose an approach to deal with out-of-vocabulary (OOV) words, i.e. words whose representations are missing in one of the bilingual embeddings vocabularies. Instead of generating random vectors or omitting these OOV words, which are common approaches to deal with them, they instead translate missing words into the other language, e.g. English, and use the word embedding corresponding to the translated word. Finally, they feed the word embeddings to the LSTM classifier.

\subsubsection{Word Embeddings Alignment}
\begin{figure}[ht!]
	\begin{subfigure}[b]{.45\linewidth}
		\centering
		\begin{adjustbox}{width=\linewidth}
			\begin{tikzpicture}
				
				\begin{axis}[ 
					xmin=-1.1,
					xmax=1.1,
					ymin=-1.1,
					ymax=1.1,
					xtick={-1,-0.5,0,0.5,1},
					ytick={-1,-0.5,0,0.5,1}
					]
					
					\node[label={270:{\scriptsize{lev}}},circle,fill, red,inner sep=1.5pt] at (axis cs:-0.88,0) {};
					\node[label={270:{\scriptsize{tygr}}},circle,fill, red,inner sep=1.5pt] at (axis cs:-0.74,0.04) {};
					\node[label={270:{\scriptsize{želva}}},circle,fill, red,inner sep=1.5pt] at (axis cs:-0.99,0.39) {};
					\node[label={270:{\scriptsize{krokodýl}}},circle,fill, red,inner sep=1.5pt] at (axis cs:-0.84,0.31) {};
					\node[label={270:{\scriptsize{auto}}},circle,fill, red,inner sep=1.5pt] at (axis cs:-0.8,0.95) {};
					\node[label={270:{\scriptsize{vlak}}},circle,fill, red,inner sep=1.5pt] at (axis cs:-0.65,0.8) {};
					\node[label={270:{\scriptsize{obličej}}},circle,fill, red,inner sep=1.5pt] at (axis cs:-0.4,0.47) {};
					\node[label={270:{\scriptsize{oko}}},circle,fill, red,inner sep=1.5pt] at (axis cs:-0.3,0.65) {};
					\node[label={270:{\scriptsize{tělo}}},circle,fill, red,inner sep=1.5pt] at (axis cs:-0.2,0.3) {};
					\node[label={90:{\scriptsize{tiger}}},rectangle,fill, green,inner sep=1.5pt] at (axis cs:0.2,-0.76) {};
					\node[label={90:{\scriptsize{lion}}},rectangle,fill, green,inner sep=1.5pt] at (axis cs:0.09,-0.8) {};
					\node[label={90:{\scriptsize{tortoise}}},rectangle,fill, green,inner sep=1.5pt] at (axis cs:0.27,-0.99) {};
					\node[label={90:{\scriptsize{crocodile}}},rectangle,fill, green,inner sep=1.5pt] at (axis cs:0.44,-0.92) {};
					\node[label={90:{\scriptsize{car}}},rectangle,fill, green,inner sep=1.5pt] at (axis cs:0.02,-0.4) {};
					\node[label={90:{\scriptsize{train}}},rectangle,fill, green,inner sep=1.5pt] at (axis cs:0.12,-0.26) {};
					\node[label={90:{\scriptsize{face}}},rectangle,fill, green,inner sep=1.5pt] at (axis cs:0.95,-0.85) {};
					\node[label={90:{\scriptsize{eye}}},rectangle,fill, green,inner sep=1.5pt] at (axis cs:0.9,-0.64) {};
					\node[label={90:{\scriptsize{body}}},rectangle,fill, green,inner sep=1.5pt] at (axis cs:0.8,-0.48) {};
				\end{axis}
			\end{tikzpicture}
		\end{adjustbox}
		\caption{Word embeddings before projection.}
	\end{subfigure}
	\hfill
	\begin{subfigure}[b]{.45\textwidth}
		\centering
		\begin{adjustbox}{width=\linewidth}
			\begin{tikzpicture}
				\begin{axis}[ 
					xmin=-1.1,
					xmax=1.1,
					ymin=-1.1,
					ymax=1.1,
					xtick={-1,-0.5,0,0.5,1},
					ytick={-1,-0.5,0,0.5,1}
					]
					\node[label={270:{\scriptsize{tygr}}},circle,fill, red,inner sep=1.5pt] at (axis cs:-0.6,-0.6) {};
					\node[label={270:{\scriptsize{lev}}},circle,fill, red,inner sep=1.5pt] at (axis cs:-0.44,-0.51) {};
					\node[label={270:{\scriptsize{želva}}},circle,fill, red,inner sep=1.5pt] at (axis cs:0.1,-0.5) {};
					\node[label={270:{\scriptsize{krokodýl}}},circle,fill, red,inner sep=1.5pt] at (axis cs:0.6,-0.8) {};
					\node[label={270:{\scriptsize{auto}}},circle,fill, red,inner sep=1.5pt] at (axis cs:-0.5,0.74) {};
					\node[label={270:{\scriptsize{vlak}}},circle,fill, red,inner sep=1.5pt] at (axis cs:-0.63,0.84) {};
					\node[label={270:{\scriptsize{obličej}}},circle,fill, red,inner sep=1.5pt] at (axis cs:0.9,0.9) {};
					\node[label={270:{\scriptsize{oko}}},circle,fill, red,inner sep=1.5pt] at (axis cs:0.91,0.49) {};
					\node[label={270:{\scriptsize{tělo}}},circle,fill, red,inner sep=1.5pt] at (axis cs:0.61,0.68) {};
					\node[label={90:{\scriptsize{tiger}}},rectangle,fill, green,inner sep=1.5pt] at (axis cs:-0.66,-0.62) {};
					\node[label={90:{\scriptsize{lion}}},rectangle,fill, green,inner sep=1.5pt] at (axis cs:-0.49,-0.55) {};
					\node[label={090:{\scriptsize{tortoise}}},rectangle,fill, green,inner sep=1.5pt] at (axis cs:0.15,-0.45) {};
					\node[label={90:{\scriptsize{crocodile}}},rectangle,fill, green,inner sep=1.5pt] at (axis cs:0.55,-0.75) {};
					\node[label={90:{\scriptsize{car}}},rectangle,fill, green,inner sep=1.5pt] at (axis cs:-0.47,0.84) {};
					\node[label={90:{\scriptsize{train}}},rectangle,fill, green,inner sep=1.5pt] at (axis cs:-0.7,0.85) {};
					\node[label={90:{\scriptsize{face}}},rectangle,fill, green,inner sep=1.5pt] at (axis cs:0.95,0.87) {};
					\node[label={90:{\scriptsize{eye}}},rectangle,fill, green,inner sep=1.5pt] at (axis cs:0.89,0.56) {};
					\node[label={90:{\scriptsize{body}}},rectangle,fill, green,inner sep=1.5pt] at (axis cs:0.56,0.64) {};
				\end{axis}
			\end{tikzpicture}
		\end{adjustbox}
		\caption{Word embeddings after projection.}
	\end{subfigure}
	\caption{Visualization of word embeddings for selected Czech and English words and their projection into a common space.}
	\label{fig:tranformation}
\end{figure}
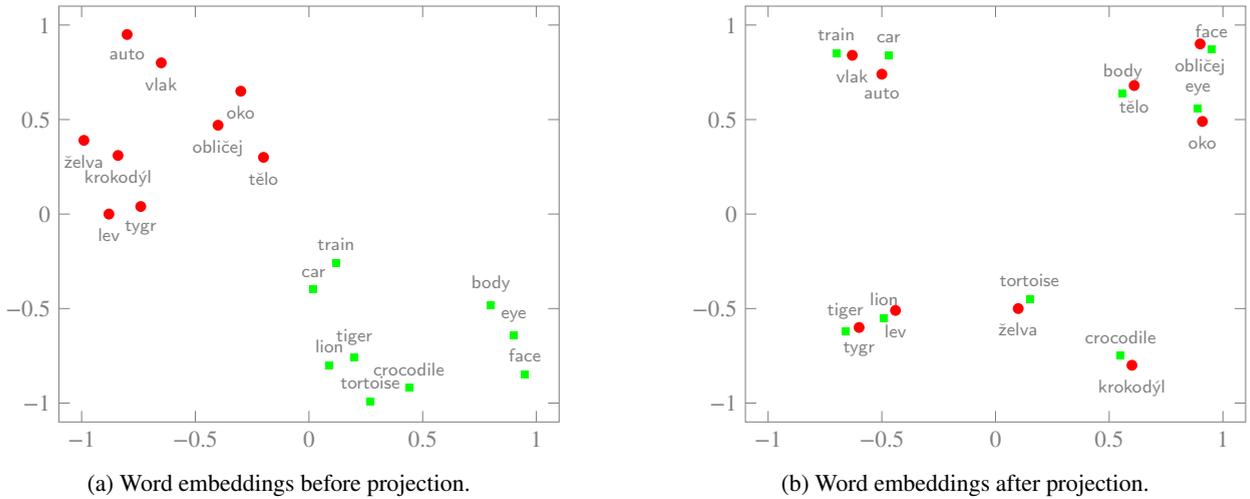

Another common approach involves aligning monolingual embeddings from different languages into a shared space, enabling cross-lingual transfer. The alignment can be achieved by learning a transformation matrix $\vec{W}$
that maps embeddings from one language into the space of another, effectively bridging the gap between languages. For example, given pre-trained word embeddings $\vec{X}^s$ for the source language $s$ and $\vec{X}^t$ for the target language $t$, a transformation matrix $\vec{W}^{s\to t}$ is learned to align these embeddings, such that the embeddings in language $s$ are mapped to the same space as those in language $t$, ideally preserving the semantic relationships between words across languages, as shown in Figure~\ref{fig:tranformation}. This can be formulated as minimizing the distance between corresponding word pairs in a bilingual dictionary, such as using regression, orthogonal or canonical methods~\cite{10.1613/jair.1.11640}. 

\citet{barnes-etal-2016-exploring} align English and Spanish embeddings by minimizing the mean square error between word pairs from a bilingual dictionary. \citet{jebbara-cimiano-2019-zero} use a singular value decomposition (SVD) on a dictionary of translated word pairs to obtain an orthogonal projection matrix from one vector space to another, preserving geometric properties during alignment. Additionally, they experiment with unsupervised methods, employing adversarial training and fine-tuning the alignment using synthetic bilingual dictionaries. \citet{wang2018transition} utilize a transition-based approach, aligning representations from different languages into a shared space with the help of an adversarial network.

\subsection{Machine Translation}
\label{sec:mt}
Another common approach in cross-lingual ABSA involves machine translation. Statistical machine translation (SMT) methods, such as phrase-based translation models, use large parallel datasets to translate input sentences from the source language into the target language. The input sentence is first translated with a translation tool from the source language to the target language, and an ABSA model can be fine-tuned with the created pseudo-labelled data in the target language. 

For tasks like aspect sentiment classification, machine translation is straightforward. However, for token-level tasks, such as aspect term extraction, the process becomes challenging due to potential discrepancies in token counts or positions between translated and original source language text. Therefore, word alignment tools such as FastAlign~\cite{dyer-etal-2013-simple} are employed. Since the effectiveness of this approach depends largely on the quality of both translation and label projection, numerous techniques have been introduced to enhance the quality of the data.

\citet{zhou2015clopinionminer} utilize Bing Translator for translation, word alignment and word segmentation for ATE. The word alignments are used to project aspect terms from annotated English data into translated Chinese data. For example, if an English aspect term aligns with multiple Chinese words, each Chinese word is labelled separately; if it aligns with consecutive Chinese words, the whole sequence is labelled as a single aspect.

\citet{klinger-cimiano-2015-instance} employ Google Translate for translation and FastAlign for labels alignment. In order to reduce the impact of translation and alignment errors, they filter out low-quality pairs. To assess translation quality, three measures are considered: the probability of the source sentence using a language model for unannotated text in the source language, the probability of the machine-translated sentence using a language model for unlabelled text in the target language, and the likelihood of correct alignment based on alignment probability.

\begin{figure}[ht!]
    \centering
    \includegraphics[width=0.8\linewidth]{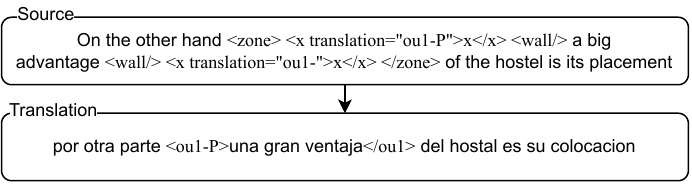}
    \caption{Source text that includes reordering constraint markup along with the code to transfer tags and their corresponding translation, as proposed by \citet{lambert-2015-aspect}.}
    \label{fig:constraintsmt}
\end{figure}

To solve the alignment problems during translation, \citet{lambert-2015-aspect} translates entire sentences while preserving opinionated segments through reordering constraints, ensuring that text between segment boundaries remains intact and unaltered by external text. They employ the Moses SMT toolkit~\cite{31b163dba28a47d2955cacf8d8908006} for this translation, utilizing \texttt{zone} and \texttt{wall} tags to enforce these constraints and enabling the passage of markup to track segment boundaries and their corresponding identifiers and polarity labels, as shown in Figure~\ref{fig:constraintsmt}.

\begin{figure}[ht!]
    \centering
    \begin{subfigure}[t]{0.48\textwidth}
        \centering
        \includegraphics[width=0.9\textwidth]{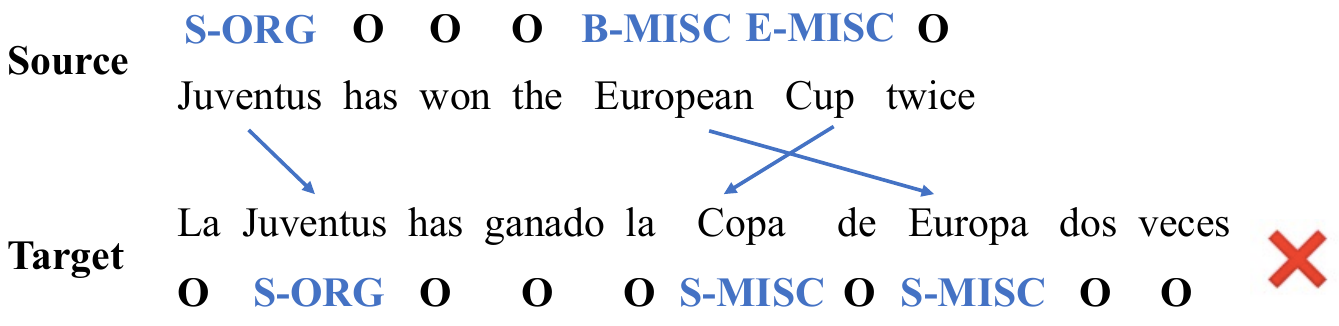}
        \caption{Word-to-word projection.}
    \end{subfigure}%
    ~
    \begin{subfigure}[t]{0.48\textwidth}
        \centering
        \includegraphics[width=0.9\textwidth]{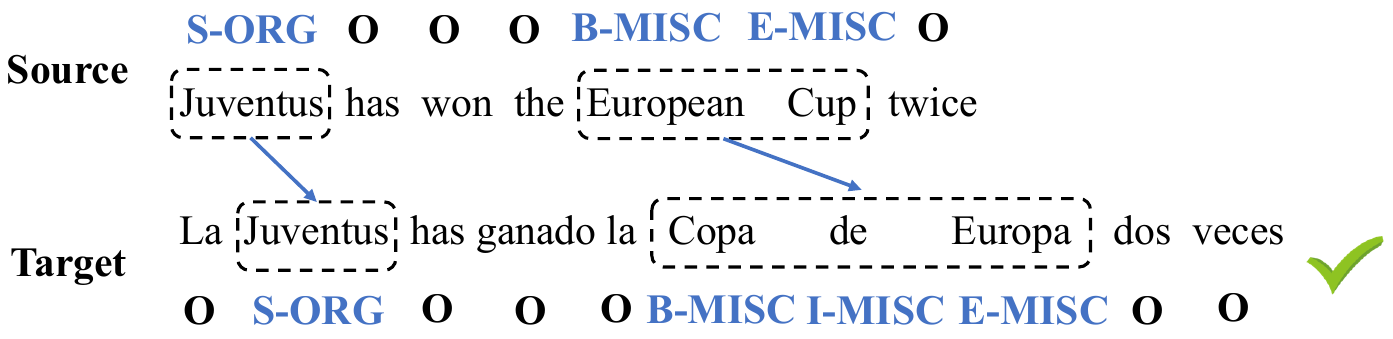}
        \caption{Span-to-span projection.}
    \end{subfigure}
    \caption{Example sentence with different label projection strategies with \texttt{BIOES} as the tagging scheme. The span-to-span strategy proposed by \citet{li2020unsupervised}.}
    \label{fig:label_proj}
\end{figure}

Traditionally, the source and translated target sentences are processed to generate word alignment links with an alignment tool indicating which words in the target language correspond to words in the source sentence. Although aligning word-level labels offers a straightforward method for token-level label projection, it is vulnerable to issues such as word order changes and alignment gaps. To enhance alignment quality, \citet{li2020unsupervised} propose a span-to-span mapping strategy, depicted in Figure~\ref{fig:label_proj}, that transforms word-based alignments into span-based alignments, allowing for more robust label propagation across aligned spans while mitigating potential problems with word order changes.

\begin{figure}[ht!]
    \centering
    \includegraphics[width=0.9\linewidth]{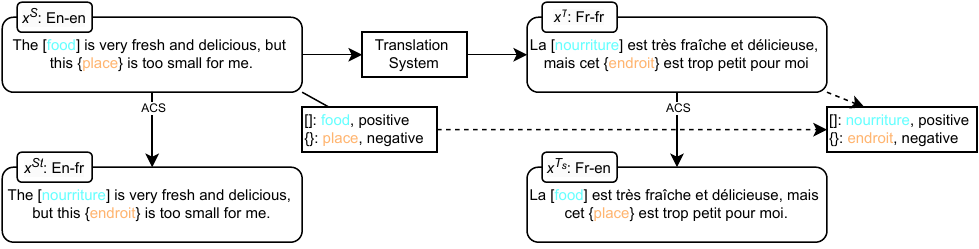}
    \caption{Example of the aspect code-switching technique (lower part) and alignment-free label projection method (upper part) proposed by \citet{zhang-etal-2021-cross}.}
    \label{fig:acs}
\end{figure}

\citet{zhang-etal-2021-cross} introduce an alignment-free label projection method to generate pseudo-labelled data in the target language. Initially, they mark each aspect term in the source sentence with unique symbols before translating it. Once the translation is completed, they extract the spans marked with these symbols and align them with the corresponding aspect terms from the source to recover the aspect boundaries and project the sentiment labels. This approach ensures that even if the order of aspect terms changes during translation, they can still be matched to the appropriate labels; however, they also account for cases where symbols may be lost during translation by filtering out sentences missing these markers. Additionally, the authors propose constructing bilingual sentences by switching aspect terms between the source sentence and its translation. This results in two datasets: one with aspect terms in the target language and the other with aspect terms in the source language. The corresponding label sequences are adjusted accordingly, creating code-switched datasets for further analysis. Figure~\ref{fig:acs} depicts the proposed method.

\citet{nam2024kpc} utilizes Google Translate to create a machine-translated version of the English SemEval-2014 dataset in Korean. Since the tasks addressed are ASC and ACD, no alignment of aspect terms and opinions is required. In contrast, other work~\cite{10031113} leverages the DeepL API to construct a multilingual dataset by translating Polish data into five different languages, employing XLM markups to align aspect terms effectively.

\subsection{Multilingual Pre-trained Language Models}
Multilingual pre-trained language models, such as mBERT~\cite{devlin-etal-2019-bert}, XLM-R~\cite{conneau-etal-2020-unsupervised} and mT5~\cite{xue-etal-2021-mt5}, have become popular for cross-lingual NLP tasks~\cite{pmlr-v119-hu20b}. These models are pre-trained on large multilingual corpora, enabling them to capture cross-lingual syntactic and semantic patterns.

Fine-tuning these models on labelled data in the source language allows \textit{zero-shot} transfer, meaning the model can perform well in the target language without any explicit training data in that language. This approach has been highly effective in many cross-lingual tasks~\cite{wu-dredze-2019-beto, wang2019cross, pires-etal-2019-multilingual}.

\begin{figure}[ht!]
    \centering
    \includegraphics{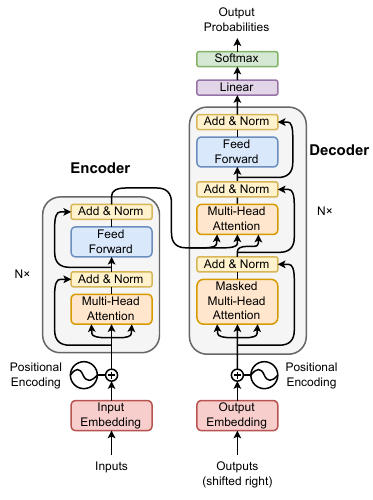}
    \caption{Transformer architecture~\cite{vaswani2023attentionneed}.}
    \label{fig:transformer}
\end{figure}

Most mPLMs are based on the Transformer architecture~\cite{vaswani2023attentionneed}, which utilizes self-attention mechanisms to capture relationships between words, regardless of their position in the sequence. Figure~\ref{fig:transformer} illustrates the architecture, with some models (like mBERT and XLM-R) using only the encoder for classification tasks, while others (like mT5) use both encoder and decoder components for tasks involving sequence generation.

In most cases, models employ subword tokenizers to handle out-of-vocabulary words. Rather than processing entire words, subword tokenizers divide words into smaller units, allowing models to effectively handle rare or unknown words. Another key advantage of mPLMs is that they produce contextualized word embeddings, which capture the meaning of words based on their context. This capability is a significant improvement over static embeddings from models like Word2Vec, FastText, or GloVe, which return the same vector for a word regardless of context – leading to identical vectors for polysemous words like \textit{\quotes{crane}} (whether referring to a bird or a machine) – mPLMs generate different vectors based on the word’s context.

\citet{9585242} explore zero-shot capabilities with mBERT and XLM-R across various source and target language combinations for ACD and ATE tasks, showing that XLM-R outperforms mBERT in most scenarios. Similarly, \citet{doi:10.1080/24751839.2023.2173843} demonstrate that XLM-R performs better in ACD and ASC tasks while also highlighting the benefits of fine-tuning the models on multiple source languages rather than just one.

\citet{nam2024kpc} leverages mBERT and XLM-R to generate pseudo-labelled data for Korean in ACD and ASC tasks, incorporating LaBSE~\cite{feng-etal-2022-language} to filter out low-quality labels. Similarly, different study~\cite{10031113} employs XLM-R and LaBSE for the E2E-ABSA task across various source and target language combinations.

Despite the promise of zero-shot approaches for cross-lingual ABSA, several challenges remain. Language-specific knowledge is vital, as user-generated content often includes abbreviations, slang, or informal language. Aspect and opinion terms can differ significantly across languages. However, zero-shot methods rely heavily on pre-trained models, where low-resource languages are often underrepresented~\cite{conneau-etal-2020-unsupervised, pfeiffer-etal-2020-mad}. This underrepresentation limits the model’s ability to capture and transfer language-specific nuances effectively.
To mitigate these limitations, some studies combine machine translation with mPLMs for E2E-ABSA, incorporating additional techniques such as parameter warm up~\cite{li2020unsupervised}, distillation on target language data~\cite{zhang-etal-2021-cross}, contrastive learning~\cite{lin2023clxabsa}, and handling imbalanced classes~\cite{LinXABSA}.

Other research bypasses fine-tuning mPLMs directly and instead leverages their ability to generate contextualized word embeddings. For instance, \citet{9550786} use these context-aware embeddings from mBERT as input to a CNN, facilitating a more nuanced understanding of word meanings.

\section{Cross-lingual ABSA Tasks}
\label{sec:cl_tasks}
This section discusses ABSA tasks explored in cross-lingual settings in more detail. For the cross-lingual E2E-ABSA task, we include tables with results as consistent datasets, models, and languages used in most of the works allow for a direct comparison of results. In contrast, methods are evaluated across diverse datasets and languages for other tasks, making it impractical to present results in a unified table. Therefore, we present an overview of the cross-lingual ABSA work alongside the solved tasks, cross-lingual transfer used, languages, dataset domains, and models in Table~\ref{tab:tasks-overview}.

\begin{table}[ht!]
\caption{Overview of key cross-lingual ABSA works, organized by year. The \textbf{Approach} column categorizes methods as classical machine learning (ML), neural networks (NN), or Transformers (T). The \textbf{Transfer Method} column specifies the approach for cross-lingual transfer: machine translation (MT), cross-lingual/bilingual word embeddings (WE), or multilingual pre-trained language models (mPLMs). The \textbf{Languages} column lists the languages using ISO 639-1 codes. *This study primarily utilizes English-centric LLMs; however, these models also incorporate data from languages other than English during their initial pre-training phase.}
\begin{adjustbox}{width=\linewidth}
\begin{tabular}{@{}llllllll@{}}
\toprule
\textbf{Source}  & \textbf{Year}                     & \textbf{Approach} & \textbf{Method (Model)}            & \textbf{Transfer Method} & \textbf{Data Domain} & \textbf{ABSA Tasks} & \textbf{Languages}     \\ \midrule
\cite{10.1145/2661829.2662019} & 2014 & ML & LDA-based & MT & Hotels, products & ATE & de, en, es, fr, it, nl, zh \\ 
\cite{klinger-cimiano-2015-instance} & 2015 & ML                & CRF                                &   MT                       & Products             & ATE                 & de, en                 \\
\cite{lambert-2015-aspect}  & 2015           & ML                & SVM                                &   MT                       & Hotels               & ASC                 & en, es                 \\
\cite{zhou2015clopinionminer} & 2015         & ML                & CRF                                &   MT                       & Electronics, cars    & ATE                 & en, zh                 \\
\cite{barnes-etal-2016-exploring}  & 2016    & ML, NN                & SVM, LSTM                          &   MT, WE                       & Hotels               & ASC                 & en, es                 \\
\cite{10.1145/3273931}& 2018 & NN  & LSTM  & WE & Restaurants, laptops & ATE & en, hi \\
\cite{akhtar-etal-2018-solving} & 2018       & NN                & LSTM                               &  WE                        & Restaurants, laptops & ASC                 & en, fr, hi             \\
\cite{wang2018transition}  & 2018            & NN                & RNN &    WE                      & Restaurants          & ATE                 & en, es, fr             \\
\cite{jebbara-cimiano-2019-zero} & 2019      & NN                & CNN                                &   WE                       & Restaurants          & ATE                 & en, es, nl, ru, tr     \\
\cite{li2020unsupervised}   & 2020           & T                & mBERT, XLM-R                       &   MT, mPLMs                       & Restaurants          & E2E-ABSA            & en, es, fr, nl, ru, tr \\
\cite{9585242} & 2021 & T & mBERT, XLM-R & mPLMs  & Restaurants & ACD, ATE & en, es, fr, nl, ru, tr \\
\cite{9550786} & 2021 & NN, T & CNN, mBERT & mPLMs  & Restaurants & E2E-ABSA & en, es, fr, nl \\
\cite{wu2021reinforced} & 2021 & T & BERT & MT & Restaurants, laptops & ASC & en, es, ru, zh \\
\cite{zhang-etal-2021-cross} & 2021            & T                & mBERT, XLM-R                       &   MT, mPLMs                       & Restaurants          & E2E-ABSA            & en, es, fr, nl, ru     \\
\cite{10031113} & 2022 & T & XLM-R, LaBSE & mPLMs & Restaurants, school, medicine, hotel, products & E2E-ABSA & cs, en, es, fr, nl, pl \\
\cite{doi:10.1080/24751839.2023.2173843} & 2023 & T & mBERT, XLM-R & mPLMs & Restaurants & ACD, ASC & en, es, fr, nl, ru \\
\cite{lin2023clxabsa} & 2023                 & T                & mBERT                              &  MT, mPLMs                        & Restaurants          & E2E-ABSA            & en, es, fr, nl, ru     \\ 
\cite{LinXABSA}  & 2024                & T                & mBERT, XLM-R                              &   MT, mPLMs                       & Restaurants          & E2E-ABSA            & en, es, fr, nl, ru     \\ 
\cite{nam2024kpc} & 2024 & T & mBERT, XLM-R, LaBSE & mPLMs & Restaurants & ACS, ASC & en, kr 
\\
\cite{wu2024evaluatingzeroshotmultilingualaspectbased} & 2024                & T                & LLMs                              &   mPLMs*                       & Restaurants          & E2E-ABSA            & en, es, fr, nl, ru \\
\bottomrule
\end{tabular}
\end{adjustbox}
\label{tab:tasks-overview}
\end{table}

\subsection{Aspect Term Extraction}
Aspect term extraction is a core task in ABSA, focusing on identifying specific aspect expressions within a given text that users express opinions about. For instance, in the sentence \textit{\quotes{The tea is great, but the ambience is disappointing}}, the aspect terms \textit{\quotes{tea}} and \textit{\quotes{ambience}} would be extracted as they represent the subjects of the user's opinions. Aspect term extraction is often framed as a token-level classification problem, where each token in the sentence is classified, as the aspect terms are usually individual words or short phrases.

\citet{zhou2015clopinionminer} employ machine translation and co-training of two labelling models based on CRF to transfer knowledge between English and Chinese. In this approach, each model annotates unlabelled datasets, selects the most confident examples, and adds them to the labelled dataset to continue training. Similarly, \citet{klinger-cimiano-2015-instance} use CRF combined with machine translation to address ATE between English and German.  \citet{10.1145/3273931} employ LSTM with bilingual embeddings to tackle ATE in English and Hindi. \citet{10.1145/2661829.2662019} propose an unsupervised LDA-based model to align aspects bilingually by assuming that reviews in different languages share the same topic distribution, using a bilingual dictionary to translate words and select the most suitable translation, enabling cross-lingual topic alignment without the need for parallel corpora.

\citet{wang2018transition} examine English, French, and Spanish and introduce a transition-based mechanism that processes one word at a time, forming a series of configurations that represent the status of the entire sentence. Each configuration is represented as a continuous feature vector, which is aligned from different languages into a shared space through an adversarial network. The transition-based adversarial network consists of a generator that produces language-invariant configuration features and a discriminator that distinguishes configurations between source and target languages. Bilingual word embeddings are the input to the network.

\citet{jebbara-cimiano-2019-zero} investigate multilingual word embeddings for English, Spanish, Dutch, Russian, and Turkish using CNNs. They experiment with various combinations of source and target languages and find that cross-lingual learning from multiple source languages enhances performance. Similarly, \citet{9585242} explore mPLMs, specifically mBERT and XLM-R, in zero-shot settings for English, Spanish, French, Dutch, Russian, and Turkish. Their findings indicate that XLM-R typically outperforms mBERT across different source and target language combinations.

\subsection{Aspect Sentiment Classification}
Aspect sentiment classification, often referred to as aspect-based or aspect-level sentiment classification, focuses on determining the sentiment polarity associated with a specific aspect within a sentence. The aspect can be either an aspect term or an aspect category, leading to two distinct ASC tasks: aspect term-based sentiment classification and aspect category-based sentiment classification. While there are minor distinctions between the two (such as utilizing the position information of a given aspect term from the sentence), the primary research question remains consistent across both: how to effectively leverage the relationship between the aspect (whether a term or category) and the surrounding sentence context to classify sentiment accurately. It is possible to address both subtasks simultaneously, employing the same model to handle them seamlessly. 

\citet{lambert-2015-aspect} uses constraint SMT (see Section~\ref{sec:mt} for more details) alongside SVM to perform aspect term-based sentiment classification in Spanish and English, adding the tokens (unigrams) of aspect terms as additional features to the classifier.

\citet{barnes-etal-2016-exploring} compare several methods for aspect term-based sentiment classification in Spanish and English with the SVM classifier. In zero-shot settings, they train a translation matrix to map two monolingual vector spaces into one. Additionally, they train bilingual word embeddings and employ stacked denoising autoencoders to encode the parallel sentences into a common latent space. Finally, they experiment with machine translation. They find that the machine translation approach performs the best, likely due to its use of high-quality, in-domain data. Additionally, they evaluate LSTM but demonstrate that SVM is more efficient in their setup.

\citet{akhtar-etal-2018-solving} utilize bilingual word embeddings and various hand-crafted features as the input to the bidirectional LSTM classifier for aspect term-based sentiment classification in English, Hindi, and French. They employ concatenation to fuse the aspect information with the sentence context. \citet{wu2021reinforced} use machine translation to build a bilingual lexicon to explore aspect category-based sentiment classification in Chinese, Russian, English, and Spanish. They introduce a reinforced transformer based on pre-trained English BERT~\cite{devlin-etal-2019-bert} with cross-lingual distillation to model aspect representations and sentence representations.

\citet{doi:10.1080/24751839.2023.2173843} explore mPLMs, specifically mBERT and XLM-R, to tackle aspect category-based sentiment classification across English, Spanish, French, Dutch, and Russian with joint training on data from multiple languages. Their models output a one-hot vector with a size equal to the number of categories, where each value represents the sentiment polarity of a category corresponding to its position in the vector.

\citet{nam2024kpc} introduces a framework that fine-tunes multilingual BERT-base models on the Korean translation of the English SemEval-2014 dataset for the aspect category-based sentiment classification and aspect category detection. The model then generates pseudo-labels for Korean data, which are used for training the final model. This final model is trained either on a combination of the translated data and the pseudo-labelled data or solely on the pseudo-labelled data. To enhance the quality of the pseudo-labels, dual filtering techniques are employed, including LaBSE-based filtering and confidence score filtering. LaBSE, originally developed as a dual-encoder for machine translation, proves equally effective for monolingual tasks such as semantic textual similarity (STS) and generating high-quality training data. This cost-efficient approach demonstrates how resource gaps can be bridged, providing an effective solution for ABSA in low-resource languages.

\subsection{Aspect Category Detection}
Aspect category detection involves identifying the relevant aspect categories discussed in a given sentence, which typically belong to a predefined set of categories that are often domain-specific. For instance, given the sentence \textit{\quotes{The tea is great, but the ambience is disappointing}}, an ACD method should accurately predict the categories of \textit{\quotes{drinks}} and \textit{\quotes{ambience}}.

Compared to the ATE task, ACD offers advantages from two main perspectives. First, while ATE focuses on predicting individual aspect terms, predicted categories of ACD represent an aggregated outcome, providing a more concise summary of the opinion targets. Second, ACD can identify opinion targets even when they are not explicitly mentioned. For example, in the sentence \textit{\quotes{The wait was long, and the lunch was overpriced}}, ACD can recognize two aspect categories: \textit{\quotes{service}} and \textit{\quotes{price}}, whereas ATE would not be applicable in this scenario.

The ACD task is often framed as a multi-label sequence-level classification problem, where each category is a label. \citet{9585242} employ mBERT and XLM-R in English, Spanish, French, Dutch, Russian, and Turkish, demonstrating the superior results with XLM-R. Similarly, \citet{doi:10.1080/24751839.2023.2173843} explore the same models and languages except for Turkish, training the models on different combinations of source languages. Additionally, \citet{nam2024kpc} proposes a pseudo-labelling framework for low-resource languages, which leverages multilingual BERT fine-tuned on translated data to generate high-quality pseudo-labels, as detailed in the ASC task.

\subsection{End-to-End ABSA}
The end-to-end ABSA task aims to simultaneously identify aspect terms and their corresponding sentiment polarities within a given sentence. This task is unique in cross-lingual settings as it is the only compound task explored extensively. The focus on end-to-end ABSA has grown significantly with the advent of multilingual pre-trained language models.

The E2E-ABSA task is often tackled using a unified tagging scheme. The first part of the scheme, such as BIO or BIOES, marks the boundaries of aspect terms, while the second part denotes the sentiment polarity. For example, the BIOES scheme uses \texttt{B} for the beginning, \texttt{I} for inside, \texttt{O} for outside, \texttt{E} for the end, and \texttt{S} for singleton, denoting the boundaries of aspect terms, while the sentiment polarity is indicated as either positive (\texttt{POS}), negative (\texttt{NEG}), or neutral (\texttt{NEU}). For instance, \texttt{B-NEU} refers to the beginning of an aspect term with a neutral sentiment polarity. This tagging structure allows end-to-end ABSA to be modelled as a token classification problem using a sequence tagger. Table~\ref{tab:e2etagging} provides an example of the BIOES and BIO tagging formats.

\begin{table}[ht!]
\centering
\caption{Example of the BIOES and BIO tagging scheme for the end-to-end ABSA task.}
\begin{adjustbox}{width=0.9\linewidth}
\begin{tabular}{@{}cccccccccccccc@{}}
\toprule
               & The & steak & with  & potatoes & was & delicious & , & but & the & service & was & horrible & . \\ \midrule
\textbf{BIOES} & O   & B-POS & I-POS & E-POS    & O   & O         & O & O   & O   & S-NEG   & O   & O        & O \\
\textbf{BIO}   & O   & B-POS & I-POS & I-POS    & O   & O         & O & O   & O   & B-NEG   & O   & O        & O \\ \bottomrule
\end{tabular}
\end{adjustbox}

\label{tab:e2etagging}
\end{table}

\citet{9550786} use part-of-speech (POS) tagging to extract tokens from the text, pre-process them by tasks like removing punctuation, and then map the tokens between languages using bilingual dictionaries. They utilize mBERT to generate embeddings and feed both the POS tags and embeddings into an attention-based CNN for further processing.

\citet{10031113} present TrAsp, a multilingual and language-agnostic approach to ABSA that employs Transformer-based methods. TrAsp utilizes a Transformer-based embedding model with two linear layers and one dropout layer, achieving superior performance compared to existing techniques. The method is evaluated on the MultiAspectEmo dataset (covering Polish and its machine-translated versions in English, Czech, Spanish, French, and Dutch) and the SemEval-2016 dataset. The results demonstrate that multilingual pre-trained models like XLM-R significantly outperform language-agnostic models such as LaBSE. The authors also compare the cross-lingual capabilities of LaBSE and XLM-R, highlighting a substantial drop in cross-lingual performance relative to monolingual settings, with XLM-R consistently outperforming LaBSE. The study underscores the advantages of multilingual training for cross-lingual knowledge transfer, achieving higher F1 scores, and investigates the potential of compressed models for resource-efficient ABSA. The multilingual capabilities of pre-trained models are shown to be pivotal in enabling robust cross-lingual performance.

\citet{wu2024evaluatingzeroshotmultilingualaspectbased} evaluate zero-shot multilingual ABSA with LLMs for the E2E-ABSA task across English, French, Spanish, Dutch, and Russian using the SemEval-2016 dataset. They explore various prompting strategies, including vanilla zero-shot, chain-of-thought, self-improvement, self-debate, and self-consistency. Their findings reveal that while LLMs show potential for multilingual ABSA, they generally underperform compared to fine-tuned, task-specific models. Performance varies significantly by language, with higher-resource languages like English achieving better results than low-resource languages and simpler zero-shot prompts often outperforming more complex strategies in these languages. However, LLMs face challenges in extracting fine-grained structured information and struggle with reasoning tasks, especially in multi-turn dialogue contexts. Closed-source models generally outperform open-source ones, and few-shot learning improves performance, although context length remains a constraint. Furthermore, cross-lingual evaluation, where models fine-tuned on English data are tested on other languages, shows significant performance drops in morphologically different languages such as Russian. These findings highlight the potential of LLMs for multilingual ABSA while emphasizing the need for further research to address their dependencies on task, language, and configuration, as well as to optimize their application for cross-lingual tasks.

\citet{li2020unsupervised} explore their span-to-span translation mapping strategy, described in Section~\ref{sec:mt}, by training XLM-R and mBERT in three different settings: using only the source language data (\textsc{Zero-shot}), using only translated target language data (\textsc{Translation-TA}), and using a combination of both (\textsc{Bilingual-TA}). While their approach proved effective for semantic role labelling, they find that for cross-lingual ABSA, fine-tuning models on the original source language data consistently outperformed training on translated data or combinations of both, which they attribute to the model's lower tolerance to translation errors. To address this, they proposed a model parameter warm-up on subsets of translated data from multiple target languages (\textsc{MTL-WS)} outperforming the multilingual training baseline (\textsc{MTL-TA}) with models fine-tuned on a combination of data translated to multiple target languages.

\citet{zhang-etal-2021-cross} demonstrate the efficacy of their alignment-free translation mechanism discussed in Section~\ref{sec:mt} in several configurations: training on pseudo-labelled target language data alone (\textsc{Translation-AF}), combining translated and source data (\textsc{Bilingual-AF}), and training on both original and translated data alongside aspect-code-switched data, where aspect terms are swapped between languages (\textsc{ACS}). In this setting, they initialize a student model using parameters from a model pre-trained on the translated target language data. The student model is then fine-tuned to match the soft probability distribution predicted by the teacher model for each token, effectively learning from the teacher's output rather than the original hard labels. This approach helps the student model generalize better to the target language by approximating the teacher's predictions, allowing it to capture more nuanced sentiment information. They experimented with single-teacher distillation, using a teacher trained on the \textsc{ACS} dataset (\textsc{ACS-Distil-S}), and multi-teacher distillation, where each teacher is trained on different datasets, such as translated and source data or translated and aspect-code switched data (\textsc{ACS-Distil-M}). Furthermore, they experiment with multilingual settings, utilizing multilingual data with alignment-free label projection (\textsc{MTL-AF}) and aspect-code-switched data (\textsc{MTL-ACS}), and distillation on multilingual unlabelled data (\textsc{MTL-ACS-D}).

\citet{lin2023clxabsa, LinXABSA} further utilize the \textsc{ACS} data and add some additional improvements. They employ contrastive learning to minimize the distance between samples with the same label in different semantic spaces~\cite{lin2023clxabsa}. Specifically, they use token-level contrastive learning to align token embeddings with similar labels (\textsc{CL-XABSA (TL)}) and sentiment-level contrastive learning for samples at the sentiment level (\textsc{CL-XABSA (SL)}). Additionally, they address class imbalances with dynamic weighted loss, which shifts the model's attention towards underperforming sentiment categories, and introduce anti-decoupling techniques to enhance semantic information utilization (\textsc{Equi-ABSA})~\cite{LinXABSA}.

Table~\ref{tab:e2eres} presents the cross-lingual results with mBERT and XLM-R on the SemEval-2016 restaurant dataset, with English as the source language and other languages as targets. The best results are typically achieved with \textsc{ACS-Distil-M}~\cite{zhang-etal-2021-cross}, which leverages aspect-code switching and distillation on unlabelled target language data, and \textsc{Equi-ABSA}~\cite{LinXABSA}, which addresses class imbalances. Table~\ref{tab:e2eresmulti} shows the multilingual result where \textsc{MTL-ACS-D}~\cite{zhang-etal-2021-cross} achieves the best results by combining aspect code-switching and multilingual distillation on unlabelled target language data.

\begin{table}[ht!]
\centering
\caption{Cross-lingual end-to-end ABSA results of different methods with English as the source language and other languages as target ones compared to supervised settings (\textsc{Supervised}). The best results for each model and language are in \textbf{bold}. *The average results exclude Turkish.}
\begin{adjustbox}{width=\linewidth}
\begin{tabular}{@{}llcccccccccccc@{}}
\toprule
\multirow{2}{*}{\textbf{Source}} & \multirow{2}{*}{\textbf{Method}} & \multicolumn{6}{c}{\textbf{mBERT}}                                                         & \multicolumn{6}{c}{\textbf{XLM-R}}                                                         \\  \cmidrule(lr){3-8} \cmidrule(lr){9-14}
                                 &                                  & Es             & Fr             & Nl             & Ru             & Tr    & Avg*           & Es             & Fr             & Nl             & Ru             & Tr    & Avg*           \\ \midrule
\cite{zhang-etal-2021-cross}                                 & \textsc{Supervised}                       & 67.88          & 61.80          & 56.80          & 58.87          & --     & 61.34          & 71.93          & 67.44          & 64.28          & 64.93          & --     & 67.15          \\ \cdashlinelr{1-14}
\multirow{3}{*}{\cite{li2020unsupervised}}               & \textsc{Zero-shot}                        & 57.32          & 45.60          & 42.68          & 36.01          & \textbf{26.58} & 45.40          & 67.10          & 56.43          & 59.03          & 56.80          & \textbf{46.21} & 59.84          \\
                                 & \textsc{Translation-TA}                   & 50.74          & 40.76          & 47.13          & 41.67          & 22.04 & 45.08          & 58.10          & 47.00          & 56.19          & 50.34          & 40.24 & 52.91          \\
                                 & \textsc{Bilingual-TA}                     & 51.23          & 41.00          & 49.72          & 43.67          & 22.64 & 46.41          & 61.87          & 49.34          & 58.64          & 52.89          & 41.44 & 55.69          \\ \cdashlinelr{1-14}
\multirow{5}{*}{\cite{zhang-etal-2021-cross}}               & \textsc{Translation-AF}                   & 59.74          & 48.03          & 49.73          & 50.17          & --     & 51.92          & 66.61          & 57.07          & 61.26          & 59.55          & --     & 61.12          \\
                                 & \textsc{Bilingual-AF}                     & 60.23          & 48.05          & 49.83          & 51.24          & --     & 52.34          & 68.04          & 57.91          & 60.80          & 60.81          & --     & 61.89          \\
                                 & \textsc{ACS}                              & 59.99          & 49.65          & 51.19          & 52.09          & --     & 53.23          & 67.32          & 59.39          & 62.83          & 60.81          & --     & 62.59          \\
                                 & \textsc{ACS-Distil-S}                    & 62.04          & 52.23          & 52.72          & 53.00          & --     & 55.00          & 68.93          & \textbf{61.00} & 62.89          & 60.97          & --     & 63.45          \\
                                 & \textsc{ACS-Distil-M}                    & 62.91          & \textbf{52.25} & \textbf{53.40} & \textbf{54.58} & --     & \textbf{55.79} & 69.24          & 59.90          & \textbf{63.74} & 62.02          & --     & \textbf{63.73} \\ \cdashlinelr{1-14}
\multirow{2}{*}{\cite{lin2023clxabsa}}               &\textsc{CL-XABSA(TL)}                     & 60.64          & 48.53          & 50.96          & 50.77          & --     & 52.73          & --              & --              & --              & --              & --     &                \\
                                 & \textsc{CL-XABSA(SL)}                     & 61.62          & 49.50          & 50.64          & 50.65          & --     & 53.10          & --              & --              & --              & --              & --    &                \\ \cdashlinelr{1-14}
\cite{LinXABSA}                                & \textsc{Equi-XABSA}                       & \textbf{63.08} & 50.08          & 51.85          & 52.59          & --     & 54.40          & \textbf{69.56} & 60.68          & 61.31          & \textbf{62.34} & --     & 63.47          \\ \bottomrule
\end{tabular}
\end{adjustbox}
\label{tab:e2eres}
\end{table}

\begin{table}[ht!]
\caption{Multilingual end-to-end ABSA results of different methods. The best results for each model and target language are in \textbf{bold}. *The average results exclude Turkish.}
\begin{adjustbox}{width=\linewidth}
\begin{tabular}{@{}llcccccccccccc@{}}
\toprule
\multirow{2}{*}{\textbf{Source}} & \multirow{2}{*}{\textbf{Method}} & \multicolumn{6}{c}{\textbf{mBERT}}            & \multicolumn{6}{c}{\textbf{XLM-R}}            \\  \cmidrule(lr){3-8} \cmidrule(lr){9-14}
                                 &                                  & Es    & Fr    & Nl    & Ru    & Tr    & Avg*  & Es    & Fr    & Nl    & Ru    & Tr    & Avg*  \\ \midrule
\multirow{2}{*}{\cite{li2020unsupervised}}               & \textsc{MTL-TA}                           & 54.14 & 40.72 & 49.06 & 43.89 & 25.77 & 46.95 & 63.56 & 52.80 & 60.37 & 55.67 & 43.04 & 58.10 \\
                                 & \textsc{MTL-WS}                           & 58.18 & 46.93 & 49.87 & 44.88 & \textbf{29.78} & 49.96 & 68.60 & 57.96 & 61.24 & 59.74 & \textbf{45.58} & 61.89 \\ \cdashlinelr{1-14}
\multirow{3}{*}{\cite{zhang-etal-2021-cross}}               & \textsc{MTL-AF}                           & 59.31 & 50.00 & 53.16 & 50.04 & --     & 53.13 & 68.20 & 59.68 & 63.33 & 61.02 & --     & 63.06 \\
                                 & \textsc{MTL-ACS}                          & 59.59 & 50.74 & 53.33 & 51.61 & --     & 53.82 & 68.88 & 59.19 & 63.06 & 61.92 & --     & 63.26 \\
                                 & \textsc{MTL-ACS-D}                        & \textbf{62.05} & \textbf{53.56} & \textbf{53.56}& \textbf{53.87} & --     & \textbf{55.76} & \textbf{70.38} & \textbf{62.17} & \textbf{65.98} & \textbf{62.79} & --     & \textbf{65.33} \\ \bottomrule 
\end{tabular}
\end{adjustbox}
\label{tab:e2eresmulti}
\end{table}

\section{Related Research}
\label{sec:related}

In this section, we describe some of the related research for ABSA, focusing on innovative methods in recent years, languages other than English, multilingual ABSA, cross-lingual sentiment analysis, and LLMs for ABSA. Additionally, we shortly summarize advancements in multilingual and cross-lingual sentiment analysis. Finally, we discuss the possibilities of utilizing large language models for ABSA.

\subsection{Monolingual ABSA}
\citet{Ma_Peng_Cambria_2018} address the TASD task with a neural architecture combining a bidirectional LSTM sequence encoder and a hierarchical attention mechanism. Their model employs target-level attention to highlight sentiment-relevant parts of aspect terms and sentence-level attention to identify aspect evidence across the sentence. Additionally, they propose Sentic LSTM, an enhanced LSTM cell incorporating affective commonsense knowledge. \citet{LIANG2022107643} introduce Sentic GCN, a graph convolutional network that augments sentence dependency graphs with affective knowledge from SenticNet. This model captures both contextual and affective relationships between words and aspects by integrating affective dependencies and focusing on aspect-specific sentiment. Both approaches can be extended to cross-lingual settings by incorporating cross-lingual word embeddings or machine translation.

\citet{d2022knowmis} present KnowMIS-ABSA, a framework emphasizing distinctions among key concepts such as sentiment, emotion, affect, and opinion. Unlike traditional methods, it advocates for tailored metrics and tools to measure these dimensions separately, ensuring a more nuanced understanding of user opinions. A qualitative case study demonstrates its advantages, particularly in marketing, where differentiating sentiments from affective reactions can enhance strategies.

Recent English ABSA research predominantly adopts the sequence-to-sequence modelling paradigm due to its flexibility in handling compound tasks involving three or four sentiment elements. Annotation- and extraction-style modelling paradigms by \citet{zhang-etal-2021-aspect-sentiment} showcase the viability of a generative framework for ABSA. Later work~\cite{zhang-etal-2021-aspect-sentiment} introduces a novel approach to the ASQP task, predicting sentiment quads in natural language format, facilitating more intuitive sentiment analysis results. \citet{gao-etal-2022-lego} propose a comprehensive framework that addresses multiple ABSA tasks by integrating element prompts, while \citet{mao-etal-2022-seq2path} generate tuples as tree paths and filter valid ones. Most of these methods generate sentiment elements in a fixed order, which may not be optimal. \citet{hu-etal-2022-improving-aspect} investigate the impact of sentiment element ordering on ABSA outcomes, optimizing the sequence generation process. Building on this, \citet{gou-etal-2023-mvp} utilize multiple-ordered permutations to enhance sentiment element prediction, merging results from different orders at the cost of extended training times.  \citet{xianlong-etal-2023-tagging} combine sequence-to-sequence and token classification paradigms. The encoder labels aspect and opinion terms, merging results with the model's text output. All these methods can be adapted to cross-lingual transfer using multilingual Transformer-based models, optionally enhanced by machine translation.

For languages beyond English, Czech ABSA research has gained traction. Earlier studies~\cite{steinberger-etal-2014-aspect,hercig2016unsupervised,tamchyna2015czech} focus on simple tasks with two sentiment elements, combining carefully selected features with classifiers like CRF and logistic regression. These methods are adaptable to cross-lingual settings via machine translation. Later study~\cite{SLON-Lenc2016Neural} explores convolutional and recurrent neural networks for ACS and ACD tasks, adaptable using machine translation or cross-lingual embeddings. Recent studies leverage Transformer-based models. For instance, \citet{smid-priban-2023-prompt} introduce prompt-based methods for Czech ABSA, demonstrating the effectiveness of multilingual sequence-to-sequence and Czech-specific BERT-base models models and in-domain pre-training in few-shot settings. Similarly, another work~\cite{priban-prazak-2023-improving} enhances ABSA with semantic role labelling in a multitask learning framework, improving ACD and ASC tasks in both Czech and English. \citet{smid-etal-2024-czech} deliver state-of-the-art Czech ABSA results using Transformer-based methods with both multilingual and Czech-specific models. The modern approaches are adaptable to cross-lingual settings through multilingual models and machine translation.

\subsubsection{Relationship to Cross-lingual ABSA}
The approaches discussed can be adapted to cross-lingual ABSA using cross-lingual transfer techniques from Section~\ref{sec:cl_transfer}, such as cross-lingual word embeddings, machine translation, multilingual pre-trained models, or a combination of them. A key distinction between monolingual and cross-lingual research lies in their focus: monolingual research aims to improve ABSA task performance, while cross-lingual research prioritizes enhancing transfer capabilities across languages. Combining the strengths of both fields could significantly improve cross-lingual ABSA. For instance, leveraging multilingual models with machine translation and target-language distillation alongside high-performing monolingual approaches like tagging-assisted generation could yield substantial advancements.

\subsection{Multilingual and Few-Shot Cross-Domain ABSA}
\citet{GARCIAPABLOS2018127} present W2VLDA, an almost unsupervised system for ABSA that requires only minimal supervision, addressing the limitations of resource-intensive supervised approaches. W2VLDA combines latent Dirichlet allocation (LDA)~\cite{10.5555/944919.944937}, continuous word embeddings, and a maximum entropy classifier to classify domain-specific aspects and sentiment polarity. It operates on unlabelled textual corpora, requiring only a single seed word per domain aspect and two polarity seed words (positive and negative). The system generates weighted lists of aspect terms, opinion words, and classified sentences, aligning these outputs with predefined domain aspects. Evaluated on the SemEval-2016 5 dataset across multiple domains (restaurants, hotels, electronic devices) and languages (English, Spanish, French, Dutch), W2VLDA achieves competitive results. Its reliance on minimal annotated data and automatic separation of aspect terms and opinion words makes it adaptable to diverse domains and languages, offering an alternative to cross-lingual methods for ABSA.

\citet{9852228} introduce LGCF, a multilingual framework for ABSA that effectively captures local and global context features to predict sentiment polarity. LGCF uses BERT-based embeddings with distinct layers to model local and global contexts. Local context modelling is enhanced by multi-headed self-attention (MHSA), dynamic masking (CDM), and dynamic weighting (CDW). A combination of bidirectional GRU~\cite{69e088c8129341ac89810907fe6b1bfe} and CNN is employed for global context modelling, enabling robust feature learning for aspect-sentiment correlations. Validated on Chinese and English datasets, LGCF achieves state-of-the-art performance, demonstrating the advantages of combined local and global context modelling.

\citet{rana2024exploring} propose a hybrid model for multilingual sentiment analysis that combines BERT with a lexicon-based approach to improve polarity estimation for reviews and tweets. The model incorporates multilingual conversion to translate non-English content into English, effectively handling multilingual text. Processing involves stages like preprocessing, aspect extraction, aspect engineering, and polarity calculation, leveraging resources such as SentiWordNet and part-of-speech taggers. While initially focused on multilingual scenarios, this approach can be adapted for cross-lingual settings by training in one language and testing in another, as machine translation is already an integral part of the workflow. 

Another study~\cite{info15110664} explores cross-domain few-shot classification for ABSA, where models trained in one domain (e.g. hotel reviews) are evaluated in another (e.g. restaurant reviews) with minimal labelled target data. While primarily focused on domain adaptation within a single language, the methods bear similarities to cross-lingual ABSA, where knowledge is transferred across languages instead of domains. Using the Polish AspectEmo dataset, the study compares gradient-based learning, zero-shot approaches, and few-shot methods like ProtoNet~\cite{10.5555/3294996.3295163} and NNShot~\cite{yang-katiyar-2020-simple}. Few-shot methods outperform other approaches, with NNShot offering a slight edge. These methods leverage embeddings to predict labels in the target domain by comparing support and query set similarities. The findings suggest the potential for few-shot learning in cross-lingual ABSA, where models could benefit from limited examples in the target language.

\subsubsection{Relationship to Cross-lingual ABSA}
Multilingual ABSA research often overlaps with cross-lingual ABSA techniques, such as machine translation and multilingual pre-trained models, making adaptation to cross-lingual settings straightforward. However, key differences must be addressed. Multilingual approaches typically include the target language in the training data, allowing models to learn language-specific nuances during training. In contrast, zero-shot cross-lingual transfer relies solely on source language data, requiring the model to generalize across languages.

Using multiple source languages can enhance performance by providing diverse linguistic patterns but may also hinder results if poorly chosen~\cite{lim-etal-2024-analysis}. Strategies like machine translation to convert source language data into the target language or carefully selecting source languages can mitigate these challenges. Machine translation-based approaches can augment original datasets, either supplementing or replacing them, to provide better language-specific adaptation. Few-shot cross-lingual approaches could also leverage minimal labelled data in the target language to fine-tune models trained on source language data, combining the strengths of multilingual and cross-lingual methods.

\subsection{Multilingual and Cross-lingual Sentiment Analysis}
This section provides an overview of multilingual and cross-lingual sentiment analysis approaches, focusing on document-level sentiment classification. Unlike aspect-based sentiment analysis, which requires fine-grained analysis at the aspect level, document-level sentiment analysis assigns sentiment to an entire document, making it a comparatively simpler task. Additionally, it is a more examined task than cross-lingual ABSA.

Early approaches to cross-lingual sentiment analysis frequently leverage machine translation~\cite{wan-2008-using, wan-2009-co, balahur-turchi-2012-multilingual, balahur2014comparative, eriguchi2018zeroshot}, often translating English training data into other languages to train sentiment models. Other work~\cite{ghorbel-2012-french} utilizes translation for specific resources, such as lexical sentiment dictionaries like SentiWordNet, to enrich datasets and improve sentiment analysis in target languages.

Another significant line of research focuses on developing cross-lingual word embeddings specifically tailored for sentiment analysis~\cite{zhou-etal-2015-learning-bilingual}. These embeddings are employed with models like SVM and neural networks such as CNN and LSTM~\cite{priban-tsd-2022, PRIBAN2024123247}. Researchers also introduce approaches to align word embeddings across languages using small bilingual dictionaries or sentiment information derived from the source language, allowing for efficient transfer of sentiment knowledge to target languages~\cite{barnes-etal-2018-bilingual}.

In recent years, Transformer-based architectures have dominated cross-lingual sentiment analysis. Multilingual versions of BERT and related models have proven particularly effective~\cite{PRIBAN2024123247, priban-steinberger-2021-multilingual, thakkar2022multi,wang-banko-2021-practical}. Studies have further enhanced these models with data augmentation techniques, such as augmenting datasets through machine translation~\cite{barriere-balahur-2020-improving}. Some studies~\cite{10.1145/3651781.3651819, 10.1145/3600229, 10.1145/3552515, nazir-2025-multi} further investigate hybrid strategies with multilingual pre-trained models and machine translation, such as mixing words between languages to enhance training, and focus on low-resource languages that inherently include code-mixed data, such as anglicisms commonly found in social media texts.

Further advancements include the exploration of zero-shot and few-shot learning strategies for cross-lingual classification tasks, enabling sentiment analysis in languages unseen during pre-training~\cite{winata-etal-2022-cross}. Researchers have also evaluated the effectiveness of linear transformations for zero-shot sentiment classification and subjectivity classification tasks, providing alternatives to Transformer-based methods~\cite{priban-tsd-2022, PRIBAN2024123247}.

\subsubsection{Relationship to Cross-lingual ABSA}
Compared to multilingual and cross-lingual sentiment analysis, cross-lingual ABSA employs many of the novel and best-performing approaches for cross-lingual transfer, such as modern pre-trained multilingual language models, machine translation, mixing words between languages, and cross-lingual embeddings. However, there is limited scope for further direct adoption of techniques from cross-lingual sentiment analysis into cross-lingual ABSA, as many of these methods have already been utilized.

Cross-lingual ABSA introduces additional complexities that are not as prominent in document-level sentiment analysis. One such challenge is the precise alignment of aspect and opinion term labels, which is crucial for accurate aspect-based sentiment predictions. This alignment task is inherently difficult due to the nuanced relationship between aspect terms and their corresponding sentiment expressions, especially when crossing linguistic boundaries. Furthermore, cross-lingual ABSA often requires greater attention to language-specific features, such as idiomatic expressions, colloquialisms, or slang, which may vary significantly across languages. These nuances demand a more tailored approach and further complicate the transfer of knowledge from one language to another.

\subsection{Large Language Models for ABSA}
Large language models refer to Transformer-based architectures with parameter counts exceeding 1 billion or even 10 billion, depending on the source~\cite{minaee2024largelanguagemodelssurvey, zhang-etal-2024-sentiment}. These generative models take text as input and produce text as output, with most of them based on the decoder stack of the Transformer. Examples of LLMs include PaLM~\cite{10.5555/3648699.3648939}, GPT-4~\cite{openai2024gpt4technicalreport}, and LLaMA~\cite{touvron2023llamaopenefficientfoundation, touvron2023llama2openfoundation, dubey2024llama3herdmodels}. LLMs are pre-trained on massive text corpora, and many are further tuned to follow instructions. This tuning has shifted the paradigm from traditional fine-tuning to \textit{prompting}, where tasks are defined directly in the input prompt without requiring labelled training data. Prompting can be zero-shot (task definition only) or few-shot (task definition with examples, also called in-context prompting), enabling effective solutions in scenarios with limited labelled data. This capability is especially beneficial in cross-lingual settings, where annotated data for specific tasks and domains might not exist in all target languages.

While LLMs have demonstrated state-of-the-art performance across many NLP tasks, including sentiment analysis~\cite{zhang-etal-2024-sentiment,PRIBAN2024123247,zhong2023chatgptunderstandtoocomparative}, they often underperform compared to smaller, fine-tuned models for ABSA tasks~\cite{gou-etal-2023-mvp,zhang-etal-2024-sentiment}. However, fine-tuning LLMs specifically for ABSA has achieved impressive results, including end-to-end ABSA in English~\cite{simmering2023largelanguagemodelsaspectbased} and complex triplet and quadruplet tasks~\cite{smid-etal-2024-llama}. Early study~\cite{wu2024evaluatingzeroshotmultilingualaspectbased} focusing on the multilingual performance of LLMs has broadly explored the cross-lingual abilities of fine-tuned LLMs with varied results.

The performance of LLMs in ABSA tasks can be affected by inconsistencies in dataset annotations, particularly for opinion terms. Some datasets label modifiers like \textit{\quotes{very}}~\cite{klinger-cimiano-2014-usage}, while others omit them~\cite{fan-etal-2019-target, aste, cai-etal-2021-aspect}, leading to inconsistencies that affect evaluation metrics. Fine-tuned models can adapt to such patterns if the annotations are consistent. However, LLMs operating in zero-shot or few-shot settings may produce outputs that do not align precisely with the evaluation criteria. To address these challenges, some better evaluation techniques of LLMs for ABSA might be needed. Additionally, techniques such as \textit{chain-of-thought} prompting~\cite{10.5555/3600270.3602070} and reasoning-based multi-prompt methods~\cite{fei-etal-2023-reasoning} may improve LLM performance in complex ABSA tasks.

Recent studies illustrate the nuanced performance of LLMs in ABSA. \citet{filip2024finetuningmultilinguallanguagemodels} use several LLMs with smaller BERT-like models for the ASC task in Czech, Slovakian, Polish, and Hungarian. Using a dataset of tweets collected via the Twitter/X academic API in 2023 and manually annotated for sentiment towards Russia and Ukraine, they fine-tuned models such as LLaMA~2~\cite{touvron2023llama2openfoundation}, LLaMA~3~\cite{dubey2024llama3herdmodels}, Mistral~\cite{jiang2023mistral7b}, BERT, and BERTweet, with GPT-4 as a reference. LLaMA~2 and Mistral achieved state-of-the-art results, while LLaMA~3 and BERTweet performed worse, likely due to the limited tunability of LLaMA~3 and outdated pre-training data of BERTweet. Fine-tuning proved significantly more effective than in-context learning, particularly for multilingual and culturally complex datasets. Translating data into English improved performance across all models; however, fine-tuned models like LLaMA~2 exhibited strong adaptability even in the original languages, surprising given their English-centric training. Challenges included misclassification in Polish tweets due to cultural nuances and limitations such as model rigidity and smaller pre-trained parameter sets. The study highlighted the effectiveness of fine-tuning on small datasets and the importance of task-specific and model-specific adaptations.

\citet{10504711} examine older deep neural networks, including LSTMs and BERT-based models, alongside LLMs for ATE and ACD tasks on several English datasets. Their findings emphasized PaLM’s robust performance, often surpassing BERT-like models across diverse domains, while identifying its limitations on challenging datasets like MAMS, which contain multiple aspects and sentiments in a single review. GPT-3.5 delivered competitive MAMS results comparable to specialized models like ATAE-LSTM, showcasing LLMs’ ability to handle complex datasets with minimal fine-tuning. The study demonstrated LLMs’ strengths in transfer learning, contextual reasoning, and domain versatility, exposing their gaps in handling nuanced tasks. Older models like ATAE-LSTM and DeBERTa~\cite{he2021debertadecodingenhancedbertdisentangled} excelled in domain-specific scenarios, underscoring the continued relevance of specialized training. The authors advocate exploring advanced LLMs, such as GPT-4, and refining models to better address nuanced sentiment and cross-domain challenges, offering valuable insights into the evolving ABSA landscape.

\citet{10.1145/3664647.3680705} introduce a groundbreaking framework for multimodal conversational ABSA, addressing gaps in integrating multimodality, conversational context, fine-grained sentiment granularity, dynamic sentiment shifts, and cognitive causal reasoning. Two new tasks were proposed: panoptic sentiment sextuple extraction, which identifies holder, target, aspect, opinion, sentiment, and rationale from multi-turn multimodal dialogues; and sentiment flipping analysis, which detects dynamic sentiment transformations and their causal factors. The PanoSent dataset, featuring multilingual, multimodal, and multi-scenario annotations, was created to benchmark these tasks. To tackle the challenges, a chain-of-sentiment reasoning framework, a novel multimodal LLM (Sentica), and a paraphrase-based verification mechanism were devised, outperforming strong baselines. This study’s multilingual approach (including English, Chinese, and Spanish) and multimodal emphasis offer a robust foundation for advancing cross-lingual ABSA, addressing gaps created by the lack of annotated data for complex tasks.

Furthermore, LLMs show potential for data augmentation~\cite{ding-etal-2024-data}, with studies leveraging LLMs to generate additional training data~\cite{li-etal-2022-controllable, moller-etal-2024-parrot}, including for English ABSA~\cite{zhong2024iterativedatagenerationlarge}. Such approaches could generate augmented data in target languages; however, their effectiveness may heavily depend on the quality of the generated data.

Despite their advantages, LLMs face several challenges. Closed-source models like GPT-4 are expensive to use and may raise privacy concerns for real-world applications. Open-source alternatives like LLaMA require substantial hardware resources for fine-tuning. For instance, fine-tuning a 13B-parameter model in full precision requires over 200 GB of GPU memory. Techniques like QLoRA~\cite{qlora}, which employ a 4-bit quantized frozen backbone with learnable LoRA~\cite{hu2022lora} parameters, reduce memory requirements and enable fine-tuning on consumer GPUs. However, even with these optimizations, LLMs remain resource-intensive compared to smaller, specialized models.

A significant limitation of many LLMs is their English-centric design~\cite{touvron2023llama2openfoundation, liu2024translationneedstudysolving}. While newer multilingual LLMs like LLaMA~3.1~\cite{dubey2024llama3herdmodels} and Aya~\cite{ustun-etal-2024-aya, aryabumi2024aya23openweight} have emerged, they often support fewer languages than multilingual pre-trained models like XLM-R. Limited language support results from the high cost and effort required to obtain quality multilingual data for instruction tuning, with quality being more critical than quantity~\cite{mitra2023orca}.

One major advantage of LLMs is their ability to generate results without labelled training data, which is costly and time-consuming to produce. This makes them especially appealing for tasks in under-resourced languages or domains where annotated datasets are scarce.

\subsection{Relationships to Cross-lingual ABSA}

Large language models have demonstrated effectiveness in simpler ABSA tasks under zero-shot and few-shot settings, offering quick solutions when labelled data is unavailable. However, smaller fine-tuned models often outperform LLMs for more complex tasks, such as predicting multiple sentiment element tuples or handling nuanced cross-lingual datasets. Techniques like chain-of-thought prompting and task-specific augmentations can enhance LLM performance in these scenarios. LLMs are also valuable for bridging language gaps, particularly when annotated data is available in a source language but not in the target language. Despite limitations such as high hardware requirements and performance gaps on challenging ABSA tasks, LLMs are useful for rapid prototyping, data augmentation, and addressing under-resourced languages.

Compared to traditional cross-lingual ABSA approaches, multilingual LLMs offer a notable advantage by eliminating the need for annotated data, even in a language different from the target language. LLMs can achieve acceptable performance for some tasks using zero-shot and few-shot prompts. However, research consistently shows that fine-tuned models outperform LLMs in most cases. That said, as LLMs continue to evolve, future iterations may match or exceed the performance of smaller multilingual models specifically fine-tuned for ABSA tasks.

\section{Challenges and Future Research}
\label{sec:future}
While monolingual ABSA has seen considerable progress with new approaches and tasks, cross-lingual ABSA remains significantly more challenging and underexplored. This section highlights key challenges and potential directions for future research.

\subsection{Quests for More Diverse Multilingual Datasets}
As noted in Section~\ref{sec:datasets}, the number of available multilingual ABSA datasets for cross-lingual transfer is extremely limited. Although some datasets cover multiple domains, only the SemEval-2016 dataset provides data in more than two languages. Most datasets focus on aspect terms and sentiment polarity, with only one offering aspect category annotations and none including opinion term annotations. This lack of annotated sentiment elements restricts the range of tasks that can be studied. Additionally, many datasets contain relatively few sentences, which hinders reliable comparison of models, especially for large pre-trained models with millions of parameters. To advance cross-lingual ABSA research, there is a need for more diverse datasets with annotations for multiple sentiment elements across different languages.

\subsection{Cross-lingual ABSA Tasks}
As discussed in Section~\ref{sec:tasks}, while numerous ABSA tasks exist, only a few have been explored in cross-lingual settings. This limitation is partly due to the lack of annotated sentiment elements, particularly opinion terms, which restricts the ability to explore opinion-related tasks. Although the SemEval-2016 dataset supports tasks with three linked sentiment elements (i.e. the TASD task), research has largely focused on tuple-based tasks, specifically E2E-ABSA, and single tasks focusing only on one sentiment element. Expanding the exploration of cross-lingual ABSA tasks requires more comprehensive multilingual datasets.

\subsection{Sequence-to-Sequence Modelling Paradigm and LLMs in Cross-lingual ABSA}

The sequence-to-sequence modelling paradigm, as discussed in Section~\ref{sec:seq2seq}, has recently gained significant traction in English ABSA research~\cite{zhang-etal-2021-aspect-sentiment, zhang-etal-2021-towards-generative, mao-etal-2022-seq2path, gou-etal-2023-mvp, xianlong-etal-2023-tagging}. However, its application in cross-lingual settings remains largely unexplored~\cite{vsmid2023cross}. This approach, particularly suitable for tasks involving multiple sentiment elements, holds great potential for addressing the complexities of cross-lingual ABSA.

Fine-tuned LLMs have demonstrated remarkable performance in monolingual ABSA~\cite{smid-etal-2024-llama}. However, their use in cross-lingual contexts has seen only limited investigation, except for a minor study focused on multilingualism~\cite{wu2024evaluatingzeroshotmultilingualaspectbased}. The development of multilingual large language models~\cite{dubey2024llama3herdmodels} presents an exciting opportunity for advancing cross-lingual ABSA. Fine-tuning these models could significantly improve addressing linguistic diversity and task complexity.

Looking ahead, the continual advancement in LLM performance for ABSA tasks opens the door to novel approaches that bypass fine-tuning entirely. Instead, LLMs could be employed with zero-shot or few-shot prompting strategies. Such methods eliminate the need for annotated training datasets, addressing the challenges posed by the scarcity of labelled resources in many languages. This paradigm shift could enable broader accessibility and applicability of cross-lingual ABSA solutions.

\subsection{Adopting Modern Approaches from Monolingual ABSA}

Cross-lingual ABSA research has predominantly concentrated on enhancing language transfer techniques to bridge the gap between source and target languages. Conversely, monolingual ABSA research has prioritized refining task-specific methodologies, resulting in several significant advancements tailored to ABSA. Despite their demonstrated success, many of these innovations from monolingual ABSA research have yet to be fully incorporated into cross-lingual studies.

Integrating state-of-the-art methods from monolingual ABSA, such as tagging-assisted generation or multi-view prompting, with the best-performing cross-lingual transfer techniques from cross-lingual ABSA could yield substantial performance improvements. Bridging these two lines of research offers a promising direction for future studies. By leveraging the complementary strengths of monolingual task-specific innovations and cross-lingual transfer methodologies, researchers can push the boundaries of what is achievable in cross-lingual ABSA.

\subsection{Dependency on Machine Translation}
Although multilingual pre-trained models offer inherent cross-lingual capabilities, they are often combined with machine translation to improve performance~\cite{zhang-etal-2021-cross, lin2023clxabsa, LinXABSA}. However, studies have shown that machine translation can sometimes degrade performance due to translation quality and label misalignment~\cite{li2020unsupervised}. Recent advancements in LLMs provide new possibilities for cross-lingual ABSA, as LLM-based data augmentation has emerged as a promising technique to enhance performance~\cite{ding2024dataaugmentationusinglarge}, with some studies using LLMs to generate additional data~\cite{li-etal-2022-controllable, moller-etal-2024-parrot, zhang2023recommendationinstructionfollowinglarge}. Future research could explore replacing reliance on machine translation with LLM-generated data.

\section{Conclusion}
\label{sec:conclusion}
This survey comprehensively reviews cross-lingual aspect-based sentiment analysis, representing the first dedicated survey to focus exclusively on this emerging field. It begins by outlining the necessary background, including the definition of aspect-based sentiment analysis, the four sentiment elements, and the various tasks and datasets used in cross-lingual ABSA research. Next, it describes the different modelling paradigms employed in ABSA, followed by an overview of approaches to cross-lingual transfer that enable sentiment analysis across languages, with examples of how these techniques have been applied in recent studies. We also offer a detailed summary of the ABSA tasks explored in cross-lingual settings, highlighting the methods used to address each task. The review also highlights the relevance of monolingual and multilingual ABSA research, as well as ABSA approaches with LLMs, in addressing challenges in cross-lingual ABSA. Finally, we discuss the main challenges facing the field and suggest potential directions for future research to advance cross-lingual ABSA.

\printcredits

\section*{Acknowledgements}
This work has been partly supported by the OP JAC project \quotes{R\&D of Technologies for Advanced Digitization in the Pilsen Metropolitan Area (DigiTech)} No.: CZ.02.01.01/00/23\_021/0008436 co-financed by the European Union. Computational resources were provided by the e-INFRA CZ project (ID:90254), supported by the Ministry of Education, Youth and Sports of the Czech Republic.

\section*{Declaration of Generative AI and AI-assisted Technologies in the Writing Process}
During the preparation of this work the authors used the ChatGPT\footnote{\url{https://chat.openai.com}} tool in order to readability, improve language, and correct grammatical errors. After using this tool, the authors reviewed and edited the content as needed and take full responsibility for the content of the publication. The tool was not used to analyse data or derive insights during the research process.

\section*{Declaration of Competing Interest}
The authors declare that they have no known competing financial interests or personal relationships that could have appeared to influence the work reported in this paper.

\bibliographystyle{model1-num-names}

\bibliography{bibliography}

\begin{thebibliography}{143}
\expandafter\ifx\csname natexlab\endcsname\relax\def\natexlab#1{#1}\fi
\providecommand{\url}[1]{\texttt{#1}}
\providecommand{\href}[2]{#2}
\providecommand{\path}[1]{#1}
\providecommand{\DOIprefix}{doi:}
\providecommand{\ArXivprefix}{arXiv:}
\providecommand{\URLprefix}{URL: }
\providecommand{\Pubmedprefix}{pmid:}
\providecommand{\doi}[1]{\href{http://dx.doi.org/#1}{\path{#1}}}
\providecommand{\Pubmed}[1]{\href{pmid:#1}{\path{#1}}}
\providecommand{\bibinfo}[2]{#2}
\ifx\xfnm\relax \def\xfnm[#1]{\unskip,\space#1}\fi
\bibitem[{Liu(2010)}]{liu2010sentiment}
\bibinfo{author}{B.~Liu},
\newblock \bibinfo{title}{Sentiment analysis and subjectivity.},
\newblock \bibinfo{journal}{Handbook of natural language processing} \bibinfo{volume}{2} (\bibinfo{year}{2010}) \bibinfo{pages}{627--666}.
\bibitem[{Zhang et~al.(2022)Zhang, Li, Deng, Bing, and Lam}]{zhang2022survey}
\bibinfo{author}{W.~Zhang}, \bibinfo{author}{X.~Li}, \bibinfo{author}{Y.~Deng}, \bibinfo{author}{L.~Bing}, \bibinfo{author}{W.~Lam},
\newblock \bibinfo{title}{A survey on aspect-based sentiment analysis: Tasks, methods, and challenges},
\newblock \bibinfo{journal}{IEEE Transactions on Knowledge and Data Engineering}  (\bibinfo{year}{2022}).
\bibitem[{Pontiki et~al.(2016)Pontiki, Galanis, Papageorgiou, Androutsopoulos, Manandhar, AL-Smadi, Al-Ayyoub, Zhao, Qin, De~Clercq, Hoste, Apidianaki, Tannier, Loukachevitch, Kotelnikov, Bel, Jim{\'e}nez-Zafra, and Eryi{\u{g}}it}]{pontiki-etal-2016-semeval}
\bibinfo{author}{M.~Pontiki}, \bibinfo{author}{D.~Galanis}, \bibinfo{author}{H.~Papageorgiou}, \bibinfo{author}{I.~Androutsopoulos}, \bibinfo{author}{S.~Manandhar}, \bibinfo{author}{M.~AL-Smadi}, \bibinfo{author}{M.~Al-Ayyoub}, \bibinfo{author}{Y.~Zhao}, \bibinfo{author}{B.~Qin}, \bibinfo{author}{O.~De~Clercq}, \bibinfo{author}{V.~Hoste}, \bibinfo{author}{M.~Apidianaki}, \bibinfo{author}{X.~Tannier}, \bibinfo{author}{N.~Loukachevitch}, \bibinfo{author}{E.~Kotelnikov}, \bibinfo{author}{N.~Bel}, \bibinfo{author}{S.~M. Jim{\'e}nez-Zafra}, \bibinfo{author}{G.~Eryi{\u{g}}it},
\newblock \bibinfo{title}{{S}em{E}val-2016 task 5: Aspect based sentiment analysis},
\newblock in: \bibinfo{editor}{S.~Bethard}, \bibinfo{editor}{M.~Carpuat}, \bibinfo{editor}{D.~Cer}, \bibinfo{editor}{D.~Jurgens}, \bibinfo{editor}{P.~Nakov}, \bibinfo{editor}{T.~Zesch} (Eds.), \bibinfo{booktitle}{Proceedings of the 10th International Workshop on Semantic Evaluation ({S}em{E}val-2016)}, \bibinfo{publisher}{Association for Computational Linguistics}, \bibinfo{address}{San Diego, California}, \bibinfo{year}{2016}, pp. \bibinfo{pages}{19--30}. \URLprefix \url{https://aclanthology.org/S16-1002}. \DOIprefix\doi{10.18653/v1/S16-1002}.
\bibitem[{Lambert(2015)}]{lambert-2015-aspect}
\bibinfo{author}{P.~Lambert},
\newblock \bibinfo{title}{Aspect-level cross-lingual sentiment classification with constrained {SMT}},
\newblock in: \bibinfo{editor}{C.~Zong}, \bibinfo{editor}{M.~Strube} (Eds.), \bibinfo{booktitle}{Proceedings of the 53rd Annual Meeting of the Association for Computational Linguistics and the 7th International Joint Conference on Natural Language Processing (Volume 2: Short Papers)}, \bibinfo{publisher}{Association for Computational Linguistics}, \bibinfo{address}{Beijing, China}, \bibinfo{year}{2015}, pp. \bibinfo{pages}{781--787}. \URLprefix \url{https://aclanthology.org/P15-2128}. \DOIprefix\doi{10.3115/v1/P15-2128}.
\bibitem[{Zhou et~al.(2015)Zhou, Wan, and Xiao}]{zhou2015clopinionminer}
\bibinfo{author}{X.~Zhou}, \bibinfo{author}{X.~Wan}, \bibinfo{author}{J.~Xiao},
\newblock \bibinfo{title}{Clopinionminer: Opinion target extraction in a cross-language scenario},
\newblock \bibinfo{journal}{IEEE/ACM Transactions on Audio, Speech, and Language Processing} \bibinfo{volume}{23} (\bibinfo{year}{2015}) \bibinfo{pages}{619--630}.
\bibitem[{Barnes et~al.(2016)Barnes, Lambert, and Badia}]{barnes-etal-2016-exploring}
\bibinfo{author}{J.~Barnes}, \bibinfo{author}{P.~Lambert}, \bibinfo{author}{T.~Badia},
\newblock \bibinfo{title}{Exploring distributional representations and machine translation for aspect-based cross-lingual sentiment classification.},
\newblock in: \bibinfo{editor}{Y.~Matsumoto}, \bibinfo{editor}{R.~Prasad} (Eds.), \bibinfo{booktitle}{Proceedings of {COLING} 2016, the 26th International Conference on Computational Linguistics: Technical Papers}, \bibinfo{publisher}{The COLING 2016 Organizing Committee}, \bibinfo{address}{Osaka, Japan}, \bibinfo{year}{2016}, pp. \bibinfo{pages}{1613--1623}. \URLprefix \url{https://aclanthology.org/C16-1152}.
\bibitem[{Akhtar et~al.(2018)Akhtar, Sawant, Sen, Ekbal, and Bhattacharyya}]{akhtar-etal-2018-solving}
\bibinfo{author}{M.~S. Akhtar}, \bibinfo{author}{P.~Sawant}, \bibinfo{author}{S.~Sen}, \bibinfo{author}{A.~Ekbal}, \bibinfo{author}{P.~Bhattacharyya},
\newblock \bibinfo{title}{Solving data sparsity for aspect based sentiment analysis using cross-linguality and multi-linguality},
\newblock in: \bibinfo{editor}{M.~Walker}, \bibinfo{editor}{H.~Ji}, \bibinfo{editor}{A.~Stent} (Eds.), \bibinfo{booktitle}{Proceedings of the 2018 Conference of the North {A}merican Chapter of the Association for Computational Linguistics: Human Language Technologies, Volume 1 (Long Papers)}, \bibinfo{publisher}{Association for Computational Linguistics}, \bibinfo{address}{New Orleans, Louisiana}, \bibinfo{year}{2018}, pp. \bibinfo{pages}{572--582}. \URLprefix \url{https://aclanthology.org/N18-1053}. \DOIprefix\doi{10.18653/v1/N18-1053}.
\bibitem[{Devlin et~al.(2019)Devlin, Chang, Lee, and Toutanova}]{devlin-etal-2019-bert}
\bibinfo{author}{J.~Devlin}, \bibinfo{author}{M.-W. Chang}, \bibinfo{author}{K.~Lee}, \bibinfo{author}{K.~Toutanova},
\newblock \bibinfo{title}{{BERT}: Pre-training of deep bidirectional transformers for language understanding},
\newblock in: \bibinfo{editor}{J.~Burstein}, \bibinfo{editor}{C.~Doran}, \bibinfo{editor}{T.~Solorio} (Eds.), \bibinfo{booktitle}{Proceedings of the 2019 Conference of the North {A}merican Chapter of the Association for Computational Linguistics: Human Language Technologies, Volume 1 (Long and Short Papers)}, \bibinfo{publisher}{Association for Computational Linguistics}, \bibinfo{address}{Minneapolis, Minnesota}, \bibinfo{year}{2019}, pp. \bibinfo{pages}{4171--4186}. \URLprefix \url{https://aclanthology.org/N19-1423}. \DOIprefix\doi{10.18653/v1/N19-1423}.
\bibitem[{Conneau et~al.(2020)Conneau, Khandelwal, Goyal, Chaudhary, Wenzek, Guzm{\'a}n, Grave, Ott, Zettlemoyer, and Stoyanov}]{conneau-etal-2020-unsupervised}
\bibinfo{author}{A.~Conneau}, \bibinfo{author}{K.~Khandelwal}, \bibinfo{author}{N.~Goyal}, \bibinfo{author}{V.~Chaudhary}, \bibinfo{author}{G.~Wenzek}, \bibinfo{author}{F.~Guzm{\'a}n}, \bibinfo{author}{E.~Grave}, \bibinfo{author}{M.~Ott}, \bibinfo{author}{L.~Zettlemoyer}, \bibinfo{author}{V.~Stoyanov},
\newblock \bibinfo{title}{Unsupervised cross-lingual representation learning at scale},
\newblock in: \bibinfo{editor}{D.~Jurafsky}, \bibinfo{editor}{J.~Chai}, \bibinfo{editor}{N.~Schluter}, \bibinfo{editor}{J.~Tetreault} (Eds.), \bibinfo{booktitle}{Proceedings of the 58th Annual Meeting of the Association for Computational Linguistics}, \bibinfo{publisher}{Association for Computational Linguistics}, \bibinfo{address}{Online}, \bibinfo{year}{2020}, pp. \bibinfo{pages}{8440--8451}. \URLprefix \url{https://aclanthology.org/2020.acl-main.747}. \DOIprefix\doi{10.18653/v1/2020.acl-main.747}.
\bibitem[{Hu et~al.(2020)Hu, Ruder, Siddhant, Neubig, Firat, and Johnson}]{pmlr-v119-hu20b}
\bibinfo{author}{J.~Hu}, \bibinfo{author}{S.~Ruder}, \bibinfo{author}{A.~Siddhant}, \bibinfo{author}{G.~Neubig}, \bibinfo{author}{O.~Firat}, \bibinfo{author}{M.~Johnson},
\newblock \bibinfo{title}{{XTREME}: A massively multilingual multi-task benchmark for evaluating cross-lingual generalisation},
\newblock in: \bibinfo{editor}{H.~D. III}, \bibinfo{editor}{A.~Singh} (Eds.), \bibinfo{booktitle}{Proceedings of the 37th International Conference on Machine Learning}, volume \bibinfo{volume}{119} of \textit{\bibinfo{series}{Proceedings of Machine Learning Research}}, \bibinfo{publisher}{PMLR}, \bibinfo{year}{2020}, pp. \bibinfo{pages}{4411--4421}. \URLprefix \url{https://proceedings.mlr.press/v119/hu20b.html}.
\bibitem[{Li et~al.(2020)Li, Bing, Zhang, Li, and Lam}]{li2020unsupervised}
\bibinfo{author}{X.~Li}, \bibinfo{author}{L.~Bing}, \bibinfo{author}{W.~Zhang}, \bibinfo{author}{Z.~Li}, \bibinfo{author}{W.~Lam},
\newblock \bibinfo{title}{Unsupervised cross-lingual adaptation for sequence tagging and beyond},
\newblock \bibinfo{journal}{arXiv preprint arXiv:2010.12405}  (\bibinfo{year}{2020}).
\bibitem[{Zhang et~al.(2021)Zhang, He, Peng, Bing, and Lam}]{zhang-etal-2021-cross}
\bibinfo{author}{W.~Zhang}, \bibinfo{author}{R.~He}, \bibinfo{author}{H.~Peng}, \bibinfo{author}{L.~Bing}, \bibinfo{author}{W.~Lam},
\newblock \bibinfo{title}{Cross-lingual aspect-based sentiment analysis with aspect term code-switching},
\newblock in: \bibinfo{editor}{M.-F. Moens}, \bibinfo{editor}{X.~Huang}, \bibinfo{editor}{L.~Specia}, \bibinfo{editor}{S.~W.-t. Yih} (Eds.), \bibinfo{booktitle}{Proceedings of the 2021 Conference on Empirical Methods in Natural Language Processing}, \bibinfo{publisher}{Association for Computational Linguistics}, \bibinfo{address}{Online and Punta Cana, Dominican Republic}, \bibinfo{year}{2021}, pp. \bibinfo{pages}{9220--9230}. \URLprefix \url{https://aclanthology.org/2021.emnlp-main.727}. \DOIprefix\doi{10.18653/v1/2021.emnlp-main.727}.
\bibitem[{Lin et~al.(2023)Lin, Fu, Lin, Zhou, Yang, and Jiang}]{lin2023clxabsa}
\bibinfo{author}{N.~Lin}, \bibinfo{author}{Y.~Fu}, \bibinfo{author}{X.~Lin}, \bibinfo{author}{D.~Zhou}, \bibinfo{author}{A.~Yang}, \bibinfo{author}{S.~Jiang},
\newblock \bibinfo{title}{Cl-xabsa: Contrastive learning for cross-lingual aspect-based sentiment analysis},
\newblock \bibinfo{journal}{IEEE/ACM Transactions on Audio, Speech, and Language Processing}  (\bibinfo{year}{2023}).
\bibitem[{Lin et~al.(2024)Lin, Zeng, Liao, Liu, Yang, and Zhou}]{LinXABSA}
\bibinfo{author}{N.~Lin}, \bibinfo{author}{M.~Zeng}, \bibinfo{author}{X.~Liao}, \bibinfo{author}{W.~Liu}, \bibinfo{author}{A.~Yang}, \bibinfo{author}{D.~Zhou},
\newblock \bibinfo{title}{Addressing class-imbalance challenges in cross-lingual aspect-based sentiment analysis: Dynamic weighted loss and anti-decoupling},
\newblock \bibinfo{journal}{Expert Systems with Applications} \bibinfo{volume}{257} (\bibinfo{year}{2024}) \bibinfo{pages}{125059}.
\bibitem[{Poria et~al.(2023)Poria, Hazarika, Majumder, and Mihalcea}]{9260964}
\bibinfo{author}{S.~Poria}, \bibinfo{author}{D.~Hazarika}, \bibinfo{author}{N.~Majumder}, \bibinfo{author}{R.~Mihalcea},
\newblock \bibinfo{title}{{ Beneath the Tip of the Iceberg: Current Challenges and New Directions in Sentiment Analysis Research }},
\newblock \bibinfo{journal}{IEEE Transactions on Affective Computing} \bibinfo{volume}{14} (\bibinfo{year}{2023}) \bibinfo{pages}{108--132}.
\bibitem[{Schouten and Frasincar(2016)}]{7286808}
\bibinfo{author}{K.~Schouten}, \bibinfo{author}{F.~Frasincar},
\newblock \bibinfo{title}{Survey on aspect-level sentiment analysis},
\newblock \bibinfo{journal}{IEEE Transactions on Knowledge and Data Engineering} \bibinfo{volume}{28} (\bibinfo{year}{2016}) \bibinfo{pages}{813--830}.
\bibitem[{Nazir et~al.(2022)Nazir, Rao, Wu, and Sun}]{8976252}
\bibinfo{author}{A.~Nazir}, \bibinfo{author}{Y.~Rao}, \bibinfo{author}{L.~Wu}, \bibinfo{author}{L.~Sun},
\newblock \bibinfo{title}{Issues and challenges of aspect-based sentiment analysis: A comprehensive survey},
\newblock \bibinfo{journal}{IEEE Transactions on Affective Computing} \bibinfo{volume}{13} (\bibinfo{year}{2022}) \bibinfo{pages}{845--863}.
\bibitem[{Brauwers and Frasincar(2022)}]{10.1145/3503044}
\bibinfo{author}{G.~Brauwers}, \bibinfo{author}{F.~Frasincar},
\newblock \bibinfo{title}{A survey on aspect-based sentiment classification},
\newblock \bibinfo{journal}{ACM Comput. Surv.} \bibinfo{volume}{55} (\bibinfo{year}{2022}).
\bibitem[{Chauhan et~al.(2023)Chauhan, Nahta, Meena, and Gopalani}]{CHAUHAN2023100576}
\bibinfo{author}{G.~S. Chauhan}, \bibinfo{author}{R.~Nahta}, \bibinfo{author}{Y.~K. Meena}, \bibinfo{author}{D.~Gopalani},
\newblock \bibinfo{title}{Aspect based sentiment analysis using deep learning approaches: A survey},
\newblock \bibinfo{journal}{Computer Science Review} \bibinfo{volume}{49} (\bibinfo{year}{2023}) \bibinfo{pages}{100576}.
\bibitem[{Zhao et~al.(2024)Zhao, ang Meng, and Song}]{ZHAO2024102552}
\bibinfo{author}{T.~Zhao}, \bibinfo{author}{L.~ang Meng}, \bibinfo{author}{D.~Song},
\newblock \bibinfo{title}{Multimodal aspect-based sentiment analysis: A survey of tasks, methods, challenges and future directions},
\newblock \bibinfo{journal}{Information Fusion} \bibinfo{volume}{112} (\bibinfo{year}{2024}) \bibinfo{pages}{102552}.
\bibitem[{Liu(2022)}]{liu2022sentiment}
\bibinfo{author}{B.~Liu}, \bibinfo{title}{Sentiment analysis and opinion mining}, \bibinfo{publisher}{Springer Nature}, \bibinfo{year}{2022}.
\bibitem[{Lin et~al.(2014)Lin, Jin, Xu, Wang, Cheng, and Wang}]{10.1145/2661829.2662019}
\bibinfo{author}{Z.~Lin}, \bibinfo{author}{X.~Jin}, \bibinfo{author}{X.~Xu}, \bibinfo{author}{W.~Wang}, \bibinfo{author}{X.~Cheng}, \bibinfo{author}{Y.~Wang},
\newblock \bibinfo{title}{A cross-lingual joint aspect/sentiment model for sentiment analysis},
\newblock in: \bibinfo{booktitle}{Proceedings of the 23rd ACM International Conference on Conference on Information and Knowledge Management}, CIKM '14, \bibinfo{publisher}{Association for Computing Machinery}, \bibinfo{address}{New York, NY, USA}, \bibinfo{year}{2014}, p. \bibinfo{pages}{1089–1098}. \URLprefix \url{https://doi.org/10.1145/2661829.2662019}. \DOIprefix\doi{10.1145/2661829.2662019}.
\bibitem[{Klinger and Cimiano(2015)}]{klinger-cimiano-2015-instance}
\bibinfo{author}{R.~Klinger}, \bibinfo{author}{P.~Cimiano},
\newblock \bibinfo{title}{Instance selection improves cross-lingual model training for fine-grained sentiment analysis},
\newblock in: \bibinfo{booktitle}{Proceedings of the Nineteenth Conference on Computational Natural Language Learning}, \bibinfo{publisher}{Association for Computational Linguistics}, \bibinfo{address}{Beijing, China}, \bibinfo{year}{2015}, pp. \bibinfo{pages}{153--163}. \URLprefix \url{https://aclanthology.org/K15-1016}. \DOIprefix\doi{10.18653/v1/K15-1016}.
\bibitem[{Akhtar et~al.(2018)Akhtar, Sawant, Sen, Ekbal, and Bhattacharyya}]{10.1145/3273931}
\bibinfo{author}{M.~S. Akhtar}, \bibinfo{author}{P.~Sawant}, \bibinfo{author}{S.~Sen}, \bibinfo{author}{A.~Ekbal}, \bibinfo{author}{P.~Bhattacharyya},
\newblock \bibinfo{title}{Improving word embedding coverage in less-resourced languages through multi-linguality and cross-linguality: A case study with aspect-based sentiment analysis},
\newblock \bibinfo{journal}{ACM Trans. Asian Low-Resour. Lang. Inf. Process.} \bibinfo{volume}{18} (\bibinfo{year}{2018}).
\bibitem[{Wang and Pan(2018)}]{wang2018transition}
\bibinfo{author}{W.~Wang}, \bibinfo{author}{S.~J. Pan},
\newblock \bibinfo{title}{Transition-based adversarial network for cross-lingual aspect extraction.},
\newblock in: \bibinfo{booktitle}{IJCAI}, \bibinfo{year}{2018}, pp. \bibinfo{pages}{4475--4481}.
\bibitem[{Jebbara and Cimiano(2019)}]{jebbara-cimiano-2019-zero}
\bibinfo{author}{S.~Jebbara}, \bibinfo{author}{P.~Cimiano},
\newblock \bibinfo{title}{{Z}ero-shot cross-lingual opinion target extraction},
\newblock in: \bibinfo{editor}{J.~Burstein}, \bibinfo{editor}{C.~Doran}, \bibinfo{editor}{T.~Solorio} (Eds.), \bibinfo{booktitle}{Proceedings of the 2019 Conference of the North {A}merican Chapter of the Association for Computational Linguistics: Human Language Technologies, Volume 1 (Long and Short Papers)}, \bibinfo{publisher}{Association for Computational Linguistics}, \bibinfo{address}{Minneapolis, Minnesota}, \bibinfo{year}{2019}, pp. \bibinfo{pages}{2486--2495}. \URLprefix \url{https://aclanthology.org/N19-1257}. \DOIprefix\doi{10.18653/v1/N19-1257}.
\bibitem[{Phan et~al.(2021)Phan, Ngoc~Hao, Thin, and Luu-Thuy~Nguyen}]{9585242}
\bibinfo{author}{K.~T.-K. Phan}, \bibinfo{author}{D.~Ngoc~Hao}, \bibinfo{author}{D.~V. Thin}, \bibinfo{author}{N.~Luu-Thuy~Nguyen},
\newblock \bibinfo{title}{Exploring zero-shot cross-lingual aspect-based sentiment analysis using pre-trained multilingual language models},
\newblock in: \bibinfo{booktitle}{2021 International Conference on Multimedia Analysis and Pattern Recognition (MAPR)}, \bibinfo{year}{2021}, pp. \bibinfo{pages}{1--6}. \DOIprefix\doi{10.1109/MAPR53640.2021.9585242}.
\bibitem[{Wu et~al.(2021)Wu, Wang, Qing, and Li}]{wu2021reinforced}
\bibinfo{author}{H.~Wu}, \bibinfo{author}{Z.~Wang}, \bibinfo{author}{F.~Qing}, \bibinfo{author}{S.~Li},
\newblock \bibinfo{title}{Reinforced transformer with cross-lingual distillation for cross-lingual aspect sentiment classification},
\newblock \bibinfo{journal}{Electronics} \bibinfo{volume}{10} (\bibinfo{year}{2021}) \bibinfo{pages}{270}.
\bibitem[{Dang Van~Thin and Nguyen(2023)}]{doi:10.1080/24751839.2023.2173843}
\bibinfo{author}{D.~N.~H. Dang Van~Thin, Hung Quoc~Ngo}, \bibinfo{author}{N.~L.-T. Nguyen},
\newblock \bibinfo{title}{Exploring zero-shot and joint training cross-lingual strategies for aspect-based sentiment analysis based on contextualized multilingual language models},
\newblock \bibinfo{journal}{Journal of Information and Telecommunication} \bibinfo{volume}{7} (\bibinfo{year}{2023}) \bibinfo{pages}{121--143}.
\bibitem[{Nam(2024)}]{nam2024kpc}
\bibinfo{author}{K.~Nam},
\newblock \bibinfo{title}{Kpc-cf: Korean aspect-based sentiment analysis via pseudo-classifier with corpus filtering for low resource society},
\newblock in: \bibinfo{booktitle}{First International Conference on Addressing Socioethical Effects of Artificial Intelligence}, \bibinfo{year}{2024}.
\bibitem[{Sattar et~al.(2021)Sattar, Umer, Vasbieva, Chung, Latif, and Lee}]{9550786}
\bibinfo{author}{K.~Sattar}, \bibinfo{author}{Q.~Umer}, \bibinfo{author}{D.~G. Vasbieva}, \bibinfo{author}{S.~Chung}, \bibinfo{author}{Z.~Latif}, \bibinfo{author}{C.~Lee},
\newblock \bibinfo{title}{A multi-layer network for aspect-based cross-lingual sentiment classification},
\newblock \bibinfo{journal}{IEEE Access} \bibinfo{volume}{9} (\bibinfo{year}{2021}) \bibinfo{pages}{133961--133973}.
\bibitem[{Szołomicka and Kocon(2022)}]{10031113}
\bibinfo{author}{J.~Szołomicka}, \bibinfo{author}{J.~Kocon},
\newblock \bibinfo{title}{Multiaspectemo: Multilingual and language-agnostic aspect-based sentiment analysis},
\newblock in: \bibinfo{booktitle}{2022 IEEE International Conference on Data Mining Workshops (ICDMW)}, \bibinfo{year}{2022}, pp. \bibinfo{pages}{443--450}. \DOIprefix\doi{10.1109/ICDMW58026.2022.00065}.
\bibitem[{Wu et~al.(2024)Wu, Ma, Zhang, Deng, He, and Xue}]{wu2024evaluatingzeroshotmultilingualaspectbased}
\bibinfo{author}{C.~Wu}, \bibinfo{author}{B.~Ma}, \bibinfo{author}{Z.~Zhang}, \bibinfo{author}{N.~Deng}, \bibinfo{author}{Y.~He}, \bibinfo{author}{Y.~Xue}, \bibinfo{title}{Evaluating zero-shot multilingual aspect-based sentiment analysis with large language models}, \bibinfo{year}{2024}. \URLprefix \url{https://arxiv.org/abs/2412.12564}. \href{http://arxiv.org/abs/2412.12564}{\tt arXiv:2412.12564}.
\bibitem[{Klinger and Cimiano(2014)}]{klinger-cimiano-2014-usage}
\bibinfo{author}{R.~Klinger}, \bibinfo{author}{P.~Cimiano},
\newblock \bibinfo{title}{The {USAGE} review corpus for fine grained multi lingual opinion analysis},
\newblock in: \bibinfo{editor}{N.~Calzolari}, \bibinfo{editor}{K.~Choukri}, \bibinfo{editor}{T.~Declerck}, \bibinfo{editor}{H.~Loftsson}, \bibinfo{editor}{B.~Maegaard}, \bibinfo{editor}{J.~Mariani}, \bibinfo{editor}{A.~Moreno}, \bibinfo{editor}{J.~Odijk}, \bibinfo{editor}{S.~Piperidis} (Eds.), \bibinfo{booktitle}{Proceedings of the Ninth International Conference on Language Resources and Evaluation ({LREC}'14)}, \bibinfo{publisher}{European Language Resources Association (ELRA)}, \bibinfo{address}{Reykjavik, Iceland}, \bibinfo{year}{2014}, pp. \bibinfo{pages}{2211--2218}. \URLprefix \url{http://www.lrec-conf.org/proceedings/lrec2014/pdf/85_Paper.pdf}.
\bibitem[{Agerri et~al.(2013)Agerri, Cuadros, Gaines, and Rigau}]{agerri2013opener}
\bibinfo{author}{R.~Agerri}, \bibinfo{author}{M.~Cuadros}, \bibinfo{author}{S.~Gaines}, \bibinfo{author}{G.~Rigau},
\newblock \bibinfo{title}{Opener: Open polarity enhanced named entity recognition},
\newblock \bibinfo{journal}{Procesamiento del Lenguaje Natural}  (\bibinfo{year}{2013}) \bibinfo{pages}{215--218}.
\bibitem[{Hyun et~al.(2020)Hyun, Cho, and Yu}]{hyun-etal-2020-building}
\bibinfo{author}{D.~Hyun}, \bibinfo{author}{J.~Cho}, \bibinfo{author}{H.~Yu},
\newblock \bibinfo{title}{Building large-scale {E}nglish and {K}orean datasets for aspect-level sentiment analysis in automotive domain},
\newblock in: \bibinfo{editor}{D.~Scott}, \bibinfo{editor}{N.~Bel}, \bibinfo{editor}{C.~Zong} (Eds.), \bibinfo{booktitle}{Proceedings of the 28th International Conference on Computational Linguistics}, \bibinfo{publisher}{International Committee on Computational Linguistics}, \bibinfo{address}{Barcelona, Spain (Online)}, \bibinfo{year}{2020}, pp. \bibinfo{pages}{961--966}. \URLprefix \url{https://aclanthology.org/2020.coling-main.83}. \DOIprefix\doi{10.18653/v1/2020.coling-main.83}.
\bibitem[{Luo et~al.(2024)Luo, Fei, Li, Wu, Liu, Poria, Cambria, Lee, and Hsu}]{10.1145/3664647.3680705}
\bibinfo{author}{M.~Luo}, \bibinfo{author}{H.~Fei}, \bibinfo{author}{B.~Li}, \bibinfo{author}{S.~Wu}, \bibinfo{author}{Q.~Liu}, \bibinfo{author}{S.~Poria}, \bibinfo{author}{E.~Cambria}, \bibinfo{author}{M.-L. Lee}, \bibinfo{author}{W.~Hsu},
\newblock \bibinfo{title}{Panosent: A panoptic sextuple extraction benchmark for multimodal conversational aspect-based sentiment analysis},
\newblock in: \bibinfo{booktitle}{Proceedings of the 32nd ACM International Conference on Multimedia}, MM '24, \bibinfo{publisher}{Association for Computing Machinery}, \bibinfo{address}{New York, NY, USA}, \bibinfo{year}{2024}, p. \bibinfo{pages}{7667–7676}. \URLprefix \url{https://doi.org/10.1145/3664647.3680705}. \DOIprefix\doi{10.1145/3664647.3680705}.
\bibitem[{Pontiki et~al.(2014)Pontiki, Galanis, Pavlopoulos, Papageorgiou, Androutsopoulos, and Manandhar}]{pontiki-etal-2014-semeval}
\bibinfo{author}{M.~Pontiki}, \bibinfo{author}{D.~Galanis}, \bibinfo{author}{J.~Pavlopoulos}, \bibinfo{author}{H.~Papageorgiou}, \bibinfo{author}{I.~Androutsopoulos}, \bibinfo{author}{S.~Manandhar},
\newblock \bibinfo{title}{{S}em{E}val-2014 task 4: Aspect based sentiment analysis},
\newblock in: \bibinfo{editor}{P.~Nakov}, \bibinfo{editor}{T.~Zesch} (Eds.), \bibinfo{booktitle}{Proceedings of the 8th International Workshop on Semantic Evaluation ({S}em{E}val 2014)}, \bibinfo{publisher}{Association for Computational Linguistics}, \bibinfo{address}{Dublin, Ireland}, \bibinfo{year}{2014}, pp. \bibinfo{pages}{27--35}. \URLprefix \url{https://aclanthology.org/S14-2004}. \DOIprefix\doi{10.3115/v1/S14-2004}.
\bibitem[{Pontiki et~al.(2015)Pontiki, Galanis, Papageorgiou, Manandhar, and Androutsopoulos}]{pontiki-etal-2015-semeval}
\bibinfo{author}{M.~Pontiki}, \bibinfo{author}{D.~Galanis}, \bibinfo{author}{H.~Papageorgiou}, \bibinfo{author}{S.~Manandhar}, \bibinfo{author}{I.~Androutsopoulos},
\newblock \bibinfo{title}{{S}em{E}val-2015 task 12: Aspect based sentiment analysis},
\newblock in: \bibinfo{editor}{P.~Nakov}, \bibinfo{editor}{T.~Zesch}, \bibinfo{editor}{D.~Cer}, \bibinfo{editor}{D.~Jurgens} (Eds.), \bibinfo{booktitle}{Proceedings of the 9th International Workshop on Semantic Evaluation ({S}em{E}val 2015)}, \bibinfo{publisher}{Association for Computational Linguistics}, \bibinfo{address}{Denver, Colorado}, \bibinfo{year}{2015}, pp. \bibinfo{pages}{486--495}. \URLprefix \url{https://aclanthology.org/S15-2082}. \DOIprefix\doi{10.18653/v1/S15-2082}.
\bibitem[{Fan et~al.(2019)Fan, Wu, Dai, Huang, and Chen}]{fan-etal-2019-target}
\bibinfo{author}{Z.~Fan}, \bibinfo{author}{Z.~Wu}, \bibinfo{author}{X.-Y. Dai}, \bibinfo{author}{S.~Huang}, \bibinfo{author}{J.~Chen},
\newblock \bibinfo{title}{Target-oriented opinion words extraction with target-fused neural sequence labeling},
\newblock in: \bibinfo{editor}{J.~Burstein}, \bibinfo{editor}{C.~Doran}, \bibinfo{editor}{T.~Solorio} (Eds.), \bibinfo{booktitle}{Proceedings of the 2019 Conference of the North {A}merican Chapter of the Association for Computational Linguistics: Human Language Technologies, Volume 1 (Long and Short Papers)}, \bibinfo{publisher}{Association for Computational Linguistics}, \bibinfo{address}{Minneapolis, Minnesota}, \bibinfo{year}{2019}, pp. \bibinfo{pages}{2509--2518}. \URLprefix \url{https://aclanthology.org/N19-1259}. \DOIprefix\doi{10.18653/v1/N19-1259}.
\bibitem[{Xu et~al.(2020)Xu, Li, Lu, and Bing}]{xu-etal-2020-position}
\bibinfo{author}{L.~Xu}, \bibinfo{author}{H.~Li}, \bibinfo{author}{W.~Lu}, \bibinfo{author}{L.~Bing},
\newblock \bibinfo{title}{Position-aware tagging for aspect sentiment triplet extraction},
\newblock in: \bibinfo{editor}{B.~Webber}, \bibinfo{editor}{T.~Cohn}, \bibinfo{editor}{Y.~He}, \bibinfo{editor}{Y.~Liu} (Eds.), \bibinfo{booktitle}{Proceedings of the 2020 Conference on Empirical Methods in Natural Language Processing (EMNLP)}, \bibinfo{publisher}{Association for Computational Linguistics}, \bibinfo{address}{Online}, \bibinfo{year}{2020}, pp. \bibinfo{pages}{2339--2349}. \URLprefix \url{https://aclanthology.org/2020.emnlp-main.183}. \DOIprefix\doi{10.18653/v1/2020.emnlp-main.183}.
\bibitem[{Peng et~al.(2020)Peng, Xu, Bing, Huang, Lu, and Si}]{aste}
\bibinfo{author}{H.~Peng}, \bibinfo{author}{L.~Xu}, \bibinfo{author}{L.~Bing}, \bibinfo{author}{F.~Huang}, \bibinfo{author}{W.~Lu}, \bibinfo{author}{L.~Si},
\newblock \bibinfo{title}{Knowing what, how and why: A near complete solution for aspect-based sentiment analysis},
\newblock \bibinfo{journal}{Proceedings of the AAAI Conference on Artificial Intelligence} \bibinfo{volume}{34} (\bibinfo{year}{2020}) \bibinfo{pages}{8600--8607}.
\bibitem[{Cai et~al.(2021)Cai, Xia, and Yu}]{cai-etal-2021-aspect}
\bibinfo{author}{H.~Cai}, \bibinfo{author}{R.~Xia}, \bibinfo{author}{J.~Yu},
\newblock \bibinfo{title}{Aspect-category-opinion-sentiment quadruple extraction with implicit aspects and opinions},
\newblock in: \bibinfo{editor}{C.~Zong}, \bibinfo{editor}{F.~Xia}, \bibinfo{editor}{W.~Li}, \bibinfo{editor}{R.~Navigli} (Eds.), \bibinfo{booktitle}{Proceedings of the 59th Annual Meeting of the Association for Computational Linguistics and the 11th International Joint Conference on Natural Language Processing (Volume 1: Long Papers)}, \bibinfo{publisher}{Association for Computational Linguistics}, \bibinfo{address}{Online}, \bibinfo{year}{2021}, pp. \bibinfo{pages}{340--350}. \URLprefix \url{https://aclanthology.org/2021.acl-long.29}. \DOIprefix\doi{10.18653/v1/2021.acl-long.29}.
\bibitem[{Zhang et~al.(2021)Zhang, Deng, Li, Yuan, Bing, and Lam}]{zhang-etal-2021-aspect-sentiment}
\bibinfo{author}{W.~Zhang}, \bibinfo{author}{Y.~Deng}, \bibinfo{author}{X.~Li}, \bibinfo{author}{Y.~Yuan}, \bibinfo{author}{L.~Bing}, \bibinfo{author}{W.~Lam},
\newblock \bibinfo{title}{Aspect sentiment quad prediction as paraphrase generation},
\newblock in: \bibinfo{editor}{M.-F. Moens}, \bibinfo{editor}{X.~Huang}, \bibinfo{editor}{L.~Specia}, \bibinfo{editor}{S.~W.-t. Yih} (Eds.), \bibinfo{booktitle}{Proceedings of the 2021 Conference on Empirical Methods in Natural Language Processing}, \bibinfo{publisher}{Association for Computational Linguistics}, \bibinfo{address}{Online and Punta Cana, Dominican Republic}, \bibinfo{year}{2021}, pp. \bibinfo{pages}{9209--9219}. \URLprefix \url{https://aclanthology.org/2021.emnlp-main.726}. \DOIprefix\doi{10.18653/v1/2021.emnlp-main.726}.
\bibitem[{Steinberger et~al.(2014)Steinberger, Brychc{\'\i}n, and Konkol}]{steinberger-etal-2014-aspect}
\bibinfo{author}{J.~Steinberger}, \bibinfo{author}{T.~Brychc{\'\i}n}, \bibinfo{author}{M.~Konkol},
\newblock \bibinfo{title}{Aspect-level sentiment analysis in {C}zech},
\newblock in: \bibinfo{editor}{A.~Balahur}, \bibinfo{editor}{E.~van~der Goot}, \bibinfo{editor}{R.~Steinberger}, \bibinfo{editor}{A.~Montoyo} (Eds.), \bibinfo{booktitle}{Proceedings of the 5th Workshop on Computational Approaches to Subjectivity, Sentiment and Social Media Analysis}, \bibinfo{publisher}{Association for Computational Linguistics}, \bibinfo{address}{Baltimore, Maryland}, \bibinfo{year}{2014}, pp. \bibinfo{pages}{24--30}. \URLprefix \url{https://aclanthology.org/W14-2605}. \DOIprefix\doi{10.3115/v1/W14-2605}.
\bibitem[{Hercig et~al.(2016)Hercig, Brychc{\'\i}n, Svoboda, Konkol, and Steinberger}]{hercig2016unsupervised}
\bibinfo{author}{T.~Hercig}, \bibinfo{author}{T.~Brychc{\'\i}n}, \bibinfo{author}{L.~Svoboda}, \bibinfo{author}{M.~Konkol}, \bibinfo{author}{J.~Steinberger},
\newblock \bibinfo{title}{Unsupervised methods to improve aspect-based sentiment analysis in czech},
\newblock \bibinfo{journal}{Computaci{\'o}n y Sistemas} \bibinfo{volume}{20} (\bibinfo{year}{2016}) \bibinfo{pages}{365--375}.
\bibitem[{{\v{S}}m{\'\i}d et~al.(2024){\v{S}}m{\'\i}d, P{\v{r}}ib{\'a}{\v{n}}, Prazak, and Kral}]{smid-etal-2024-czech}
\bibinfo{author}{J.~{\v{S}}m{\'\i}d}, \bibinfo{author}{P.~P{\v{r}}ib{\'a}{\v{n}}}, \bibinfo{author}{O.~Prazak}, \bibinfo{author}{P.~Kral},
\newblock \bibinfo{title}{{C}zech dataset for complex aspect-based sentiment analysis tasks},
\newblock in: \bibinfo{editor}{N.~Calzolari}, \bibinfo{editor}{M.-Y. Kan}, \bibinfo{editor}{V.~Hoste}, \bibinfo{editor}{A.~Lenci}, \bibinfo{editor}{S.~Sakti}, \bibinfo{editor}{N.~Xue} (Eds.), \bibinfo{booktitle}{Proceedings of the 2024 Joint International Conference on Computational Linguistics, Language Resources and Evaluation (LREC-COLING 2024)}, \bibinfo{publisher}{ELRA and ICCL}, \bibinfo{address}{Torino, Italia}, \bibinfo{year}{2024}, pp. \bibinfo{pages}{4299--4310}. \URLprefix \url{https://aclanthology.org/2024.lrec-main.384}.
\bibitem[{Akhtar et~al.(2016)Akhtar, Ekbal, and Bhattacharyya}]{akhtar-etal-2016-aspect}
\bibinfo{author}{M.~S. Akhtar}, \bibinfo{author}{A.~Ekbal}, \bibinfo{author}{P.~Bhattacharyya},
\newblock \bibinfo{title}{Aspect based sentiment analysis in {H}indi: Resource creation and evaluation},
\newblock in: \bibinfo{editor}{N.~Calzolari}, \bibinfo{editor}{K.~Choukri}, \bibinfo{editor}{T.~Declerck}, \bibinfo{editor}{S.~Goggi}, \bibinfo{editor}{M.~Grobelnik}, \bibinfo{editor}{B.~Maegaard}, \bibinfo{editor}{J.~Mariani}, \bibinfo{editor}{H.~Mazo}, \bibinfo{editor}{A.~Moreno}, \bibinfo{editor}{J.~Odijk}, \bibinfo{editor}{S.~Piperidis} (Eds.), \bibinfo{booktitle}{Proceedings of the Tenth International Conference on Language Resources and Evaluation ({LREC}'16)}, \bibinfo{publisher}{European Language Resources Association (ELRA)}, \bibinfo{address}{Portoro{\v{z}}, Slovenia}, \bibinfo{year}{2016}, pp. \bibinfo{pages}{2703--2709}. \URLprefix \url{https://aclanthology.org/L16-1429}.
\bibitem[{Nakayama et~al.(2022)Nakayama, Murakami, Kumar, Bhingardive, and Hardaway}]{nakayama-etal-2022-large}
\bibinfo{author}{Y.~Nakayama}, \bibinfo{author}{K.~Murakami}, \bibinfo{author}{G.~Kumar}, \bibinfo{author}{S.~Bhingardive}, \bibinfo{author}{I.~Hardaway},
\newblock \bibinfo{title}{A large-scale {J}apanese dataset for aspect-based sentiment analysis},
\newblock in: \bibinfo{editor}{N.~Calzolari}, \bibinfo{editor}{F.~B{\'e}chet}, \bibinfo{editor}{P.~Blache}, \bibinfo{editor}{K.~Choukri}, \bibinfo{editor}{C.~Cieri}, \bibinfo{editor}{T.~Declerck}, \bibinfo{editor}{S.~Goggi}, \bibinfo{editor}{H.~Isahara}, \bibinfo{editor}{B.~Maegaard}, \bibinfo{editor}{J.~Mariani}, \bibinfo{editor}{H.~Mazo}, \bibinfo{editor}{J.~Odijk}, \bibinfo{editor}{S.~Piperidis} (Eds.), \bibinfo{booktitle}{Proceedings of the Thirteenth Language Resources and Evaluation Conference}, \bibinfo{publisher}{European Language Resources Association}, \bibinfo{address}{Marseille, France}, \bibinfo{year}{2022}, pp. \bibinfo{pages}{7014--7021}. \URLprefix \url{https://aclanthology.org/2022.lrec-1.758}.
\bibitem[{Wang et~al.(2019)Wang, Sun, Li, Liu, Si, Zhang, and Zhou}]{wang-etal-2019-aspect}
\bibinfo{author}{J.~Wang}, \bibinfo{author}{C.~Sun}, \bibinfo{author}{S.~Li}, \bibinfo{author}{X.~Liu}, \bibinfo{author}{L.~Si}, \bibinfo{author}{M.~Zhang}, \bibinfo{author}{G.~Zhou},
\newblock \bibinfo{title}{Aspect sentiment classification towards question-answering with reinforced bidirectional attention network},
\newblock in: \bibinfo{editor}{A.~Korhonen}, \bibinfo{editor}{D.~Traum}, \bibinfo{editor}{L.~M{\`a}rquez} (Eds.), \bibinfo{booktitle}{Proceedings of the 57th Annual Meeting of the Association for Computational Linguistics}, \bibinfo{publisher}{Association for Computational Linguistics}, \bibinfo{address}{Florence, Italy}, \bibinfo{year}{2019}, pp. \bibinfo{pages}{3548--3557}. \URLprefix \url{https://aclanthology.org/P19-1345}. \DOIprefix\doi{10.18653/v1/P19-1345}.
\bibitem[{Bu et~al.(2021)Bu, Ren, Zheng, Yang, Wang, Zhang, and Wu}]{bu-etal-2021-asap}
\bibinfo{author}{J.~Bu}, \bibinfo{author}{L.~Ren}, \bibinfo{author}{S.~Zheng}, \bibinfo{author}{Y.~Yang}, \bibinfo{author}{J.~Wang}, \bibinfo{author}{F.~Zhang}, \bibinfo{author}{W.~Wu},
\newblock \bibinfo{title}{{ASAP}: A {C}hinese review dataset towards aspect category sentiment analysis and rating prediction},
\newblock in: \bibinfo{editor}{K.~Toutanova}, \bibinfo{editor}{A.~Rumshisky}, \bibinfo{editor}{L.~Zettlemoyer}, \bibinfo{editor}{D.~Hakkani-Tur}, \bibinfo{editor}{I.~Beltagy}, \bibinfo{editor}{S.~Bethard}, \bibinfo{editor}{R.~Cotterell}, \bibinfo{editor}{T.~Chakraborty}, \bibinfo{editor}{Y.~Zhou} (Eds.), \bibinfo{booktitle}{Proceedings of the 2021 Conference of the North American Chapter of the Association for Computational Linguistics: Human Language Technologies}, \bibinfo{publisher}{Association for Computational Linguistics}, \bibinfo{address}{Online}, \bibinfo{year}{2021}, pp. \bibinfo{pages}{2069--2079}. \URLprefix \url{https://aclanthology.org/2021.naacl-main.167}. \DOIprefix\doi{10.18653/v1/2021.naacl-main.167}.
\bibitem[{Saeidi et~al.(2016)Saeidi, Bouchard, Liakata, and Riedel}]{saeidi-etal-2016-sentihood}
\bibinfo{author}{M.~Saeidi}, \bibinfo{author}{G.~Bouchard}, \bibinfo{author}{M.~Liakata}, \bibinfo{author}{S.~Riedel},
\newblock \bibinfo{title}{{S}enti{H}ood: Targeted aspect based sentiment analysis dataset for urban neighbourhoods},
\newblock in: \bibinfo{editor}{Y.~Matsumoto}, \bibinfo{editor}{R.~Prasad} (Eds.), \bibinfo{booktitle}{Proceedings of {COLING} 2016, the 26th International Conference on Computational Linguistics: Technical Papers}, \bibinfo{publisher}{The COLING 2016 Organizing Committee}, \bibinfo{address}{Osaka, Japan}, \bibinfo{year}{2016}, pp. \bibinfo{pages}{1546--1556}. \URLprefix \url{https://aclanthology.org/C16-1146}.
\bibitem[{Jiang et~al.(2019)Jiang, Chen, Xu, Ao, and Yang}]{jiang-etal-2019-challenge}
\bibinfo{author}{Q.~Jiang}, \bibinfo{author}{L.~Chen}, \bibinfo{author}{R.~Xu}, \bibinfo{author}{X.~Ao}, \bibinfo{author}{M.~Yang},
\newblock \bibinfo{title}{A challenge dataset and effective models for aspect-based sentiment analysis},
\newblock in: \bibinfo{editor}{K.~Inui}, \bibinfo{editor}{J.~Jiang}, \bibinfo{editor}{V.~Ng}, \bibinfo{editor}{X.~Wan} (Eds.), \bibinfo{booktitle}{Proceedings of the 2019 Conference on Empirical Methods in Natural Language Processing and the 9th International Joint Conference on Natural Language Processing (EMNLP-IJCNLP)}, \bibinfo{publisher}{Association for Computational Linguistics}, \bibinfo{address}{Hong Kong, China}, \bibinfo{year}{2019}, pp. \bibinfo{pages}{6280--6285}. \URLprefix \url{https://aclanthology.org/D19-1654}. \DOIprefix\doi{10.18653/v1/D19-1654}.
\bibitem[{Xing et~al.(2020)Xing, Jin, Jin, Wang, Zhang, and Huang}]{xing-etal-2020-tasty}
\bibinfo{author}{X.~Xing}, \bibinfo{author}{Z.~Jin}, \bibinfo{author}{D.~Jin}, \bibinfo{author}{B.~Wang}, \bibinfo{author}{Q.~Zhang}, \bibinfo{author}{X.~Huang},
\newblock \bibinfo{title}{Tasty burgers, soggy fries: Probing aspect robustness in aspect-based sentiment analysis},
\newblock in: \bibinfo{editor}{B.~Webber}, \bibinfo{editor}{T.~Cohn}, \bibinfo{editor}{Y.~He}, \bibinfo{editor}{Y.~Liu} (Eds.), \bibinfo{booktitle}{Proceedings of the 2020 Conference on Empirical Methods in Natural Language Processing (EMNLP)}, \bibinfo{publisher}{Association for Computational Linguistics}, \bibinfo{address}{Online}, \bibinfo{year}{2020}, pp. \bibinfo{pages}{3594--3605}. \URLprefix \url{https://aclanthology.org/2020.emnlp-main.292}. \DOIprefix\doi{10.18653/v1/2020.emnlp-main.292}.
\bibitem[{Fukushima(2007)}]{fukushima2007neocognitron}
\bibinfo{author}{K.~Fukushima},
\newblock \bibinfo{title}{Neocognitron},
\newblock \bibinfo{journal}{Scholarpedia} \bibinfo{volume}{2} (\bibinfo{year}{2007}) \bibinfo{pages}{1717}.
\bibitem[{Hochreiter(1997)}]{hochreiter1997long}
\bibinfo{author}{S.~Hochreiter},
\newblock \bibinfo{title}{Long short-term memory},
\newblock \bibinfo{journal}{Neural Computation MIT-Press}  (\bibinfo{year}{1997}).
\bibitem[{Vaswani et~al.(2017)Vaswani, Shazeer, Parmar, Uszkoreit, Jones, Gomez, Kaiser, and Polosukhin}]{vaswani2023attentionneed}
\bibinfo{author}{A.~Vaswani}, \bibinfo{author}{N.~Shazeer}, \bibinfo{author}{N.~Parmar}, \bibinfo{author}{J.~Uszkoreit}, \bibinfo{author}{L.~Jones}, \bibinfo{author}{A.~N. Gomez}, \bibinfo{author}{L.~u. Kaiser}, \bibinfo{author}{I.~Polosukhin},
\newblock \bibinfo{title}{Attention is all you need},
\newblock in: \bibinfo{editor}{I.~Guyon}, \bibinfo{editor}{U.~V. Luxburg}, \bibinfo{editor}{S.~Bengio}, \bibinfo{editor}{H.~Wallach}, \bibinfo{editor}{R.~Fergus}, \bibinfo{editor}{S.~Vishwanathan}, \bibinfo{editor}{R.~Garnett} (Eds.), \bibinfo{booktitle}{Advances in Neural Information Processing Systems}, volume~\bibinfo{volume}{30}, \bibinfo{publisher}{Curran Associates, Inc.}, \bibinfo{year}{2017}, p. \bibinfo{pages}{6000–6010}. \URLprefix \url{https://proceedings.neurips.cc/paper_files/paper/2017/file/3f5ee243547dee91fbd053c1c4a845aa-Paper.pdf}.
\bibitem[{Lafferty et~al.(2001)Lafferty, McCallum, and Pereira}]{lafferty2001conditional}
\bibinfo{author}{J.~D. Lafferty}, \bibinfo{author}{A.~McCallum}, \bibinfo{author}{F.~C.~N. Pereira},
\newblock \bibinfo{title}{Conditional random fields: Probabilistic models for segmenting and labeling sequence data},
\newblock in: \bibinfo{booktitle}{Proceedings of the Eighteenth International Conference on Machine Learning}, ICML '01, \bibinfo{publisher}{Morgan Kaufmann Publishers Inc.}, \bibinfo{address}{San Francisco, CA, USA}, \bibinfo{year}{2001}, p. \bibinfo{pages}{282–289}.
\bibitem[{Tjong Kim~Sang and Veenstra(1999)}]{tjong-kim-sang-veenstra-1999-representing}
\bibinfo{author}{E.~F. Tjong Kim~Sang}, \bibinfo{author}{J.~Veenstra},
\newblock \bibinfo{title}{Representing text chunks},
\newblock in: \bibinfo{editor}{H.~S. Thompson}, \bibinfo{editor}{A.~Lascarides} (Eds.), \bibinfo{booktitle}{Ninth Conference of the {E}uropean Chapter of the Association for Computational Linguistics}, \bibinfo{publisher}{Association for Computational Linguistics}, \bibinfo{address}{Bergen, Norway}, \bibinfo{year}{1999}, pp. \bibinfo{pages}{173--179}. \URLprefix \url{https://aclanthology.org/E99-1023}.
\bibitem[{Zhang et~al.(2021)Zhang, Li, Deng, Bing, and Lam}]{zhang-etal-2021-towards-generative}
\bibinfo{author}{W.~Zhang}, \bibinfo{author}{X.~Li}, \bibinfo{author}{Y.~Deng}, \bibinfo{author}{L.~Bing}, \bibinfo{author}{W.~Lam},
\newblock \bibinfo{title}{Towards generative aspect-based sentiment analysis},
\newblock in: \bibinfo{editor}{C.~Zong}, \bibinfo{editor}{F.~Xia}, \bibinfo{editor}{W.~Li}, \bibinfo{editor}{R.~Navigli} (Eds.), \bibinfo{booktitle}{Proceedings of the 59th Annual Meeting of the Association for Computational Linguistics and the 11th International Joint Conference on Natural Language Processing (Volume 2: Short Papers)}, \bibinfo{publisher}{Association for Computational Linguistics}, \bibinfo{address}{Online}, \bibinfo{year}{2021}, pp. \bibinfo{pages}{504--510}. \URLprefix \url{https://aclanthology.org/2021.acl-short.64}. \DOIprefix\doi{10.18653/v1/2021.acl-short.64}.
\bibitem[{Mao et~al.(2022)Mao, Shen, Yang, Zhu, and Cai}]{mao-etal-2022-seq2path}
\bibinfo{author}{Y.~Mao}, \bibinfo{author}{Y.~Shen}, \bibinfo{author}{J.~Yang}, \bibinfo{author}{X.~Zhu}, \bibinfo{author}{L.~Cai},
\newblock \bibinfo{title}{{S}eq2{P}ath: Generating sentiment tuples as paths of a tree},
\newblock in: \bibinfo{editor}{S.~Muresan}, \bibinfo{editor}{P.~Nakov}, \bibinfo{editor}{A.~Villavicencio} (Eds.), \bibinfo{booktitle}{Findings of the Association for Computational Linguistics: ACL 2022}, \bibinfo{publisher}{Association for Computational Linguistics}, \bibinfo{address}{Dublin, Ireland}, \bibinfo{year}{2022}, pp. \bibinfo{pages}{2215--2225}. \URLprefix \url{https://aclanthology.org/2022.findings-acl.174}. \DOIprefix\doi{10.18653/v1/2022.findings-acl.174}.
\bibitem[{Gou et~al.(2023)Gou, Guo, and Yang}]{gou-etal-2023-mvp}
\bibinfo{author}{Z.~Gou}, \bibinfo{author}{Q.~Guo}, \bibinfo{author}{Y.~Yang},
\newblock \bibinfo{title}{{M}v{P}: Multi-view prompting improves aspect sentiment tuple prediction},
\newblock in: \bibinfo{editor}{A.~Rogers}, \bibinfo{editor}{J.~Boyd-Graber}, \bibinfo{editor}{N.~Okazaki} (Eds.), \bibinfo{booktitle}{Proceedings of the 61st Annual Meeting of the Association for Computational Linguistics (Volume 1: Long Papers)}, \bibinfo{publisher}{Association for Computational Linguistics}, \bibinfo{address}{Toronto, Canada}, \bibinfo{year}{2023}, pp. \bibinfo{pages}{4380--4397}. \URLprefix \url{https://aclanthology.org/2023.acl-long.240}. \DOIprefix\doi{10.18653/v1/2023.acl-long.240}.
\bibitem[{Xianlong et~al.(2023)Xianlong, Yang, and Wang}]{xianlong-etal-2023-tagging}
\bibinfo{author}{L.~Xianlong}, \bibinfo{author}{M.~Yang}, \bibinfo{author}{Y.~Wang},
\newblock \bibinfo{title}{Tagging-assisted generation model with encoder and decoder supervision for aspect sentiment triplet extraction},
\newblock in: \bibinfo{editor}{H.~Bouamor}, \bibinfo{editor}{J.~Pino}, \bibinfo{editor}{K.~Bali} (Eds.), \bibinfo{booktitle}{Proceedings of the 2023 Conference on Empirical Methods in Natural Language Processing}, \bibinfo{publisher}{Association for Computational Linguistics}, \bibinfo{address}{Singapore}, \bibinfo{year}{2023}, pp. \bibinfo{pages}{2078--2093}. \URLprefix \url{https://aclanthology.org/2023.emnlp-main.129}. \DOIprefix\doi{10.18653/v1/2023.emnlp-main.129}.
\bibitem[{{\v{S}}m{\'\i}d and P{\v{r}}ib{\'a}{\v{n}}(2023)}]{smid-priban-2023-prompt}
\bibinfo{author}{J.~{\v{S}}m{\'\i}d}, \bibinfo{author}{P.~P{\v{r}}ib{\'a}{\v{n}}},
\newblock \bibinfo{title}{Prompt-based approach for {C}zech sentiment analysis},
\newblock in: \bibinfo{editor}{R.~Mitkov}, \bibinfo{editor}{G.~Angelova} (Eds.), \bibinfo{booktitle}{Proceedings of the 14th International Conference on Recent Advances in Natural Language Processing}, \bibinfo{publisher}{INCOMA Ltd., Shoumen, Bulgaria}, \bibinfo{address}{Varna, Bulgaria}, \bibinfo{year}{2023}, pp. \bibinfo{pages}{1110--1120}. \URLprefix \url{https://aclanthology.org/2023.ranlp-1.118}.
\bibitem[{Zhang et~al.(2024)Zhang, Deng, Liu, Pan, and Bing}]{zhang-etal-2024-sentiment}
\bibinfo{author}{W.~Zhang}, \bibinfo{author}{Y.~Deng}, \bibinfo{author}{B.~Liu}, \bibinfo{author}{S.~Pan}, \bibinfo{author}{L.~Bing},
\newblock \bibinfo{title}{Sentiment analysis in the era of large language models: A reality check},
\newblock in: \bibinfo{editor}{K.~Duh}, \bibinfo{editor}{H.~Gomez}, \bibinfo{editor}{S.~Bethard} (Eds.), \bibinfo{booktitle}{Findings of the Association for Computational Linguistics: NAACL 2024}, \bibinfo{publisher}{Association for Computational Linguistics}, \bibinfo{address}{Mexico City, Mexico}, \bibinfo{year}{2024}, pp. \bibinfo{pages}{3881--3906}. \URLprefix \url{https://aclanthology.org/2024.findings-naacl.246}. \DOIprefix\doi{10.18653/v1/2024.findings-naacl.246}.
\bibitem[{{\v{S}}m{\'\i}d et~al.(2024){\v{S}}m{\'\i}d, Priban, and Kral}]{smid-etal-2024-llama}
\bibinfo{author}{J.~{\v{S}}m{\'\i}d}, \bibinfo{author}{P.~Priban}, \bibinfo{author}{P.~Kral},
\newblock \bibinfo{title}{{LL}a{MA}-based models for aspect-based sentiment analysis},
\newblock in: \bibinfo{editor}{O.~De~Clercq}, \bibinfo{editor}{V.~Barriere}, \bibinfo{editor}{J.~Barnes}, \bibinfo{editor}{R.~Klinger}, \bibinfo{editor}{J.~Sedoc}, \bibinfo{editor}{S.~Tafreshi} (Eds.), \bibinfo{booktitle}{Proceedings of the 14th Workshop on Computational Approaches to Subjectivity, Sentiment, {\&} Social Media Analysis}, \bibinfo{publisher}{Association for Computational Linguistics}, \bibinfo{address}{Bangkok, Thailand}, \bibinfo{year}{2024}, pp. \bibinfo{pages}{63--70}. \URLprefix \url{https://aclanthology.org/2024.wassa-1.6}.
\bibitem[{Mikolov et~al.(2013)Mikolov, Sutskever, Chen, Corrado, and Dean}]{word2vec}
\bibinfo{author}{T.~Mikolov}, \bibinfo{author}{I.~Sutskever}, \bibinfo{author}{K.~Chen}, \bibinfo{author}{G.~S. Corrado}, \bibinfo{author}{J.~Dean},
\newblock \bibinfo{title}{Distributed representations of words and phrases and their compositionality},
\newblock in: \bibinfo{editor}{C.~Burges}, \bibinfo{editor}{L.~Bottou}, \bibinfo{editor}{M.~Welling}, \bibinfo{editor}{Z.~Ghahramani}, \bibinfo{editor}{K.~Weinberger} (Eds.), \bibinfo{booktitle}{Advances in Neural Information Processing Systems}, volume~\bibinfo{volume}{26}, \bibinfo{publisher}{Curran Associates, Inc.}, \bibinfo{year}{2013}, p. \bibinfo{pages}{3111–3119}. \URLprefix \url{https://proceedings.neurips.cc/paper_files/paper/2013/file/9aa42b31882ec039965f3c4923ce901b-Paper.pdf}.
\bibitem[{Bojanowski et~al.(2017)Bojanowski, Grave, Joulin, and Mikolov}]{bojanowski-etal-2017-enriching}
\bibinfo{author}{P.~Bojanowski}, \bibinfo{author}{E.~Grave}, \bibinfo{author}{A.~Joulin}, \bibinfo{author}{T.~Mikolov},
\newblock \bibinfo{title}{Enriching word vectors with subword information},
\newblock \bibinfo{journal}{Transactions of the Association for Computational Linguistics} \bibinfo{volume}{5} (\bibinfo{year}{2017}) \bibinfo{pages}{135--146}.
\bibitem[{Pennington et~al.(2014)Pennington, Socher, and Manning}]{pennington-etal-2014-glove}
\bibinfo{author}{J.~Pennington}, \bibinfo{author}{R.~Socher}, \bibinfo{author}{C.~Manning},
\newblock \bibinfo{title}{{G}lo{V}e: Global vectors for word representation},
\newblock in: \bibinfo{editor}{A.~Moschitti}, \bibinfo{editor}{B.~Pang}, \bibinfo{editor}{W.~Daelemans} (Eds.), \bibinfo{booktitle}{Proceedings of the 2014 Conference on Empirical Methods in Natural Language Processing ({EMNLP})}, \bibinfo{publisher}{Association for Computational Linguistics}, \bibinfo{address}{Doha, Qatar}, \bibinfo{year}{2014}, pp. \bibinfo{pages}{1532--1543}. \URLprefix \url{https://aclanthology.org/D14-1162}. \DOIprefix\doi{10.3115/v1/D14-1162}.
\bibitem[{Luong et~al.(2015)Luong, Pham, and Manning}]{luong-etal-2015-bilingual}
\bibinfo{author}{T.~Luong}, \bibinfo{author}{H.~Pham}, \bibinfo{author}{C.~D. Manning},
\newblock \bibinfo{title}{Bilingual word representations with monolingual quality in mind},
\newblock in: \bibinfo{editor}{P.~Blunsom}, \bibinfo{editor}{S.~Cohen}, \bibinfo{editor}{P.~Dhillon}, \bibinfo{editor}{P.~Liang} (Eds.), \bibinfo{booktitle}{Proceedings of the 1st Workshop on Vector Space Modeling for Natural Language Processing}, \bibinfo{publisher}{Association for Computational Linguistics}, \bibinfo{address}{Denver, Colorado}, \bibinfo{year}{2015}, pp. \bibinfo{pages}{151--159}. \URLprefix \url{https://aclanthology.org/W15-1521}. \DOIprefix\doi{10.3115/v1/W15-1521}.
\bibitem[{Koehn(2005)}]{koehn-2005-europarl}
\bibinfo{author}{P.~Koehn},
\newblock \bibinfo{title}{{E}uroparl: A parallel corpus for statistical machine translation},
\newblock in: \bibinfo{booktitle}{Proceedings of Machine Translation Summit X: Papers}, \bibinfo{address}{Phuket, Thailand}, \bibinfo{year}{2005}, pp. \bibinfo{pages}{79--86}. \URLprefix \url{https://aclanthology.org/2005.mtsummit-papers.11}.
\bibitem[{Ruder et~al.(2019)Ruder, Vuli\'{c}, and S\o{}gaard}]{10.1613/jair.1.11640}
\bibinfo{author}{S.~Ruder}, \bibinfo{author}{I.~Vuli\'{c}}, \bibinfo{author}{A.~S\o{}gaard},
\newblock \bibinfo{title}{A survey of cross-lingual word embedding models},
\newblock \bibinfo{journal}{J. Artif. Int. Res.} \bibinfo{volume}{65} (\bibinfo{year}{2019}) \bibinfo{pages}{569–630}.
\bibitem[{Dyer et~al.(2013)Dyer, Chahuneau, and Smith}]{dyer-etal-2013-simple}
\bibinfo{author}{C.~Dyer}, \bibinfo{author}{V.~Chahuneau}, \bibinfo{author}{N.~A. Smith},
\newblock \bibinfo{title}{A simple, fast, and effective reparameterization of {IBM} model 2},
\newblock in: \bibinfo{editor}{L.~Vanderwende}, \bibinfo{editor}{H.~Daum{\'e}~III}, \bibinfo{editor}{K.~Kirchhoff} (Eds.), \bibinfo{booktitle}{Proceedings of the 2013 Conference of the North {A}merican Chapter of the Association for Computational Linguistics: Human Language Technologies}, \bibinfo{publisher}{Association for Computational Linguistics}, \bibinfo{address}{Atlanta, Georgia}, \bibinfo{year}{2013}, pp. \bibinfo{pages}{644--648}. \URLprefix \url{https://aclanthology.org/N13-1073}.
\bibitem[{Koehn(2010)}]{31b163dba28a47d2955cacf8d8908006}
\bibinfo{author}{P.~Koehn},
\newblock \bibinfo{title}{An experimental management system},
\newblock \bibinfo{journal}{Prague Bulletin of Mathematical Linguistics} \bibinfo{volume}{94} (\bibinfo{year}{2010}) \bibinfo{pages}{87--96}.
\bibitem[{Xue et~al.(2021)Xue, Constant, Roberts, Kale, Al-Rfou, Siddhant, Barua, and Raffel}]{xue-etal-2021-mt5}
\bibinfo{author}{L.~Xue}, \bibinfo{author}{N.~Constant}, \bibinfo{author}{A.~Roberts}, \bibinfo{author}{M.~Kale}, \bibinfo{author}{R.~Al-Rfou}, \bibinfo{author}{A.~Siddhant}, \bibinfo{author}{A.~Barua}, \bibinfo{author}{C.~Raffel},
\newblock \bibinfo{title}{m{T}5: A massively multilingual pre-trained text-to-text transformer},
\newblock in: \bibinfo{editor}{K.~Toutanova}, \bibinfo{editor}{A.~Rumshisky}, \bibinfo{editor}{L.~Zettlemoyer}, \bibinfo{editor}{D.~Hakkani-Tur}, \bibinfo{editor}{I.~Beltagy}, \bibinfo{editor}{S.~Bethard}, \bibinfo{editor}{R.~Cotterell}, \bibinfo{editor}{T.~Chakraborty}, \bibinfo{editor}{Y.~Zhou} (Eds.), \bibinfo{booktitle}{Proceedings of the 2021 Conference of the North American Chapter of the Association for Computational Linguistics: Human Language Technologies}, \bibinfo{publisher}{Association for Computational Linguistics}, \bibinfo{address}{Online}, \bibinfo{year}{2021}, pp. \bibinfo{pages}{483--498}. \URLprefix \url{https://aclanthology.org/2021.naacl-main.41}. \DOIprefix\doi{10.18653/v1/2021.naacl-main.41}.
\bibitem[{Wu and Dredze(2019)}]{wu-dredze-2019-beto}
\bibinfo{author}{S.~Wu}, \bibinfo{author}{M.~Dredze},
\newblock \bibinfo{title}{Beto, bentz, becas: The surprising cross-lingual effectiveness of {BERT}},
\newblock in: \bibinfo{editor}{K.~Inui}, \bibinfo{editor}{J.~Jiang}, \bibinfo{editor}{V.~Ng}, \bibinfo{editor}{X.~Wan} (Eds.), \bibinfo{booktitle}{Proceedings of the 2019 Conference on Empirical Methods in Natural Language Processing and the 9th International Joint Conference on Natural Language Processing (EMNLP-IJCNLP)}, \bibinfo{publisher}{Association for Computational Linguistics}, \bibinfo{address}{Hong Kong, China}, \bibinfo{year}{2019}, pp. \bibinfo{pages}{833--844}. \URLprefix \url{https://aclanthology.org/D19-1077}. \DOIprefix\doi{10.18653/v1/D19-1077}.
\bibitem[{Wang et~al.(2019)Wang, Mayhew, Roth et~al.}]{wang2019cross}
\bibinfo{author}{Z.~Wang}, \bibinfo{author}{S.~Mayhew}, \bibinfo{author}{D.~Roth}, et~al.,
\newblock \bibinfo{title}{Cross-lingual ability of multilingual bert: An empirical study},
\newblock \bibinfo{journal}{arXiv preprint arXiv:1912.07840}  (\bibinfo{year}{2019}).
\bibitem[{Pires et~al.(2019)Pires, Schlinger, and Garrette}]{pires-etal-2019-multilingual}
\bibinfo{author}{T.~Pires}, \bibinfo{author}{E.~Schlinger}, \bibinfo{author}{D.~Garrette},
\newblock \bibinfo{title}{How multilingual is multilingual {BERT}?},
\newblock in: \bibinfo{editor}{A.~Korhonen}, \bibinfo{editor}{D.~Traum}, \bibinfo{editor}{L.~M{\`a}rquez} (Eds.), \bibinfo{booktitle}{Proceedings of the 57th Annual Meeting of the Association for Computational Linguistics}, \bibinfo{publisher}{Association for Computational Linguistics}, \bibinfo{address}{Florence, Italy}, \bibinfo{year}{2019}, pp. \bibinfo{pages}{4996--5001}. \URLprefix \url{https://aclanthology.org/P19-1493}. \DOIprefix\doi{10.18653/v1/P19-1493}.
\bibitem[{Feng et~al.(2022)Feng, Yang, Cer, Arivazhagan, and Wang}]{feng-etal-2022-language}
\bibinfo{author}{F.~Feng}, \bibinfo{author}{Y.~Yang}, \bibinfo{author}{D.~Cer}, \bibinfo{author}{N.~Arivazhagan}, \bibinfo{author}{W.~Wang},
\newblock \bibinfo{title}{Language-agnostic {BERT} sentence embedding},
\newblock in: \bibinfo{editor}{S.~Muresan}, \bibinfo{editor}{P.~Nakov}, \bibinfo{editor}{A.~Villavicencio} (Eds.), \bibinfo{booktitle}{Proceedings of the 60th Annual Meeting of the Association for Computational Linguistics (Volume 1: Long Papers)}, \bibinfo{publisher}{Association for Computational Linguistics}, \bibinfo{address}{Dublin, Ireland}, \bibinfo{year}{2022}, pp. \bibinfo{pages}{878--891}. \URLprefix \url{https://aclanthology.org/2022.acl-long.62/}. \DOIprefix\doi{10.18653/v1/2022.acl-long.62}.
\bibitem[{Pfeiffer et~al.(2020)Pfeiffer, Vuli{\'c}, Gurevych, and Ruder}]{pfeiffer-etal-2020-mad}
\bibinfo{author}{J.~Pfeiffer}, \bibinfo{author}{I.~Vuli{\'c}}, \bibinfo{author}{I.~Gurevych}, \bibinfo{author}{S.~Ruder},
\newblock \bibinfo{title}{{MAD-X}: {A}n {A}dapter-{B}ased {F}ramework for {M}ulti-{T}ask {C}ross-{L}ingual {T}ransfer},
\newblock in: \bibinfo{editor}{B.~Webber}, \bibinfo{editor}{T.~Cohn}, \bibinfo{editor}{Y.~He}, \bibinfo{editor}{Y.~Liu} (Eds.), \bibinfo{booktitle}{Proceedings of the 2020 Conference on Empirical Methods in Natural Language Processing (EMNLP)}, \bibinfo{publisher}{Association for Computational Linguistics}, \bibinfo{address}{Online}, \bibinfo{year}{2020}, pp. \bibinfo{pages}{7654--7673}. \URLprefix \url{https://aclanthology.org/2020.emnlp-main.617}. \DOIprefix\doi{10.18653/v1/2020.emnlp-main.617}.
\bibitem[{Ma et~al.(2018)Ma, Peng, and Cambria}]{Ma_Peng_Cambria_2018}
\bibinfo{author}{Y.~Ma}, \bibinfo{author}{H.~Peng}, \bibinfo{author}{E.~Cambria},
\newblock \bibinfo{title}{Targeted aspect-based sentiment analysis via embedding commonsense knowledge into an attentive lstm},
\newblock \bibinfo{journal}{Proceedings of the AAAI Conference on Artificial Intelligence} \bibinfo{volume}{32} (\bibinfo{year}{2018}).
\bibitem[{Liang et~al.(2022)Liang, Su, Gui, Cambria, and Xu}]{LIANG2022107643}
\bibinfo{author}{B.~Liang}, \bibinfo{author}{H.~Su}, \bibinfo{author}{L.~Gui}, \bibinfo{author}{E.~Cambria}, \bibinfo{author}{R.~Xu},
\newblock \bibinfo{title}{Aspect-based sentiment analysis via affective knowledge enhanced graph convolutional networks},
\newblock \bibinfo{journal}{Knowledge-Based Systems} \bibinfo{volume}{235} (\bibinfo{year}{2022}) \bibinfo{pages}{107643}.
\bibitem[{D’Aniello et~al.(2022)D’Aniello, Gaeta, and La~Rocca}]{d2022knowmis}
\bibinfo{author}{G.~D’Aniello}, \bibinfo{author}{M.~Gaeta}, \bibinfo{author}{I.~La~Rocca},
\newblock \bibinfo{title}{Knowmis-absa: an overview and a reference model for applications of sentiment analysis and aspect-based sentiment analysis},
\newblock \bibinfo{journal}{Artificial Intelligence Review} \bibinfo{volume}{55} (\bibinfo{year}{2022}) \bibinfo{pages}{5543--5574}.
\bibitem[{Gao et~al.(2022)Gao, Fang, Liu, Liu, Liu, Liu, Bao, and Yan}]{gao-etal-2022-lego}
\bibinfo{author}{T.~Gao}, \bibinfo{author}{J.~Fang}, \bibinfo{author}{H.~Liu}, \bibinfo{author}{Z.~Liu}, \bibinfo{author}{C.~Liu}, \bibinfo{author}{P.~Liu}, \bibinfo{author}{Y.~Bao}, \bibinfo{author}{W.~Yan},
\newblock \bibinfo{title}{{LEGO}-{ABSA}: A prompt-based task assemblable unified generative framework for multi-task aspect-based sentiment analysis},
\newblock in: \bibinfo{editor}{N.~Calzolari}, \bibinfo{editor}{C.-R. Huang}, \bibinfo{editor}{H.~Kim}, \bibinfo{editor}{J.~Pustejovsky}, \bibinfo{editor}{L.~Wanner}, \bibinfo{editor}{K.-S. Choi}, \bibinfo{editor}{P.-M. Ryu}, \bibinfo{editor}{H.-H. Chen}, \bibinfo{editor}{L.~Donatelli}, \bibinfo{editor}{H.~Ji}, \bibinfo{editor}{S.~Kurohashi}, \bibinfo{editor}{P.~Paggio}, \bibinfo{editor}{N.~Xue}, \bibinfo{editor}{S.~Kim}, \bibinfo{editor}{Y.~Hahm}, \bibinfo{editor}{Z.~He}, \bibinfo{editor}{T.~K. Lee}, \bibinfo{editor}{E.~Santus}, \bibinfo{editor}{F.~Bond}, \bibinfo{editor}{S.-H. Na} (Eds.), \bibinfo{booktitle}{Proceedings of the 29th International Conference on Computational Linguistics}, \bibinfo{publisher}{International Committee on Computational Linguistics}, \bibinfo{address}{Gyeongju, Republic of Korea}, \bibinfo{year}{2022}, pp. \bibinfo{pages}{7002--7012}. \URLprefix \url{https://aclanthology.org/2022.coling-1.610}.
\bibitem[{Hu et~al.(2022)Hu, Wu, Gao, Bai, and Zhao}]{hu-etal-2022-improving-aspect}
\bibinfo{author}{M.~Hu}, \bibinfo{author}{Y.~Wu}, \bibinfo{author}{H.~Gao}, \bibinfo{author}{Y.~Bai}, \bibinfo{author}{S.~Zhao},
\newblock \bibinfo{title}{Improving aspect sentiment quad prediction via template-order data augmentation},
\newblock in: \bibinfo{booktitle}{Proceedings of the 2022 Conference on Empirical Methods in Natural Language Processing}, \bibinfo{publisher}{Association for Computational Linguistics}, \bibinfo{address}{Abu Dhabi, United Arab Emirates}, \bibinfo{year}{2022}, pp. \bibinfo{pages}{7889--7900}. \URLprefix \url{https://aclanthology.org/2022.emnlp-main.538}. \DOIprefix\doi{10.18653/v1/2022.emnlp-main.538}.
\bibitem[{Tamchyna et~al.(2015)Tamchyna, Fiala, and Veselovsk{\'a}}]{tamchyna2015czech}
\bibinfo{author}{A.~Tamchyna}, \bibinfo{author}{O.~Fiala}, \bibinfo{author}{K.~Veselovsk{\'a}},
\newblock \bibinfo{title}{Czech aspect-based sentiment analysis: A new dataset and preliminary results.},
\newblock in: \bibinfo{booktitle}{ITAT}, \bibinfo{year}{2015}, pp. \bibinfo{pages}{95--99}.
\bibitem[{Lenc and Hercig(2016)}]{SLON-Lenc2016Neural}
\bibinfo{author}{L.~Lenc}, \bibinfo{author}{T.~Hercig},
\newblock \bibinfo{title}{Neural networks for sentiment analysis in czech},
\newblock in: \bibinfo{editor}{B.~Brejov{\'{a}}} (Ed.), \bibinfo{booktitle}{Proceedings of the 16th {ITAT}: Slovensko{\v{c}}esk{\'{y}} {NLP} workshop (Slo{NLP} 2016)}, volume \bibinfo{volume}{1649} of \textit{\bibinfo{series}{{CEUR} Workshop Proceedings}}, \bibinfo{organization}{Comenius University in Bratislava, Faculty of Mathematics, Physics and Informatics}, \bibinfo{publisher}{CreateSpace Independent Publishing Platform}, \bibinfo{address}{Bratislava, Slovakia}, \bibinfo{year}{2016}, pp. \bibinfo{pages}{48--55}.
\bibitem[{P{\v{r}}ib{\'a}{\v{n}} and Pra\v{z}{\'a}k(2023)}]{priban-prazak-2023-improving}
\bibinfo{author}{P.~P{\v{r}}ib{\'a}{\v{n}}}, \bibinfo{author}{O.~Pra\v{z}{\'a}k},
\newblock \bibinfo{title}{Improving aspect-based sentiment with end-to-end semantic role labeling model},
\newblock in: \bibinfo{editor}{R.~Mitkov}, \bibinfo{editor}{G.~Angelova} (Eds.), \bibinfo{booktitle}{Proceedings of the 14th International Conference on Recent Advances in Natural Language Processing}, \bibinfo{publisher}{INCOMA Ltd., Shoumen, Bulgaria}, \bibinfo{address}{Varna, Bulgaria}, \bibinfo{year}{2023}, pp. \bibinfo{pages}{888--897}. \URLprefix \url{https://aclanthology.org/2023.ranlp-1.96}.
\bibitem[{García-Pablos et~al.(2018)García-Pablos, Cuadros, and Rigau}]{GARCIAPABLOS2018127}
\bibinfo{author}{A.~García-Pablos}, \bibinfo{author}{M.~Cuadros}, \bibinfo{author}{G.~Rigau},
\newblock \bibinfo{title}{W2vlda: Almost unsupervised system for aspect based sentiment analysis},
\newblock \bibinfo{journal}{Expert Systems with Applications} \bibinfo{volume}{91} (\bibinfo{year}{2018}) \bibinfo{pages}{127--137}.
\bibitem[{Blei et~al.(2003)Blei, Ng, and Jordan}]{10.5555/944919.944937}
\bibinfo{author}{D.~M. Blei}, \bibinfo{author}{A.~Y. Ng}, \bibinfo{author}{M.~I. Jordan},
\newblock \bibinfo{title}{Latent dirichlet allocation},
\newblock \bibinfo{journal}{J. Mach. Learn. Res.} \bibinfo{volume}{3} (\bibinfo{year}{2003}) \bibinfo{pages}{993–1022}.
\bibitem[{He et~al.(2022)He, Wumaier, Kadeer, Sun, Xin, and Zheng}]{9852228}
\bibinfo{author}{J.~He}, \bibinfo{author}{A.~Wumaier}, \bibinfo{author}{Z.~Kadeer}, \bibinfo{author}{W.~Sun}, \bibinfo{author}{X.~Xin}, \bibinfo{author}{L.~Zheng},
\newblock \bibinfo{title}{A local and global context focus multilingual learning model for aspect-based sentiment analysis},
\newblock \bibinfo{journal}{IEEE Access} \bibinfo{volume}{10} (\bibinfo{year}{2022}) \bibinfo{pages}{84135--84146}.
\bibitem[{Chung et~al.(2014)Chung, Gulcehre, Cho, and Bengio}]{69e088c8129341ac89810907fe6b1bfe}
\bibinfo{author}{J.~Chung}, \bibinfo{author}{C.~Gulcehre}, \bibinfo{author}{K.~Cho}, \bibinfo{author}{Y.~Bengio},
\newblock \bibinfo{title}{Empirical evaluation of gated recurrent neural networks on sequence modeling},
\newblock in: \bibinfo{booktitle}{NIPS 2014 Workshop on Deep Learning, December 2014}, \bibinfo{year}{2014}.
\bibitem[{Rana et~al.(2024)Rana, Ali, and Nawaz}]{rana2024exploring}
\bibinfo{author}{M.~R.~R. Rana}, \bibinfo{author}{T.~Ali}, \bibinfo{author}{A.~Nawaz},
\newblock \bibinfo{title}{Exploring multilingual reviews for aspect-based sentiment analysis using lexicon and bert},
\newblock \bibinfo{journal}{Sigma Journal of Engineering and Natural Sciences} \bibinfo{volume}{42} (\bibinfo{year}{2024}) \bibinfo{pages}{1469--1479}.
\bibitem[{Wawer(2024)}]{info15110664}
\bibinfo{author}{A.~Wawer},
\newblock \bibinfo{title}{Few-shot methods for aspect-level sentiment analysis},
\newblock \bibinfo{journal}{Information} \bibinfo{volume}{15} (\bibinfo{year}{2024}).
\bibitem[{Snell et~al.(2017)Snell, Swersky, and Zemel}]{10.5555/3294996.3295163}
\bibinfo{author}{J.~Snell}, \bibinfo{author}{K.~Swersky}, \bibinfo{author}{R.~Zemel},
\newblock \bibinfo{title}{Prototypical networks for few-shot learning},
\newblock in: \bibinfo{booktitle}{Proceedings of the 31st International Conference on Neural Information Processing Systems}, NIPS'17, \bibinfo{publisher}{Curran Associates Inc.}, \bibinfo{address}{Red Hook, NY, USA}, \bibinfo{year}{2017}, p. \bibinfo{pages}{4080–4090}.
\bibitem[{Yang and Katiyar(2020)}]{yang-katiyar-2020-simple}
\bibinfo{author}{Y.~Yang}, \bibinfo{author}{A.~Katiyar},
\newblock \bibinfo{title}{Simple and effective few-shot named entity recognition with structured nearest neighbor learning},
\newblock in: \bibinfo{editor}{B.~Webber}, \bibinfo{editor}{T.~Cohn}, \bibinfo{editor}{Y.~He}, \bibinfo{editor}{Y.~Liu} (Eds.), \bibinfo{booktitle}{Proceedings of the 2020 Conference on Empirical Methods in Natural Language Processing (EMNLP)}, \bibinfo{publisher}{Association for Computational Linguistics}, \bibinfo{address}{Online}, \bibinfo{year}{2020}, pp. \bibinfo{pages}{6365--6375}. \URLprefix \url{https://aclanthology.org/2020.emnlp-main.516/}. \DOIprefix\doi{10.18653/v1/2020.emnlp-main.516}.
\bibitem[{Lim et~al.(2024)Lim, Yun, Kim, Choi, and Kim}]{lim-etal-2024-analysis}
\bibinfo{author}{S.~Lim}, \bibinfo{author}{T.~Yun}, \bibinfo{author}{J.~Kim}, \bibinfo{author}{J.~Choi}, \bibinfo{author}{T.~Kim},
\newblock \bibinfo{title}{Analysis of multi-source language training in cross-lingual transfer},
\newblock in: \bibinfo{editor}{L.-W. Ku}, \bibinfo{editor}{A.~Martins}, \bibinfo{editor}{V.~Srikumar} (Eds.), \bibinfo{booktitle}{Proceedings of the 62nd Annual Meeting of the Association for Computational Linguistics (Volume 1: Long Papers)}, \bibinfo{publisher}{Association for Computational Linguistics}, \bibinfo{address}{Bangkok, Thailand}, \bibinfo{year}{2024}, pp. \bibinfo{pages}{712--725}. \URLprefix \url{https://aclanthology.org/2024.acl-long.42/}. \DOIprefix\doi{10.18653/v1/2024.acl-long.42}.
\bibitem[{Wan(2008)}]{wan-2008-using}
\bibinfo{author}{X.~Wan},
\newblock \bibinfo{title}{Using bilingual knowledge and ensemble techniques for unsupervised {C}hinese sentiment analysis},
\newblock in: \bibinfo{editor}{M.~Lapata}, \bibinfo{editor}{H.~T. Ng} (Eds.), \bibinfo{booktitle}{Proceedings of the 2008 Conference on Empirical Methods in Natural Language Processing}, \bibinfo{publisher}{Association for Computational Linguistics}, \bibinfo{address}{Honolulu, Hawaii}, \bibinfo{year}{2008}, pp. \bibinfo{pages}{553--561}. \URLprefix \url{https://aclanthology.org/D08-1058}.
\bibitem[{Wan(2009)}]{wan-2009-co}
\bibinfo{author}{X.~Wan},
\newblock \bibinfo{title}{Co-training for cross-lingual sentiment classification},
\newblock in: \bibinfo{editor}{K.-Y. Su}, \bibinfo{editor}{J.~Su}, \bibinfo{editor}{J.~Wiebe}, \bibinfo{editor}{H.~Li} (Eds.), \bibinfo{booktitle}{Proceedings of the Joint Conference of the 47th Annual Meeting of the {ACL} and the 4th International Joint Conference on Natural Language Processing of the {AFNLP}}, \bibinfo{publisher}{Association for Computational Linguistics}, \bibinfo{address}{Suntec, Singapore}, \bibinfo{year}{2009}, pp. \bibinfo{pages}{235--243}. \URLprefix \url{https://aclanthology.org/P09-1027}.
\bibitem[{Balahur and Turchi(2012)}]{balahur-turchi-2012-multilingual}
\bibinfo{author}{A.~Balahur}, \bibinfo{author}{M.~Turchi},
\newblock \bibinfo{title}{Multilingual sentiment analysis using machine translation?},
\newblock in: \bibinfo{editor}{A.~Balahur}, \bibinfo{editor}{A.~Montoyo}, \bibinfo{editor}{P.~M. Barco}, \bibinfo{editor}{E.~Boldrini} (Eds.), \bibinfo{booktitle}{Proceedings of the 3rd Workshop in Computational Approaches to Subjectivity and Sentiment Analysis}, \bibinfo{publisher}{Association for Computational Linguistics}, \bibinfo{address}{Jeju, Korea}, \bibinfo{year}{2012}, pp. \bibinfo{pages}{52--60}. \URLprefix \url{https://aclanthology.org/W12-3709}.
\bibitem[{Balahur and Turchi(2014)}]{balahur2014comparative}
\bibinfo{author}{A.~Balahur}, \bibinfo{author}{M.~Turchi},
\newblock \bibinfo{title}{Comparative experiments using supervised learning and machine translation for multilingual sentiment analysis},
\newblock \bibinfo{journal}{Computer Speech \& Language} \bibinfo{volume}{28} (\bibinfo{year}{2014}) \bibinfo{pages}{56--75}.
\bibitem[{Eriguchi et~al.(2018)Eriguchi, Johnson, Firat, Kazawa, and Macherey}]{eriguchi2018zeroshot}
\bibinfo{author}{A.~Eriguchi}, \bibinfo{author}{M.~Johnson}, \bibinfo{author}{O.~Firat}, \bibinfo{author}{H.~Kazawa}, \bibinfo{author}{W.~Macherey}, \bibinfo{title}{Zero-shot cross-lingual classification using multilingual neural machine translation}, \bibinfo{year}{2018}. \href{http://arxiv.org/abs/1809.04686}{\tt arXiv:1809.04686}.
\bibitem[{Ghorbel(2012)}]{ghorbel-2012-french}
\bibinfo{author}{H.~Ghorbel},
\newblock \bibinfo{title}{Experiments in cross-lingual sentiment analysis in discussion forums},
\newblock in: \bibinfo{editor}{K.~Aberer}, \bibinfo{editor}{A.~Flache}, \bibinfo{editor}{W.~Jager}, \bibinfo{editor}{L.~Liu}, \bibinfo{editor}{J.~Tang}, \bibinfo{editor}{C.~Gu{\'e}ret} (Eds.), \bibinfo{booktitle}{Social Informatics}, \bibinfo{publisher}{Springer Berlin Heidelberg}, \bibinfo{address}{Berlin, Heidelberg}, \bibinfo{year}{2012}, pp. \bibinfo{pages}{138--151}.
\bibitem[{Zhou et~al.(2015)Zhou, Chen, Shi, and Huang}]{zhou-etal-2015-learning-bilingual}
\bibinfo{author}{H.~Zhou}, \bibinfo{author}{L.~Chen}, \bibinfo{author}{F.~Shi}, \bibinfo{author}{D.~Huang},
\newblock \bibinfo{title}{Learning bilingual sentiment word embeddings for cross-language sentiment classification},
\newblock in: \bibinfo{editor}{C.~Zong}, \bibinfo{editor}{M.~Strube} (Eds.), \bibinfo{booktitle}{Proceedings of the 53rd Annual Meeting of the Association for Computational Linguistics and the 7th International Joint Conference on Natural Language Processing (Volume 1: Long Papers)}, \bibinfo{publisher}{Association for Computational Linguistics}, \bibinfo{address}{Beijing, China}, \bibinfo{year}{2015}, pp. \bibinfo{pages}{430--440}. \URLprefix \url{https://aclanthology.org/P15-1042}. \DOIprefix\doi{10.3115/v1/P15-1042}.
\bibitem[{P{\v{r}}ib{\'{a}}{\v{n}} et~al.(2022)P{\v{r}}ib{\'{a}}{\v{n}}, {\v{S}}m{\'{\i}}d, Mi{\v{s}}tera, and Kr{\'{a}}l}]{priban-tsd-2022}
\bibinfo{author}{P.~P{\v{r}}ib{\'{a}}{\v{n}}}, \bibinfo{author}{J.~{\v{S}}m{\'{\i}}d}, \bibinfo{author}{A.~Mi{\v{s}}tera}, \bibinfo{author}{P.~Kr{\'{a}}l},
\newblock \bibinfo{title}{Linear transformations for cross-lingual sentiment analysis},
\newblock in: \bibinfo{editor}{P.~Sojka}, \bibinfo{editor}{A.~Hor{\'{a}}k}, \bibinfo{editor}{I.~Kopecek}, \bibinfo{editor}{K.~Pala} (Eds.), \bibinfo{booktitle}{Text, Speech, and Dialogue - 25th International Conference, {TSD} 2022, Brno, Czech Republic, September 6-9, 2022, Proceedings}, volume \bibinfo{volume}{13502} of \textit{\bibinfo{series}{Lecture Notes in Computer Science}}, \bibinfo{publisher}{Springer}, \bibinfo{year}{2022}, pp. \bibinfo{pages}{125--137}. \URLprefix \url{https://doi.org/10.1007/978-3-031-16270-1\_11}. \DOIprefix\doi{10.1007/978-3-031-16270-1\_11}.
\bibitem[{Přibáň et~al.(2024)Přibáň, Šmíd, Steinberger, and Mištera}]{PRIBAN2024123247}
\bibinfo{author}{P.~Přibáň}, \bibinfo{author}{J.~Šmíd}, \bibinfo{author}{J.~Steinberger}, \bibinfo{author}{A.~Mištera},
\newblock \bibinfo{title}{A comparative study of cross-lingual sentiment analysis},
\newblock \bibinfo{journal}{Expert Systems with Applications} \bibinfo{volume}{247} (\bibinfo{year}{2024}) \bibinfo{pages}{123247}.
\bibitem[{Barnes et~al.(2018)Barnes, Klinger, and Schulte~im Walde}]{barnes-etal-2018-bilingual}
\bibinfo{author}{J.~Barnes}, \bibinfo{author}{R.~Klinger}, \bibinfo{author}{S.~Schulte~im Walde},
\newblock \bibinfo{title}{Bilingual sentiment embeddings: Joint projection of sentiment across languages},
\newblock in: \bibinfo{editor}{I.~Gurevych}, \bibinfo{editor}{Y.~Miyao} (Eds.), \bibinfo{booktitle}{Proceedings of the 56th Annual Meeting of the Association for Computational Linguistics (Volume 1: Long Papers)}, \bibinfo{publisher}{Association for Computational Linguistics}, \bibinfo{address}{Melbourne, Australia}, \bibinfo{year}{2018}, pp. \bibinfo{pages}{2483--2493}. \URLprefix \url{https://aclanthology.org/P18-1231/}. \DOIprefix\doi{10.18653/v1/P18-1231}.
\bibitem[{P{\v{r}}ib{\'a}{\v{n}} and Steinberger(2021)}]{priban-steinberger-2021-multilingual}
\bibinfo{author}{P.~P{\v{r}}ib{\'a}{\v{n}}}, \bibinfo{author}{J.~Steinberger},
\newblock \bibinfo{title}{Are the multilingual models better? improving {C}zech sentiment with transformers},
\newblock in: \bibinfo{editor}{R.~Mitkov}, \bibinfo{editor}{G.~Angelova} (Eds.), \bibinfo{booktitle}{Proceedings of the International Conference on Recent Advances in Natural Language Processing (RANLP 2021)}, \bibinfo{publisher}{INCOMA Ltd.}, \bibinfo{address}{Held Online}, \bibinfo{year}{2021}, pp. \bibinfo{pages}{1138--1149}. \URLprefix \url{https://aclanthology.org/2021.ranlp-1.128}.
\bibitem[{Thakkar et~al.(2022)Thakkar, Preradovic, and Tadic}]{thakkar2022multi}
\bibinfo{author}{G.~Thakkar}, \bibinfo{author}{N.~M. Preradovic}, \bibinfo{author}{M.~Tadic},
\newblock \bibinfo{title}{Multi-task learning for cross-lingual sentiment analysis},
\newblock \bibinfo{journal}{arXiv preprint arXiv:2212.07160}  (\bibinfo{year}{2022}).
\bibitem[{Wang and Banko(2021)}]{wang-banko-2021-practical}
\bibinfo{author}{C.~Wang}, \bibinfo{author}{M.~Banko},
\newblock \bibinfo{title}{Practical transformer-based multilingual text classification},
\newblock in: \bibinfo{editor}{Y.-b. Kim}, \bibinfo{editor}{Y.~Li}, \bibinfo{editor}{O.~Rambow} (Eds.), \bibinfo{booktitle}{Proceedings of the 2021 Conference of the North American Chapter of the Association for Computational Linguistics: Human Language Technologies: Industry Papers}, \bibinfo{publisher}{Association for Computational Linguistics}, \bibinfo{address}{Online}, \bibinfo{year}{2021}, pp. \bibinfo{pages}{121--129}. \URLprefix \url{https://aclanthology.org/2021.naacl-industry.16/}. \DOIprefix\doi{10.18653/v1/2021.naacl-industry.16}.
\bibitem[{Barriere and Balahur(2020)}]{barriere-balahur-2020-improving}
\bibinfo{author}{V.~Barriere}, \bibinfo{author}{A.~Balahur},
\newblock \bibinfo{title}{Improving sentiment analysis over non-{E}nglish tweets using multilingual transformers and automatic translation for data-augmentation},
\newblock in: \bibinfo{editor}{D.~Scott}, \bibinfo{editor}{N.~Bel}, \bibinfo{editor}{C.~Zong} (Eds.), \bibinfo{booktitle}{Proceedings of the 28th International Conference on Computational Linguistics}, \bibinfo{publisher}{International Committee on Computational Linguistics}, \bibinfo{address}{Barcelona, Spain (Online)}, \bibinfo{year}{2020}, pp. \bibinfo{pages}{266--271}. \URLprefix \url{https://aclanthology.org/2020.coling-main.23}. \DOIprefix\doi{10.18653/v1/2020.coling-main.23}.
\bibitem[{Ho et~al.(2024)Ho, Chean, and Lim}]{10.1145/3651781.3651819}
\bibinfo{author}{C.~F. Ho}, \bibinfo{author}{K.~L. Chean}, \bibinfo{author}{T.~M. Lim},
\newblock \bibinfo{title}{Leveraging machine translation to enhance sentiment analysis on multilingual text},
\newblock in: \bibinfo{booktitle}{Proceedings of the 2024 13th International Conference on Software and Computer Applications}, ICSCA '24, \bibinfo{publisher}{Association for Computing Machinery}, \bibinfo{address}{New York, NY, USA}, \bibinfo{year}{2024}, p. \bibinfo{pages}{242–248}. \URLprefix \url{https://doi.org/10.1145/3651781.3651819}. \DOIprefix\doi{10.1145/3651781.3651819}.
\bibitem[{Roy(2024)}]{10.1145/3600229}
\bibinfo{author}{P.~K. Roy},
\newblock \bibinfo{title}{Deep ensemble network for sentiment analysis in bi-lingual low-resource languages},
\newblock \bibinfo{journal}{ACM Trans. Asian Low-Resour. Lang. Inf. Process.} \bibinfo{volume}{23} (\bibinfo{year}{2024}).
\bibitem[{Ilyas et~al.(2023)Ilyas, Shahzad, and Kamran~Malik}]{10.1145/3552515}
\bibinfo{author}{A.~Ilyas}, \bibinfo{author}{K.~Shahzad}, \bibinfo{author}{M.~Kamran~Malik},
\newblock \bibinfo{title}{Emotion detection in code-mixed roman urdu - english text},
\newblock \bibinfo{journal}{ACM Trans. Asian Low-Resour. Lang. Inf. Process.} \bibinfo{volume}{22} (\bibinfo{year}{2023}).
\bibitem[{Nazir et~al.(2025)Nazir, Faisal, Habib, and Ahmad}]{nazir-2025-multi}
\bibinfo{author}{M.~Nazir}, \bibinfo{author}{C.~Faisal}, \bibinfo{author}{M.~A. Habib}, \bibinfo{author}{H.~Ahmad},
\newblock \bibinfo{title}{Leveraging multilingual transformer for multiclass sentiment analysis in code-mixed data of low-resource languages},
\newblock \bibinfo{journal}{IEEE Access} \bibinfo{volume}{PP} (\bibinfo{year}{2025}).
\bibitem[{Winata et~al.(2022)Winata, Wu, Kulkarni, Solorio, and Preotiuc-Pietro}]{winata-etal-2022-cross}
\bibinfo{author}{G.~Winata}, \bibinfo{author}{S.~Wu}, \bibinfo{author}{M.~Kulkarni}, \bibinfo{author}{T.~Solorio}, \bibinfo{author}{D.~Preotiuc-Pietro},
\newblock \bibinfo{title}{Cross-lingual few-shot learning on unseen languages},
\newblock in: \bibinfo{editor}{Y.~He}, \bibinfo{editor}{H.~Ji}, \bibinfo{editor}{S.~Li}, \bibinfo{editor}{Y.~Liu}, \bibinfo{editor}{C.-H. Chang} (Eds.), \bibinfo{booktitle}{Proceedings of the 2nd Conference of the Asia-Pacific Chapter of the Association for Computational Linguistics and the 12th International Joint Conference on Natural Language Processing (Volume 1: Long Papers)}, \bibinfo{publisher}{Association for Computational Linguistics}, \bibinfo{address}{Online only}, \bibinfo{year}{2022}, pp. \bibinfo{pages}{777--791}. \URLprefix \url{https://aclanthology.org/2022.aacl-main.59/}. \DOIprefix\doi{10.18653/v1/2022.aacl-main.59}.
\bibitem[{Minaee et~al.(2024)Minaee, Mikolov, Nikzad, Chenaghlu, Socher, Amatriain, and Gao}]{minaee2024largelanguagemodelssurvey}
\bibinfo{author}{S.~Minaee}, \bibinfo{author}{T.~Mikolov}, \bibinfo{author}{N.~Nikzad}, \bibinfo{author}{M.~Chenaghlu}, \bibinfo{author}{R.~Socher}, \bibinfo{author}{X.~Amatriain}, \bibinfo{author}{J.~Gao}, \bibinfo{title}{Large language models: A survey}, \bibinfo{year}{2024}. \URLprefix \url{https://arxiv.org/abs/2402.06196}. \href{http://arxiv.org/abs/2402.06196}{\tt arXiv:2402.06196}.
\bibitem[{Chowdhery et~al.(2024)Chowdhery, Narang, Devlin, Bosma, Mishra, Roberts, Barham, Chung, Sutton, Gehrmann, Schuh, Shi, Tsvyashchenko, Maynez, Rao, Barnes, Tay, Shazeer, Prabhakaran, Reif, Du, Hutchinson, Pope, Bradbury, Austin, Isard, Gur-Ari, Yin, Duke, Levskaya, Ghemawat, Dev, Michalewski, Garcia, Misra, Robinson, Fedus, Zhou, Ippolito, Luan, Lim, Zoph, Spiridonov, Sepassi, Dohan, Agrawal, Omernick, Dai, Pillai, Pellat, Lewkowycz, Moreira, Child, Polozov, Lee, Zhou, Wang, Saeta, Diaz, Firat, Catasta, Wei, Meier-Hellstern, Eck, Dean, Petrov, and Fiedel}]{10.5555/3648699.3648939}
\bibinfo{author}{A.~Chowdhery}, \bibinfo{author}{S.~Narang}, \bibinfo{author}{J.~Devlin}, \bibinfo{author}{M.~Bosma}, \bibinfo{author}{G.~Mishra}, \bibinfo{author}{A.~Roberts}, \bibinfo{author}{P.~Barham}, \bibinfo{author}{H.~W. Chung}, \bibinfo{author}{C.~Sutton}, \bibinfo{author}{S.~Gehrmann}, \bibinfo{author}{P.~Schuh}, \bibinfo{author}{K.~Shi}, \bibinfo{author}{S.~Tsvyashchenko}, \bibinfo{author}{J.~Maynez}, \bibinfo{author}{A.~Rao}, \bibinfo{author}{P.~Barnes}, \bibinfo{author}{Y.~Tay}, \bibinfo{author}{N.~Shazeer}, \bibinfo{author}{V.~Prabhakaran}, \bibinfo{author}{E.~Reif}, \bibinfo{author}{N.~Du}, \bibinfo{author}{B.~Hutchinson}, \bibinfo{author}{R.~Pope}, \bibinfo{author}{J.~Bradbury}, \bibinfo{author}{J.~Austin}, \bibinfo{author}{M.~Isard}, \bibinfo{author}{G.~Gur-Ari}, \bibinfo{author}{P.~Yin}, \bibinfo{author}{T.~Duke}, \bibinfo{author}{A.~Levskaya}, \bibinfo{author}{S.~Ghemawat}, \bibinfo{author}{S.~Dev}, \bibinfo{author}{H.~Michalewski}, \bibinfo{author}{X.~Garcia}, \bibinfo{author}{V.~Misra},
  \bibinfo{author}{K.~Robinson}, \bibinfo{author}{L.~Fedus}, \bibinfo{author}{D.~Zhou}, \bibinfo{author}{D.~Ippolito}, \bibinfo{author}{D.~Luan}, \bibinfo{author}{H.~Lim}, \bibinfo{author}{B.~Zoph}, \bibinfo{author}{A.~Spiridonov}, \bibinfo{author}{R.~Sepassi}, \bibinfo{author}{D.~Dohan}, \bibinfo{author}{S.~Agrawal}, \bibinfo{author}{M.~Omernick}, \bibinfo{author}{A.~M. Dai}, \bibinfo{author}{T.~S. Pillai}, \bibinfo{author}{M.~Pellat}, \bibinfo{author}{A.~Lewkowycz}, \bibinfo{author}{E.~Moreira}, \bibinfo{author}{R.~Child}, \bibinfo{author}{O.~Polozov}, \bibinfo{author}{K.~Lee}, \bibinfo{author}{Z.~Zhou}, \bibinfo{author}{X.~Wang}, \bibinfo{author}{B.~Saeta}, \bibinfo{author}{M.~Diaz}, \bibinfo{author}{O.~Firat}, \bibinfo{author}{M.~Catasta}, \bibinfo{author}{J.~Wei}, \bibinfo{author}{K.~Meier-Hellstern}, \bibinfo{author}{D.~Eck}, \bibinfo{author}{J.~Dean}, \bibinfo{author}{S.~Petrov}, \bibinfo{author}{N.~Fiedel},
\newblock \bibinfo{title}{Palm: scaling language modeling with pathways},
\newblock \bibinfo{journal}{J. Mach. Learn. Res.} \bibinfo{volume}{24} (\bibinfo{year}{2024}).
\bibitem[{OpenAI et~al.(2024)OpenAI, Achiam, Adler, Agarwal, Ahmad, Akkaya, Aleman, Almeida, Altenschmidt, Altman, Anadkat, Avila, Babuschkin, Balaji, Balcom, Baltescu, Bao, Bavarian, Belgum, Bello, Berdine, Bernadett-Shapiro, Berner, Bogdonoff, Boiko, Boyd, Brakman, Brockman, Brooks, Brundage, Button, Cai, Campbell, Cann, Carey, Carlson, Carmichael, Chan, Chang, Chantzis, Chen, Chen, Chen, Chen, Chen, Chess, Cho, Chu, Chung, Cummings, Currier, Dai, Decareaux, Degry, Deutsch, Deville, Dhar, Dohan, Dowling, Dunning, Ecoffet, Eleti, Eloundou, Farhi, Fedus, Felix, Fishman, Forte, Fulford, Gao, Georges, Gibson, Goel, Gogineni, Goh, Gontijo-Lopes, Gordon, Grafstein, Gray, Greene, Gross, Gu, Guo, Hallacy, Han, Harris, He, Heaton, Heidecke, Hesse, Hickey, Hickey, Hoeschele, Houghton, Hsu, Hu, Hu, Huizinga, Jain, Jain, Jang, Jiang, Jiang, Jin, Jin, Jomoto, Jonn, Jun, Kaftan, Łukasz Kaiser, Kamali, Kanitscheider, Keskar, Khan, Kilpatrick, Kim, Kim, Kim, Kirchner, Kiros, Knight, Kokotajlo, Łukasz Kondraciuk,
  Kondrich, Konstantinidis, Kosic, Krueger, Kuo, Lampe, Lan, Lee, Leike, Leung, Levy, Li, Lim, Lin, Lin, Litwin, Lopez, Lowe, Lue, Makanju, Malfacini, Manning, Markov, Markovski, Martin, Mayer, Mayne, McGrew, McKinney, McLeavey, McMillan, McNeil, Medina, Mehta, Menick, Metz, Mishchenko, Mishkin, Monaco, Morikawa, Mossing, Mu, Murati, Murk, Mély, Nair, Nakano, Nayak, Neelakantan, Ngo, Noh, Ouyang, O'Keefe, Pachocki, Paino, Palermo, Pantuliano, Parascandolo, Parish, Parparita, Passos, Pavlov, Peng, Perelman, de~Avila Belbute~Peres, Petrov, de~Oliveira~Pinto, Michael, Pokorny, Pokrass, Pong, Powell, Power, Power, Proehl, Puri, Radford, Rae, Ramesh, Raymond, Real, Rimbach, Ross, Rotsted, Roussez, Ryder, Saltarelli, Sanders, Santurkar, Sastry, Schmidt, Schnurr, Schulman, Selsam, Sheppard, Sherbakov, Shieh, Shoker, Shyam, Sidor, Sigler, Simens, Sitkin, Slama, Sohl, Sokolowsky, Song, Staudacher, Such, Summers, Sutskever, Tang, Tezak, Thompson, Tillet, Tootoonchian, Tseng, Tuggle, Turley, Tworek, Uribe, Vallone,
  Vijayvergiya, Voss, Wainwright, Wang, Wang, Wang, Ward, Wei, Weinmann, Welihinda, Welinder, Weng, Weng, Wiethoff, Willner, Winter, Wolrich, Wong, Workman, Wu, Wu, Wu, Xiao, Xu, Yoo, Yu, Yuan, Zaremba, Zellers, Zhang, Zhang, Zhao, Zheng, Zhuang, Zhuk, and Zoph}]{openai2024gpt4technicalreport}
\bibinfo{author}{OpenAI}, \bibinfo{author}{J.~Achiam}, \bibinfo{author}{S.~Adler}, \bibinfo{author}{S.~Agarwal}, \bibinfo{author}{L.~Ahmad}, \bibinfo{author}{I.~Akkaya}, \bibinfo{author}{F.~L. Aleman}, \bibinfo{author}{D.~Almeida}, \bibinfo{author}{J.~Altenschmidt}, \bibinfo{author}{S.~Altman}, \bibinfo{author}{S.~Anadkat}, \bibinfo{author}{R.~Avila}, \bibinfo{author}{I.~Babuschkin}, \bibinfo{author}{S.~Balaji}, \bibinfo{author}{V.~Balcom}, \bibinfo{author}{P.~Baltescu}, \bibinfo{author}{H.~Bao}, \bibinfo{author}{M.~Bavarian}, \bibinfo{author}{J.~Belgum}, \bibinfo{author}{I.~Bello}, \bibinfo{author}{J.~Berdine}, \bibinfo{author}{G.~Bernadett-Shapiro}, \bibinfo{author}{C.~Berner}, \bibinfo{author}{L.~Bogdonoff}, \bibinfo{author}{O.~Boiko}, \bibinfo{author}{M.~Boyd}, \bibinfo{author}{A.-L. Brakman}, \bibinfo{author}{G.~Brockman}, \bibinfo{author}{T.~Brooks}, \bibinfo{author}{M.~Brundage}, \bibinfo{author}{K.~Button}, \bibinfo{author}{T.~Cai}, \bibinfo{author}{R.~Campbell}, \bibinfo{author}{A.~Cann},
  \bibinfo{author}{B.~Carey}, \bibinfo{author}{C.~Carlson}, \bibinfo{author}{R.~Carmichael}, \bibinfo{author}{B.~Chan}, \bibinfo{author}{C.~Chang}, \bibinfo{author}{F.~Chantzis}, \bibinfo{author}{D.~Chen}, \bibinfo{author}{S.~Chen}, \bibinfo{author}{R.~Chen}, \bibinfo{author}{J.~Chen}, \bibinfo{author}{M.~Chen}, \bibinfo{author}{B.~Chess}, \bibinfo{author}{C.~Cho}, \bibinfo{author}{C.~Chu}, \bibinfo{author}{H.~W. Chung}, \bibinfo{author}{D.~Cummings}, \bibinfo{author}{J.~Currier}, \bibinfo{author}{Y.~Dai}, \bibinfo{author}{C.~Decareaux}, \bibinfo{author}{T.~Degry}, \bibinfo{author}{N.~Deutsch}, \bibinfo{author}{D.~Deville}, \bibinfo{author}{A.~Dhar}, \bibinfo{author}{D.~Dohan}, \bibinfo{author}{S.~Dowling}, \bibinfo{author}{S.~Dunning}, \bibinfo{author}{A.~Ecoffet}, \bibinfo{author}{A.~Eleti}, \bibinfo{author}{T.~Eloundou}, \bibinfo{author}{D.~Farhi}, \bibinfo{author}{L.~Fedus}, \bibinfo{author}{N.~Felix}, \bibinfo{author}{S.~P. Fishman}, \bibinfo{author}{J.~Forte}, \bibinfo{author}{I.~Fulford},
  \bibinfo{author}{L.~Gao}, \bibinfo{author}{E.~Georges}, \bibinfo{author}{C.~Gibson}, \bibinfo{author}{V.~Goel}, \bibinfo{author}{T.~Gogineni}, \bibinfo{author}{G.~Goh}, \bibinfo{author}{R.~Gontijo-Lopes}, \bibinfo{author}{J.~Gordon}, \bibinfo{author}{M.~Grafstein}, \bibinfo{author}{S.~Gray}, \bibinfo{author}{R.~Greene}, \bibinfo{author}{J.~Gross}, \bibinfo{author}{S.~S. Gu}, \bibinfo{author}{Y.~Guo}, \bibinfo{author}{C.~Hallacy}, \bibinfo{author}{J.~Han}, \bibinfo{author}{J.~Harris}, \bibinfo{author}{Y.~He}, \bibinfo{author}{M.~Heaton}, \bibinfo{author}{J.~Heidecke}, \bibinfo{author}{C.~Hesse}, \bibinfo{author}{A.~Hickey}, \bibinfo{author}{W.~Hickey}, \bibinfo{author}{P.~Hoeschele}, \bibinfo{author}{B.~Houghton}, \bibinfo{author}{K.~Hsu}, \bibinfo{author}{S.~Hu}, \bibinfo{author}{X.~Hu}, \bibinfo{author}{J.~Huizinga}, \bibinfo{author}{S.~Jain}, \bibinfo{author}{S.~Jain}, \bibinfo{author}{J.~Jang}, \bibinfo{author}{A.~Jiang}, \bibinfo{author}{R.~Jiang}, \bibinfo{author}{H.~Jin}, \bibinfo{author}{D.~Jin},
  \bibinfo{author}{S.~Jomoto}, \bibinfo{author}{B.~Jonn}, \bibinfo{author}{H.~Jun}, \bibinfo{author}{T.~Kaftan}, \bibinfo{author}{Łukasz Kaiser}, \bibinfo{author}{A.~Kamali}, \bibinfo{author}{I.~Kanitscheider}, \bibinfo{author}{N.~S. Keskar}, \bibinfo{author}{T.~Khan}, \bibinfo{author}{L.~Kilpatrick}, \bibinfo{author}{J.~W. Kim}, \bibinfo{author}{C.~Kim}, \bibinfo{author}{Y.~Kim}, \bibinfo{author}{J.~H. Kirchner}, \bibinfo{author}{J.~Kiros}, \bibinfo{author}{M.~Knight}, \bibinfo{author}{D.~Kokotajlo}, \bibinfo{author}{Łukasz Kondraciuk}, \bibinfo{author}{A.~Kondrich}, \bibinfo{author}{A.~Konstantinidis}, \bibinfo{author}{K.~Kosic}, \bibinfo{author}{G.~Krueger}, \bibinfo{author}{V.~Kuo}, \bibinfo{author}{M.~Lampe}, \bibinfo{author}{I.~Lan}, \bibinfo{author}{T.~Lee}, \bibinfo{author}{J.~Leike}, \bibinfo{author}{J.~Leung}, \bibinfo{author}{D.~Levy}, \bibinfo{author}{C.~M. Li}, \bibinfo{author}{R.~Lim}, \bibinfo{author}{M.~Lin}, \bibinfo{author}{S.~Lin}, \bibinfo{author}{M.~Litwin}, \bibinfo{author}{T.~Lopez},
  \bibinfo{author}{R.~Lowe}, \bibinfo{author}{P.~Lue}, \bibinfo{author}{A.~Makanju}, \bibinfo{author}{K.~Malfacini}, \bibinfo{author}{S.~Manning}, \bibinfo{author}{T.~Markov}, \bibinfo{author}{Y.~Markovski}, \bibinfo{author}{B.~Martin}, \bibinfo{author}{K.~Mayer}, \bibinfo{author}{A.~Mayne}, \bibinfo{author}{B.~McGrew}, \bibinfo{author}{S.~M. McKinney}, \bibinfo{author}{C.~McLeavey}, \bibinfo{author}{P.~McMillan}, \bibinfo{author}{J.~McNeil}, \bibinfo{author}{D.~Medina}, \bibinfo{author}{A.~Mehta}, \bibinfo{author}{J.~Menick}, \bibinfo{author}{L.~Metz}, \bibinfo{author}{A.~Mishchenko}, \bibinfo{author}{P.~Mishkin}, \bibinfo{author}{V.~Monaco}, \bibinfo{author}{E.~Morikawa}, \bibinfo{author}{D.~Mossing}, \bibinfo{author}{T.~Mu}, \bibinfo{author}{M.~Murati}, \bibinfo{author}{O.~Murk}, \bibinfo{author}{D.~Mély}, \bibinfo{author}{A.~Nair}, \bibinfo{author}{R.~Nakano}, \bibinfo{author}{R.~Nayak}, \bibinfo{author}{A.~Neelakantan}, \bibinfo{author}{R.~Ngo}, \bibinfo{author}{H.~Noh}, \bibinfo{author}{L.~Ouyang},
  \bibinfo{author}{C.~O'Keefe}, \bibinfo{author}{J.~Pachocki}, \bibinfo{author}{A.~Paino}, \bibinfo{author}{J.~Palermo}, \bibinfo{author}{A.~Pantuliano}, \bibinfo{author}{G.~Parascandolo}, \bibinfo{author}{J.~Parish}, \bibinfo{author}{E.~Parparita}, \bibinfo{author}{A.~Passos}, \bibinfo{author}{M.~Pavlov}, \bibinfo{author}{A.~Peng}, \bibinfo{author}{A.~Perelman}, \bibinfo{author}{F.~de~Avila Belbute~Peres}, \bibinfo{author}{M.~Petrov}, \bibinfo{author}{H.~P. de~Oliveira~Pinto}, \bibinfo{author}{Michael}, \bibinfo{author}{Pokorny}, \bibinfo{author}{M.~Pokrass}, \bibinfo{author}{V.~H. Pong}, \bibinfo{author}{T.~Powell}, \bibinfo{author}{A.~Power}, \bibinfo{author}{B.~Power}, \bibinfo{author}{E.~Proehl}, \bibinfo{author}{R.~Puri}, \bibinfo{author}{A.~Radford}, \bibinfo{author}{J.~Rae}, \bibinfo{author}{A.~Ramesh}, \bibinfo{author}{C.~Raymond}, \bibinfo{author}{F.~Real}, \bibinfo{author}{K.~Rimbach}, \bibinfo{author}{C.~Ross}, \bibinfo{author}{B.~Rotsted}, \bibinfo{author}{H.~Roussez}, \bibinfo{author}{N.~Ryder},
  \bibinfo{author}{M.~Saltarelli}, \bibinfo{author}{T.~Sanders}, \bibinfo{author}{S.~Santurkar}, \bibinfo{author}{G.~Sastry}, \bibinfo{author}{H.~Schmidt}, \bibinfo{author}{D.~Schnurr}, \bibinfo{author}{J.~Schulman}, \bibinfo{author}{D.~Selsam}, \bibinfo{author}{K.~Sheppard}, \bibinfo{author}{T.~Sherbakov}, \bibinfo{author}{J.~Shieh}, \bibinfo{author}{S.~Shoker}, \bibinfo{author}{P.~Shyam}, \bibinfo{author}{S.~Sidor}, \bibinfo{author}{E.~Sigler}, \bibinfo{author}{M.~Simens}, \bibinfo{author}{J.~Sitkin}, \bibinfo{author}{K.~Slama}, \bibinfo{author}{I.~Sohl}, \bibinfo{author}{B.~Sokolowsky}, \bibinfo{author}{Y.~Song}, \bibinfo{author}{N.~Staudacher}, \bibinfo{author}{F.~P. Such}, \bibinfo{author}{N.~Summers}, \bibinfo{author}{I.~Sutskever}, \bibinfo{author}{J.~Tang}, \bibinfo{author}{N.~Tezak}, \bibinfo{author}{M.~B. Thompson}, \bibinfo{author}{P.~Tillet}, \bibinfo{author}{A.~Tootoonchian}, \bibinfo{author}{E.~Tseng}, \bibinfo{author}{P.~Tuggle}, \bibinfo{author}{N.~Turley}, \bibinfo{author}{J.~Tworek},
  \bibinfo{author}{J.~F.~C. Uribe}, \bibinfo{author}{A.~Vallone}, \bibinfo{author}{A.~Vijayvergiya}, \bibinfo{author}{C.~Voss}, \bibinfo{author}{C.~Wainwright}, \bibinfo{author}{J.~J. Wang}, \bibinfo{author}{A.~Wang}, \bibinfo{author}{B.~Wang}, \bibinfo{author}{J.~Ward}, \bibinfo{author}{J.~Wei}, \bibinfo{author}{C.~Weinmann}, \bibinfo{author}{A.~Welihinda}, \bibinfo{author}{P.~Welinder}, \bibinfo{author}{J.~Weng}, \bibinfo{author}{L.~Weng}, \bibinfo{author}{M.~Wiethoff}, \bibinfo{author}{D.~Willner}, \bibinfo{author}{C.~Winter}, \bibinfo{author}{S.~Wolrich}, \bibinfo{author}{H.~Wong}, \bibinfo{author}{L.~Workman}, \bibinfo{author}{S.~Wu}, \bibinfo{author}{J.~Wu}, \bibinfo{author}{M.~Wu}, \bibinfo{author}{K.~Xiao}, \bibinfo{author}{T.~Xu}, \bibinfo{author}{S.~Yoo}, \bibinfo{author}{K.~Yu}, \bibinfo{author}{Q.~Yuan}, \bibinfo{author}{W.~Zaremba}, \bibinfo{author}{R.~Zellers}, \bibinfo{author}{C.~Zhang}, \bibinfo{author}{M.~Zhang}, \bibinfo{author}{S.~Zhao}, \bibinfo{author}{T.~Zheng},
  \bibinfo{author}{J.~Zhuang}, \bibinfo{author}{W.~Zhuk}, \bibinfo{author}{B.~Zoph}, \bibinfo{title}{Gpt-4 technical report}, \bibinfo{year}{2024}. \URLprefix \url{https://arxiv.org/abs/2303.08774}. \href{http://arxiv.org/abs/2303.08774}{\tt arXiv:2303.08774}.
\bibitem[{Touvron et~al.(2023{\natexlab{a}})Touvron, Lavril, Izacard, Martinet, Lachaux, Lacroix, Rozière, Goyal, Hambro, Azhar, Rodriguez, Joulin, Grave, and Lample}]{touvron2023llamaopenefficientfoundation}
\bibinfo{author}{H.~Touvron}, \bibinfo{author}{T.~Lavril}, \bibinfo{author}{G.~Izacard}, \bibinfo{author}{X.~Martinet}, \bibinfo{author}{M.-A. Lachaux}, \bibinfo{author}{T.~Lacroix}, \bibinfo{author}{B.~Rozière}, \bibinfo{author}{N.~Goyal}, \bibinfo{author}{E.~Hambro}, \bibinfo{author}{F.~Azhar}, \bibinfo{author}{A.~Rodriguez}, \bibinfo{author}{A.~Joulin}, \bibinfo{author}{E.~Grave}, \bibinfo{author}{G.~Lample}, \bibinfo{title}{Llama: Open and efficient foundation language models}, \bibinfo{year}{2023}{\natexlab{a}}. \URLprefix \url{https://arxiv.org/abs/2302.13971}. \href{http://arxiv.org/abs/2302.13971}{\tt arXiv:2302.13971}.
\bibitem[{Touvron et~al.(2023{\natexlab{b}})Touvron, Martin, Stone, Albert, Almahairi, Babaei, Bashlykov, Batra, Bhargava, Bhosale, Bikel, Blecher, Ferrer, Chen, Cucurull, Esiobu, Fernandes, Fu, Fu, Fuller, Gao, Goswami, Goyal, Hartshorn, Hosseini, Hou, Inan, Kardas, Kerkez, Khabsa, Kloumann, Korenev, Koura, Lachaux, Lavril, Lee, Liskovich, Lu, Mao, Martinet, Mihaylov, Mishra, Molybog, Nie, Poulton, Reizenstein, Rungta, Saladi, Schelten, Silva, Smith, Subramanian, Tan, Tang, Taylor, Williams, Kuan, Xu, Yan, Zarov, Zhang, Fan, Kambadur, Narang, Rodriguez, Stojnic, Edunov, and Scialom}]{touvron2023llama2openfoundation}
\bibinfo{author}{H.~Touvron}, \bibinfo{author}{L.~Martin}, \bibinfo{author}{K.~Stone}, \bibinfo{author}{P.~Albert}, \bibinfo{author}{A.~Almahairi}, \bibinfo{author}{Y.~Babaei}, \bibinfo{author}{N.~Bashlykov}, \bibinfo{author}{S.~Batra}, \bibinfo{author}{P.~Bhargava}, \bibinfo{author}{S.~Bhosale}, \bibinfo{author}{D.~Bikel}, \bibinfo{author}{L.~Blecher}, \bibinfo{author}{C.~C. Ferrer}, \bibinfo{author}{M.~Chen}, \bibinfo{author}{G.~Cucurull}, \bibinfo{author}{D.~Esiobu}, \bibinfo{author}{J.~Fernandes}, \bibinfo{author}{J.~Fu}, \bibinfo{author}{W.~Fu}, \bibinfo{author}{B.~Fuller}, \bibinfo{author}{C.~Gao}, \bibinfo{author}{V.~Goswami}, \bibinfo{author}{N.~Goyal}, \bibinfo{author}{A.~Hartshorn}, \bibinfo{author}{S.~Hosseini}, \bibinfo{author}{R.~Hou}, \bibinfo{author}{H.~Inan}, \bibinfo{author}{M.~Kardas}, \bibinfo{author}{V.~Kerkez}, \bibinfo{author}{M.~Khabsa}, \bibinfo{author}{I.~Kloumann}, \bibinfo{author}{A.~Korenev}, \bibinfo{author}{P.~S. Koura}, \bibinfo{author}{M.-A. Lachaux},
  \bibinfo{author}{T.~Lavril}, \bibinfo{author}{J.~Lee}, \bibinfo{author}{D.~Liskovich}, \bibinfo{author}{Y.~Lu}, \bibinfo{author}{Y.~Mao}, \bibinfo{author}{X.~Martinet}, \bibinfo{author}{T.~Mihaylov}, \bibinfo{author}{P.~Mishra}, \bibinfo{author}{I.~Molybog}, \bibinfo{author}{Y.~Nie}, \bibinfo{author}{A.~Poulton}, \bibinfo{author}{J.~Reizenstein}, \bibinfo{author}{R.~Rungta}, \bibinfo{author}{K.~Saladi}, \bibinfo{author}{A.~Schelten}, \bibinfo{author}{R.~Silva}, \bibinfo{author}{E.~M. Smith}, \bibinfo{author}{R.~Subramanian}, \bibinfo{author}{X.~E. Tan}, \bibinfo{author}{B.~Tang}, \bibinfo{author}{R.~Taylor}, \bibinfo{author}{A.~Williams}, \bibinfo{author}{J.~X. Kuan}, \bibinfo{author}{P.~Xu}, \bibinfo{author}{Z.~Yan}, \bibinfo{author}{I.~Zarov}, \bibinfo{author}{Y.~Zhang}, \bibinfo{author}{A.~Fan}, \bibinfo{author}{M.~Kambadur}, \bibinfo{author}{S.~Narang}, \bibinfo{author}{A.~Rodriguez}, \bibinfo{author}{R.~Stojnic}, \bibinfo{author}{S.~Edunov}, \bibinfo{author}{T.~Scialom}, \bibinfo{title}{Llama 2: Open
  foundation and fine-tuned chat models}, \bibinfo{year}{2023}{\natexlab{b}}. \URLprefix \url{https://arxiv.org/abs/2307.09288}. \href{http://arxiv.org/abs/2307.09288}{\tt arXiv:2307.09288}.
\bibitem[{Dubey et~al.(2024)Dubey, Jauhri, Pandey, Kadian, Al-Dahle et~al.}]{dubey2024llama3herdmodels}
\bibinfo{author}{A.~Dubey}, \bibinfo{author}{A.~Jauhri}, \bibinfo{author}{A.~Pandey}, \bibinfo{author}{A.~Kadian}, \bibinfo{author}{A.~Al-Dahle}, et~al., \bibinfo{title}{The llama 3 herd of models}, \bibinfo{year}{2024}. \URLprefix \url{https://arxiv.org/abs/2407.21783}. \href{http://arxiv.org/abs/2407.21783}{\tt arXiv:2407.21783}.
\bibitem[{Zhong et~al.(2023)Zhong, Ding, Liu, Du, and Tao}]{zhong2023chatgptunderstandtoocomparative}
\bibinfo{author}{Q.~Zhong}, \bibinfo{author}{L.~Ding}, \bibinfo{author}{J.~Liu}, \bibinfo{author}{B.~Du}, \bibinfo{author}{D.~Tao}, \bibinfo{title}{Can chatgpt understand too? a comparative study on chatgpt and fine-tuned bert}, \bibinfo{year}{2023}. \URLprefix \url{https://arxiv.org/abs/2302.10198}. \href{http://arxiv.org/abs/2302.10198}{\tt arXiv:2302.10198}.
\bibitem[{Simmering and Huoviala(2023)}]{simmering2023largelanguagemodelsaspectbased}
\bibinfo{author}{P.~F. Simmering}, \bibinfo{author}{P.~Huoviala}, \bibinfo{title}{Large language models for aspect-based sentiment analysis}, \bibinfo{year}{2023}. \URLprefix \url{https://arxiv.org/abs/2310.18025}. \href{http://arxiv.org/abs/2310.18025}{\tt arXiv:2310.18025}.
\bibitem[{Wei et~al.(2024)Wei, Wang, Schuurmans, Bosma, Ichter, Xia, Chi, Le, and Zhou}]{10.5555/3600270.3602070}
\bibinfo{author}{J.~Wei}, \bibinfo{author}{X.~Wang}, \bibinfo{author}{D.~Schuurmans}, \bibinfo{author}{M.~Bosma}, \bibinfo{author}{B.~Ichter}, \bibinfo{author}{F.~Xia}, \bibinfo{author}{E.~H. Chi}, \bibinfo{author}{Q.~V. Le}, \bibinfo{author}{D.~Zhou},
\newblock \bibinfo{title}{Chain-of-thought prompting elicits reasoning in large language models},
\newblock in: \bibinfo{booktitle}{Proceedings of the 36th International Conference on Neural Information Processing Systems}, NIPS '22, \bibinfo{publisher}{Curran Associates Inc.}, \bibinfo{address}{Red Hook, NY, USA}, \bibinfo{year}{2024}.
\bibitem[{Fei et~al.(2023)Fei, Li, Liu, Bing, Li, and Chua}]{fei-etal-2023-reasoning}
\bibinfo{author}{H.~Fei}, \bibinfo{author}{B.~Li}, \bibinfo{author}{Q.~Liu}, \bibinfo{author}{L.~Bing}, \bibinfo{author}{F.~Li}, \bibinfo{author}{T.-S. Chua},
\newblock \bibinfo{title}{Reasoning implicit sentiment with chain-of-thought prompting},
\newblock in: \bibinfo{editor}{A.~Rogers}, \bibinfo{editor}{J.~Boyd-Graber}, \bibinfo{editor}{N.~Okazaki} (Eds.), \bibinfo{booktitle}{Proceedings of the 61st Annual Meeting of the Association for Computational Linguistics (Volume 2: Short Papers)}, \bibinfo{publisher}{Association for Computational Linguistics}, \bibinfo{address}{Toronto, Canada}, \bibinfo{year}{2023}, pp. \bibinfo{pages}{1171--1182}. \URLprefix \url{https://aclanthology.org/2023.acl-short.101/}. \DOIprefix\doi{10.18653/v1/2023.acl-short.101}.
\bibitem[{Filip et~al.(2024)Filip, Pavlíček, and Sosík}]{filip2024finetuningmultilinguallanguagemodels}
\bibinfo{author}{T.~Filip}, \bibinfo{author}{M.~Pavlíček}, \bibinfo{author}{P.~Sosík}, \bibinfo{title}{Fine-tuning multilingual language models in twitter/x sentiment analysis: a study on eastern-european v4 languages}, \bibinfo{year}{2024}. \URLprefix \url{https://arxiv.org/abs/2408.02044}. \href{http://arxiv.org/abs/2408.02044}{\tt arXiv:2408.02044}.
\bibitem[{Jiang et~al.(2023)Jiang, Sablayrolles, Mensch, Bamford, Chaplot, de~las Casas, Bressand, Lengyel, Lample, Saulnier, Lavaud, Lachaux, Stock, Scao, Lavril, Wang, Lacroix, and Sayed}]{jiang2023mistral7b}
\bibinfo{author}{A.~Q. Jiang}, \bibinfo{author}{A.~Sablayrolles}, \bibinfo{author}{A.~Mensch}, \bibinfo{author}{C.~Bamford}, \bibinfo{author}{D.~S. Chaplot}, \bibinfo{author}{D.~de~las Casas}, \bibinfo{author}{F.~Bressand}, \bibinfo{author}{G.~Lengyel}, \bibinfo{author}{G.~Lample}, \bibinfo{author}{L.~Saulnier}, \bibinfo{author}{L.~R. Lavaud}, \bibinfo{author}{M.-A. Lachaux}, \bibinfo{author}{P.~Stock}, \bibinfo{author}{T.~L. Scao}, \bibinfo{author}{T.~Lavril}, \bibinfo{author}{T.~Wang}, \bibinfo{author}{T.~Lacroix}, \bibinfo{author}{W.~E. Sayed}, \bibinfo{title}{Mistral 7b}, \bibinfo{year}{2023}. \URLprefix \url{https://arxiv.org/abs/2310.06825}. \href{http://arxiv.org/abs/2310.06825}{\tt arXiv:2310.06825}.
\bibitem[{Mughal et~al.(2024)Mughal, Mujtaba, Shaikh, Kumar, and Daudpota}]{10504711}
\bibinfo{author}{N.~Mughal}, \bibinfo{author}{G.~Mujtaba}, \bibinfo{author}{S.~Shaikh}, \bibinfo{author}{A.~Kumar}, \bibinfo{author}{S.~M. Daudpota},
\newblock \bibinfo{title}{Comparative analysis of deep natural networks and large language models for aspect-based sentiment analysis},
\newblock \bibinfo{journal}{IEEE Access} \bibinfo{volume}{12} (\bibinfo{year}{2024}) \bibinfo{pages}{60943--60959}.
\bibitem[{He et~al.(2021)He, Liu, Gao, and Chen}]{he2021debertadecodingenhancedbertdisentangled}
\bibinfo{author}{P.~He}, \bibinfo{author}{X.~Liu}, \bibinfo{author}{J.~Gao}, \bibinfo{author}{W.~Chen}, \bibinfo{title}{Deberta: Decoding-enhanced bert with disentangled attention}, \bibinfo{year}{2021}. \URLprefix \url{https://arxiv.org/abs/2006.03654}. \href{http://arxiv.org/abs/2006.03654}{\tt arXiv:2006.03654}.
\bibitem[{Ding et~al.(2024)Ding, Qin, Zhao, Luo, Li, Chen, Xia, Hu, Luu, and Joty}]{ding-etal-2024-data}
\bibinfo{author}{B.~Ding}, \bibinfo{author}{C.~Qin}, \bibinfo{author}{R.~Zhao}, \bibinfo{author}{T.~Luo}, \bibinfo{author}{X.~Li}, \bibinfo{author}{G.~Chen}, \bibinfo{author}{W.~Xia}, \bibinfo{author}{J.~Hu}, \bibinfo{author}{A.~T. Luu}, \bibinfo{author}{S.~Joty},
\newblock \bibinfo{title}{Data augmentation using {LLM}s: Data perspectives, learning paradigms and challenges},
\newblock in: \bibinfo{editor}{L.-W. Ku}, \bibinfo{editor}{A.~Martins}, \bibinfo{editor}{V.~Srikumar} (Eds.), \bibinfo{booktitle}{Findings of the Association for Computational Linguistics: ACL 2024}, \bibinfo{publisher}{Association for Computational Linguistics}, \bibinfo{address}{Bangkok, Thailand}, \bibinfo{year}{2024}, pp. \bibinfo{pages}{1679--1705}. \URLprefix \url{https://aclanthology.org/2024.findings-acl.97}. \DOIprefix\doi{10.18653/v1/2024.findings-acl.97}.
\bibitem[{Li et~al.(2022)Li, Chen, Li, Wang, Qian, and Yan}]{li-etal-2022-controllable}
\bibinfo{author}{Z.~Li}, \bibinfo{author}{W.~Chen}, \bibinfo{author}{S.~Li}, \bibinfo{author}{H.~Wang}, \bibinfo{author}{J.~Qian}, \bibinfo{author}{X.~Yan},
\newblock \bibinfo{title}{Controllable dialogue simulation with in-context learning},
\newblock in: \bibinfo{editor}{Y.~Goldberg}, \bibinfo{editor}{Z.~Kozareva}, \bibinfo{editor}{Y.~Zhang} (Eds.), \bibinfo{booktitle}{Findings of the Association for Computational Linguistics: EMNLP 2022}, \bibinfo{publisher}{Association for Computational Linguistics}, \bibinfo{address}{Abu Dhabi, United Arab Emirates}, \bibinfo{year}{2022}, pp. \bibinfo{pages}{4330--4347}. \URLprefix \url{https://aclanthology.org/2022.findings-emnlp.318}. \DOIprefix\doi{10.18653/v1/2022.findings-emnlp.318}.
\bibitem[{M{\o}ller et~al.(2024)M{\o}ller, Pera, Dalsgaard, and Aiello}]{moller-etal-2024-parrot}
\bibinfo{author}{A.~G. M{\o}ller}, \bibinfo{author}{A.~Pera}, \bibinfo{author}{J.~Dalsgaard}, \bibinfo{author}{L.~Aiello},
\newblock \bibinfo{title}{The parrot dilemma: Human-labeled vs. {LLM}-augmented data in classification tasks},
\newblock in: \bibinfo{editor}{Y.~Graham}, \bibinfo{editor}{M.~Purver} (Eds.), \bibinfo{booktitle}{Proceedings of the 18th Conference of the European Chapter of the Association for Computational Linguistics (Volume 2: Short Papers)}, \bibinfo{publisher}{Association for Computational Linguistics}, \bibinfo{address}{St. Julian{'}s, Malta}, \bibinfo{year}{2024}, pp. \bibinfo{pages}{179--192}. \URLprefix \url{https://aclanthology.org/2024.eacl-short.17}.
\bibitem[{Zhong et~al.(2024)Zhong, Li, Zhuang, Liu, and Du}]{zhong2024iterativedatagenerationlarge}
\bibinfo{author}{Q.~Zhong}, \bibinfo{author}{H.~Li}, \bibinfo{author}{L.~Zhuang}, \bibinfo{author}{J.~Liu}, \bibinfo{author}{B.~Du}, \bibinfo{title}{Iterative data generation with large language models for aspect-based sentiment analysis}, \bibinfo{year}{2024}. \URLprefix \url{https://arxiv.org/abs/2407.00341}. \href{http://arxiv.org/abs/2407.00341}{\tt arXiv:2407.00341}.
\bibitem[{Dettmers et~al.(2023)Dettmers, Pagnoni, Holtzman, and Zettlemoyer}]{qlora}
\bibinfo{author}{T.~Dettmers}, \bibinfo{author}{A.~Pagnoni}, \bibinfo{author}{A.~Holtzman}, \bibinfo{author}{L.~Zettlemoyer},
\newblock \bibinfo{title}{Qlora: Efficient finetuning of quantized llms},
\newblock in: \bibinfo{editor}{A.~Oh}, \bibinfo{editor}{T.~Naumann}, \bibinfo{editor}{A.~Globerson}, \bibinfo{editor}{K.~Saenko}, \bibinfo{editor}{M.~Hardt}, \bibinfo{editor}{S.~Levine} (Eds.), \bibinfo{booktitle}{Advances in Neural Information Processing Systems}, volume~\bibinfo{volume}{36}, \bibinfo{publisher}{Curran Associates, Inc.}, \bibinfo{year}{2023}, pp. \bibinfo{pages}{10088--10115}. \URLprefix \url{https://proceedings.neurips.cc/paper_files/paper/2023/file/1feb87871436031bdc0f2beaa62a049b-Paper-Conference.pdf}.
\bibitem[{Hu et~al.(2022)Hu, Shen, Wallis, Allen-Zhu, Li, Wang, Wang, and Chen}]{hu2022lora}
\bibinfo{author}{E.~J. Hu}, \bibinfo{author}{Y.~Shen}, \bibinfo{author}{P.~Wallis}, \bibinfo{author}{Z.~Allen-Zhu}, \bibinfo{author}{Y.~Li}, \bibinfo{author}{S.~Wang}, \bibinfo{author}{L.~Wang}, \bibinfo{author}{W.~Chen},
\newblock \bibinfo{title}{Lo{RA}: Low-rank adaptation of large language models},
\newblock in: \bibinfo{booktitle}{International Conference on Learning Representations}, \bibinfo{year}{2022}. \URLprefix \url{https://openreview.net/forum?id=nZeVKeeFYf9}.
\bibitem[{Liu et~al.(2024)Liu, Zhang, Zhao, Luu, and Bing}]{liu2024translationneedstudysolving}
\bibinfo{author}{C.~Liu}, \bibinfo{author}{W.~Zhang}, \bibinfo{author}{Y.~Zhao}, \bibinfo{author}{A.~T. Luu}, \bibinfo{author}{L.~Bing}, \bibinfo{title}{Is translation all you need? a study on solving multilingual tasks with large language models}, \bibinfo{year}{2024}. \URLprefix \url{https://arxiv.org/abs/2403.10258}. \href{http://arxiv.org/abs/2403.10258}{\tt arXiv:2403.10258}.
\bibitem[{{\"U}st{\"u}n et~al.(2024){\"U}st{\"u}n, Aryabumi, Yong, Ko, D{'}souza, Onilude, Bhandari, Singh, Ooi, Kayid, Vargus, Blunsom, Longpre, Muennighoff, Fadaee, Kreutzer, and Hooker}]{ustun-etal-2024-aya}
\bibinfo{author}{A.~{\"U}st{\"u}n}, \bibinfo{author}{V.~Aryabumi}, \bibinfo{author}{Z.~Yong}, \bibinfo{author}{W.-Y. Ko}, \bibinfo{author}{D.~D{'}souza}, \bibinfo{author}{G.~Onilude}, \bibinfo{author}{N.~Bhandari}, \bibinfo{author}{S.~Singh}, \bibinfo{author}{H.-L. Ooi}, \bibinfo{author}{A.~Kayid}, \bibinfo{author}{F.~Vargus}, \bibinfo{author}{P.~Blunsom}, \bibinfo{author}{S.~Longpre}, \bibinfo{author}{N.~Muennighoff}, \bibinfo{author}{M.~Fadaee}, \bibinfo{author}{J.~Kreutzer}, \bibinfo{author}{S.~Hooker},
\newblock \bibinfo{title}{Aya model: An instruction finetuned open-access multilingual language model},
\newblock in: \bibinfo{editor}{L.-W. Ku}, \bibinfo{editor}{A.~Martins}, \bibinfo{editor}{V.~Srikumar} (Eds.), \bibinfo{booktitle}{Proceedings of the 62nd Annual Meeting of the Association for Computational Linguistics (Volume 1: Long Papers)}, \bibinfo{publisher}{Association for Computational Linguistics}, \bibinfo{address}{Bangkok, Thailand}, \bibinfo{year}{2024}, pp. \bibinfo{pages}{15894--15939}. \URLprefix \url{https://aclanthology.org/2024.acl-long.845/}. \DOIprefix\doi{10.18653/v1/2024.acl-long.845}.
\bibitem[{Aryabumi et~al.(2024)Aryabumi, Dang, Talupuru, Dash, Cairuz, Lin, Venkitesh, Smith, Campos, Tan, Marchisio, Bartolo, Ruder, Locatelli, Kreutzer, Frosst, Gomez, Blunsom, Fadaee, Üstün, and Hooker}]{aryabumi2024aya23openweight}
\bibinfo{author}{V.~Aryabumi}, \bibinfo{author}{J.~Dang}, \bibinfo{author}{D.~Talupuru}, \bibinfo{author}{S.~Dash}, \bibinfo{author}{D.~Cairuz}, \bibinfo{author}{H.~Lin}, \bibinfo{author}{B.~Venkitesh}, \bibinfo{author}{M.~Smith}, \bibinfo{author}{J.~A. Campos}, \bibinfo{author}{Y.~C. Tan}, \bibinfo{author}{K.~Marchisio}, \bibinfo{author}{M.~Bartolo}, \bibinfo{author}{S.~Ruder}, \bibinfo{author}{A.~Locatelli}, \bibinfo{author}{J.~Kreutzer}, \bibinfo{author}{N.~Frosst}, \bibinfo{author}{A.~Gomez}, \bibinfo{author}{P.~Blunsom}, \bibinfo{author}{M.~Fadaee}, \bibinfo{author}{A.~Üstün}, \bibinfo{author}{S.~Hooker}, \bibinfo{title}{Aya 23: Open weight releases to further multilingual progress}, \bibinfo{year}{2024}. \URLprefix \url{https://arxiv.org/abs/2405.15032}. \href{http://arxiv.org/abs/2405.15032}{\tt arXiv:2405.15032}.
\bibitem[{Mitra et~al.(2023)Mitra, Corro, Mahajan, Codas, Simoes, Agarwal, Chen, Razdaibiedina, Jones, Aggarwal, Palangi, Zheng, Rosset, Khanpour, and Awadallah}]{mitra2023orca}
\bibinfo{author}{A.~Mitra}, \bibinfo{author}{L.~D. Corro}, \bibinfo{author}{S.~Mahajan}, \bibinfo{author}{A.~Codas}, \bibinfo{author}{C.~Simoes}, \bibinfo{author}{S.~Agarwal}, \bibinfo{author}{X.~Chen}, \bibinfo{author}{A.~Razdaibiedina}, \bibinfo{author}{E.~Jones}, \bibinfo{author}{K.~Aggarwal}, \bibinfo{author}{H.~Palangi}, \bibinfo{author}{G.~Zheng}, \bibinfo{author}{C.~Rosset}, \bibinfo{author}{H.~Khanpour}, \bibinfo{author}{A.~Awadallah}, \bibinfo{title}{Orca 2: Teaching small language models how to reason}, \bibinfo{year}{2023}. \URLprefix \url{https://arxiv.org/abs/2311.11045}. \href{http://arxiv.org/abs/2311.11045}{\tt arXiv:2311.11045}.
\bibitem[{{\v{S}}m{\'\i}d(2023)}]{vsmid2023cross}
\bibinfo{author}{J.~{\v{S}}m{\'\i}d}, \bibinfo{title}{Cross-lingual Aspect-Based Sentiment Analysis}, Master's thesis, University of West Bohemia, Faculty of Applied Sciences, Plzeň, \bibinfo{year}{2023}.
\bibitem[{Ding et~al.(2024)Ding, Qin, Zhao, Luo, Li, Chen, Xia, Hu, Luu, and Joty}]{ding2024dataaugmentationusinglarge}
\bibinfo{author}{B.~Ding}, \bibinfo{author}{C.~Qin}, \bibinfo{author}{R.~Zhao}, \bibinfo{author}{T.~Luo}, \bibinfo{author}{X.~Li}, \bibinfo{author}{G.~Chen}, \bibinfo{author}{W.~Xia}, \bibinfo{author}{J.~Hu}, \bibinfo{author}{A.~T. Luu}, \bibinfo{author}{S.~Joty}, \bibinfo{title}{Data augmentation using large language models: Data perspectives, learning paradigms and challenges}, \bibinfo{year}{2024}. \URLprefix \url{https://arxiv.org/abs/2403.02990}. \href{http://arxiv.org/abs/2403.02990}{\tt arXiv:2403.02990}.
\bibitem[{Zhang et~al.(2024)Zhang, Xie, Hou, Zhao, Lin, and Wen}]{zhang2023recommendationinstructionfollowinglarge}
\bibinfo{author}{J.~Zhang}, \bibinfo{author}{R.~Xie}, \bibinfo{author}{Y.~Hou}, \bibinfo{author}{X.~Zhao}, \bibinfo{author}{L.~Lin}, \bibinfo{author}{J.-R. Wen},
\newblock \bibinfo{title}{Recommendation as instruction following: A large language model empowered recommendation approach},
\newblock \bibinfo{journal}{ACM Trans. Inf. Syst.}  (\bibinfo{year}{2024}). \bibinfo{note}{Just Accepted}.

\end{thebibliography}

\end{document}